\pdfoutput=1
\documentclass[screen,acmsmall]{acmart} 
\usepackage{multirow}
\usepackage{graphicx}
\usepackage[english]{babel}
\usepackage{enumitem}
\usepackage{listings}
\usepackage{amsmath,mathrsfs}
\usepackage{tcolorbox}
\usepackage{enumitem} 
\usepackage{ifthen}
\usepackage{hyperref}
\usepackage{svg}
\usepackage{subcaption}
\usepackage{makecell}
\definecolor{DarkGreen}{rgb}{0.0,0.5,0.0}
\definecolor{sustainability}{HTML}{2ECC71} 
\definecolor{popularity}{HTML}{3498DB}     
\definecolor{personalization}{HTML}{FFBF00} 
\definecolor{moderator}{HTML}{9B59B6}
\definecolor{SASI}{HTML}{E74C3C}      
\definecolor{MASI}{HTML}{2754F5}      
\definecolor{MAMI}{HTML}{2ECC71}      
\usepackage{fontawesome5}
\usepackage{pifont}
\usepackage{booktabs}
\usepackage{xcolor}
\usepackage{colortbl}

\newcommand{\minSign}{\textcolor{blue}{\ensuremath{\mathbf{\downarrow}}}}
\newcommand{\maxSign}{\textcolor{blue}{\ensuremath{\mathbf{\uparrow}}}}

\newcommand{\maxHighlight}[1]{\textbf{#1}}

\newcommand{\sigcell}{\cellcolor{green!20}}

\newcommand{\change}[1]{\textcolor{black}{#1}}

\newcommand{\yashar}[1]{\textcolor{black}{#1}}

\newcommand{\tors}[1]{\textcolor{black}{#1}}
\newcommand{\torsRTwo}[1]{\textcolor{black}{#1}}
\newcommand{\torsRThree}[1]{\textcolor{black}{#1}}

\newcommand{\gemini}{\textit{Gemini}}
\newcommand{\olmo}{\textit{Olmo-7b}}
\newcommand{\gemmaTwelve}{\textit{Gemma-12b}}
\newcommand{\gemmaFour}{\textit{Gemma-4b}}
\newcommand{\gptOss}{\textit{GPT-OSS}}

\newcommand{\claude}{\textit{Claude}}

\newcommand{\MILP}{\textsc{MILP}}

\usepackage{xcolor}
\definecolor{rqboxframe}{HTML}{0B6E69} 
\definecolor{rqboxbg}{HTML}{E7F4F3}    

\newcommand{\RQBox}[2]{%
  \begin{center}
    \fcolorbox{rqboxframe}{rqboxbg}{%
      \parbox{0.97\linewidth}{\textbf{#1}~#2}%
    }%
  \end{center}
}

\newcommand{\Query}{\mathbb{Q}}

\newcommand{\filters}{\mathcal{F}}
\newcommand{\agent}{\mathit{a}_i}

\newcommand{\round}{\mathit{t}}

\newcommand{\collectiverejection}{\Phi'_\round}

\newcommand{\collectiveoffer}{\Phi_\round}

\newcommand{\candidate}{\mathit{L}_{\agent,\round}}

\newcommand{\reliabilityscore}{\ensuremath{d}_{\agent, \round}}

\newcommand{\moderator}{\ensuremath{\mathcal{M}}}
\newcommand{\relevancescore}{\ensuremath{r}_{\agent, \round}}

\newcommand{\hallucinationscore}{\ensuremath{h}_{\agent, \round}}

\newcommand{\MAMIearly}{\ensuremath{\mathsf{M}_{\text{early}}}}
\newcommand{\MAMIfull}{\ensuremath{\mathsf{M}_{10}}}
\newcommand{\MASI}{\ensuremath{\mathsf{M}_{1}}}

\newcommand{\sysname}{\textsc{Collab-Rec}}

\usepackage{tcolorbox}

\newtcolorbox{promptbox}[1][]{
  colback=yellow!10,
  colframe=DarkGreen!90,
  boxrule=0.5pt,
  arc=4pt,
  left=6pt,
  right=6pt,
  top=6pt,
  bottom=6pt,
  fonttitle=\bfseries,
  fontupper=\ttfamily,
  title=#1,
}
\AtBeginDocument{%
  }
\addto\extrasenglish{%
}
\setcopyright{cc}
\setcctype{by-nc-nd}
\acmJournal{TORS}
\acmYear{2026} \acmVolume{1} \acmNumber{1} \acmArticle{}
\acmMonth{8} \acmDOI{10.1145/3837862}

\begin{document}

\title[\sysname{}]{\sysname{}: An LLM-based Agentic Framework for Balancing Recommendations in Tourism}

\author{Ashmi Banerjee}
\email{ashmi.banerjee@tum.de}
\orcid{https://orcid.org/0000-0002-4217-3888}
\affiliation{%
  \institution{Technical University of Munich}
  \city{Munich}
  \country{Germany}
}
\author{Adithi Satish}
\email{adithi.satish@tum.de}
\orcid{https://orcid.org/0000-0003-2602-5356}
\affiliation{%
  \institution{Technical University of Munich}
  \city{Munich}
  \country{Germany}
}

\author{Fitri Nur Aisyah}
\email{fitri.aisyah@tum.de}
\orcid{https://orcid.org/0009-0003-2423-9788}
\affiliation{%
  \institution{Technical University of Munich}
  \city{Munich}
  \country{Germany}
}

\author{Wolfgang W\"orndl}
\email{woerndl@in.tum.de}
\orcid{https://orcid.org/0000-0003-2972-5817}
\affiliation{%
  \institution{Technical University of Munich}
  \city{Munich}
  \country{Germany}
}

\author{Yashar Deldjoo}
\email{yashar.deldjoo@poliba.it}
\orcid{https://orcid.org/0000-0002-6767-358X}

\affiliation{
  \institution{Polytechnic University of Bari}
  \city{Bari} 
  \country{Italy} 
}

\renewcommand{\shortauthors}{Banerjee et al.}


\begin{abstract}
We propose \sysname{}, a multi-agent framework designed to counteract popularity bias and enhance diversity in tourism recommendations. In our setting, three LLM-based agents (\textit{Personalization}, \textit{Popularity}, and \textit{Sustainability}) generate city suggestions from complementary perspectives. A non-LLM moderator then merges and refines these proposals through \torsRTwo{iterative constrained refinement}, ensuring each agent's viewpoint is incorporated while penalizing spurious or repeated responses.

\tors{Extensive \torsRTwo{offline} experiments on European city queries using LLMs from different sizes and model families} demonstrate that \sysname{} enhances diversity and overall relevance compared to a single-agent baseline, surfacing lesser-visited locales that are often overlooked. 
\torsRTwo{This balanced, context-aware approach better reflects a broader range of user and system-level considerations, highlighting the potential of multi-stakeholder collaboration in LLM-driven recommender systems.}

Code, data, and other artifacts are available here:
\url{https://github.com/ashmibanerjee/collab-rec} while the prompts used are included in the appendix.

\end{abstract}






\maketitle

\section{Introduction}
\label{section: intro}

Tourism recommender systems (RSs) \tors{that suggest travel destinations} are increasingly expected to serve \emph{multiple stakeholders} simultaneously.
Beyond tailoring suggestions to a traveler’s constraints and interests, platforms often optimize for engagement and business
objectives (frequently correlated with destination popularity), while destinations and policymakers increasingly require
\emph{sustainability-aware demand shaping} to mitigate overtourism, seasonal concentration, and spatial congestion.
These requirements naturally induce \emph{competing objectives}: recommendations that are highly personalized can still
over-concentrate demand on a small set of iconic hubs; conversely, aggressively pushing long-tail destinations can degrade
relevance when user constraints are not respected~\cite{balakrishnan2021multistakeholder, abdollahpouri2021multistakeholder}.

Unlike many retail settings, where recommending a popular product mostly affects conversion, tourism recommendations can
shape physical flows of visitors across space and time.
Repeatedly steering demand toward a few ``must-see'' cities can exacerbate congestion externalities and reduce the quality of
experience for both visitors and residents~\cite{dodds2019phenomena}.
At the same time, tourism RSs remain accountable to the individual traveler: a recommendation list is only useful if it satisfies
hard constraints (e.g., travel dates, budget, seasonal preferences) and aligns with stated interests (e.g., museums, nature, nightlife).
This combination makes tourism a prototypical \emph{multi-stakeholder} recommendation setting, where user utility, platform utility,
and destination-level sustainability objectives must be balanced rather than optimized in isolation~\cite{abdollahpouri2020multistakeholder, Jannach}. 

\subsection{LLMs as Travel Recommenders: Capabilities and Shortcomings}
Large language models (LLMs) enable conversational travel recommendation, allowing users to express complex, multi-intent
requirements in natural language. \change{For example: \textit{``walkable European cities in September, mid-budget, with museums and cultural
events, but not overcrowded''.}}
Recent generative recommenders demonstrate that LLMs can improve interaction quality through dialogue, explanations,
and nuanced preference elicitation~\cite{gao2023chatrec, lubos2024llm, lyu2024llm, yang2023palr}.
However, monolithic LLM recommenders remain brittle when asked to satisfy multiple simultaneous constraints and to
balance stakeholder objectives.
Two failure modes are especially problematic in tourism: (\emph{i}) \textbf{popularity dominance}, where the model repeatedly
returns canonical tourist hubs even when users ask for ``hidden gems'' or when sustainability goals discourage concentration; and
\change{(\emph{ii}) \textbf{hallucinations}}, where the model fabricates destinations or attributes (e.g., incorrect sustainability claims),
or returns out-of-catalog entities that cannot be validated in a deployment setting~\cite{deldjoo2025toward, staab2023beyond, jiang2025beyond, sakib2024challenging}.
Since these decisions are produced end-to-end inside the model, it is often unclear \emph{why} a particular trade-off
was made, and how the output would change under different stakeholder priorities~\cite{li2024survey}.


\subsection{Limitations of Prior LLM-Based Approaches: Why Agentic Design?}
A natural baseline for LLM recommenders is single-shot prompt engineering: ask one model to satisfy all constraints and
``balance'' objectives.
In our setting, this approach has three practical limitations.
First, mixing multiple objectives in a single prompt often produces unstable behavior: the model may implicitly prioritize
popular options or ignore sustainability-oriented constraints.
Second, prompt-only specialization is sensitive to surface form and decoding choices; small paraphrases can change which objective dominates.
Third, single-shot pipelines offer limited mechanisms for \emph{auditing} and \emph{controlling} how constraints are enforced,
especially when recommendations must be grounded in a fixed catalog.
\change{Recent work on agentic and multi-agent LLM systems --- including dedicated surveys of agentic recommender systems --- synthesizes architectures and open challenges, highlighting controllability, trustworthiness, and efficiency as central barriers to deployment~\cite{maragheh2025future}.}

Distributing objectives across specialist agents allows each agent to focus on a single stakeholder dimension, which can reduce the
``objective collapse'' often observed in monolithic LLM outputs and can surface a broader candidate set before aggregation.
However, specialization alone is insufficient: each agent can still overfit to its own biases, and hallucinations remain possible.


We therefore combine role specialization with a \emph{deterministic, non-LLM moderator} that implements two \change{practical desiderata for deployment}:
(\emph{i}) \textbf{catalog grounding} (ensuring outputs map to a known inventory), and
(\emph{ii}) \textbf{transparent aggregation} (making the trade-off policy explicit and reproducible).
The moderator does not reason about proposals; instead, it applies explicit constraint checks and feedback signals that determine whether agents regenerate lists in subsequent rounds. 
While an LLM could in principle act as a flexible moderator, doing so would introduce additional cost and prompt sensitivity,
and would make it harder to disentangle improvements due to role specialization versus improvements due to an additional model.
A deterministic moderator provides a strong, auditable control point that supports ablation analysis. 

\subsection{Our Approach}
\begin{figure*}[htbp]
    \centering
    \includegraphics[width=\linewidth]{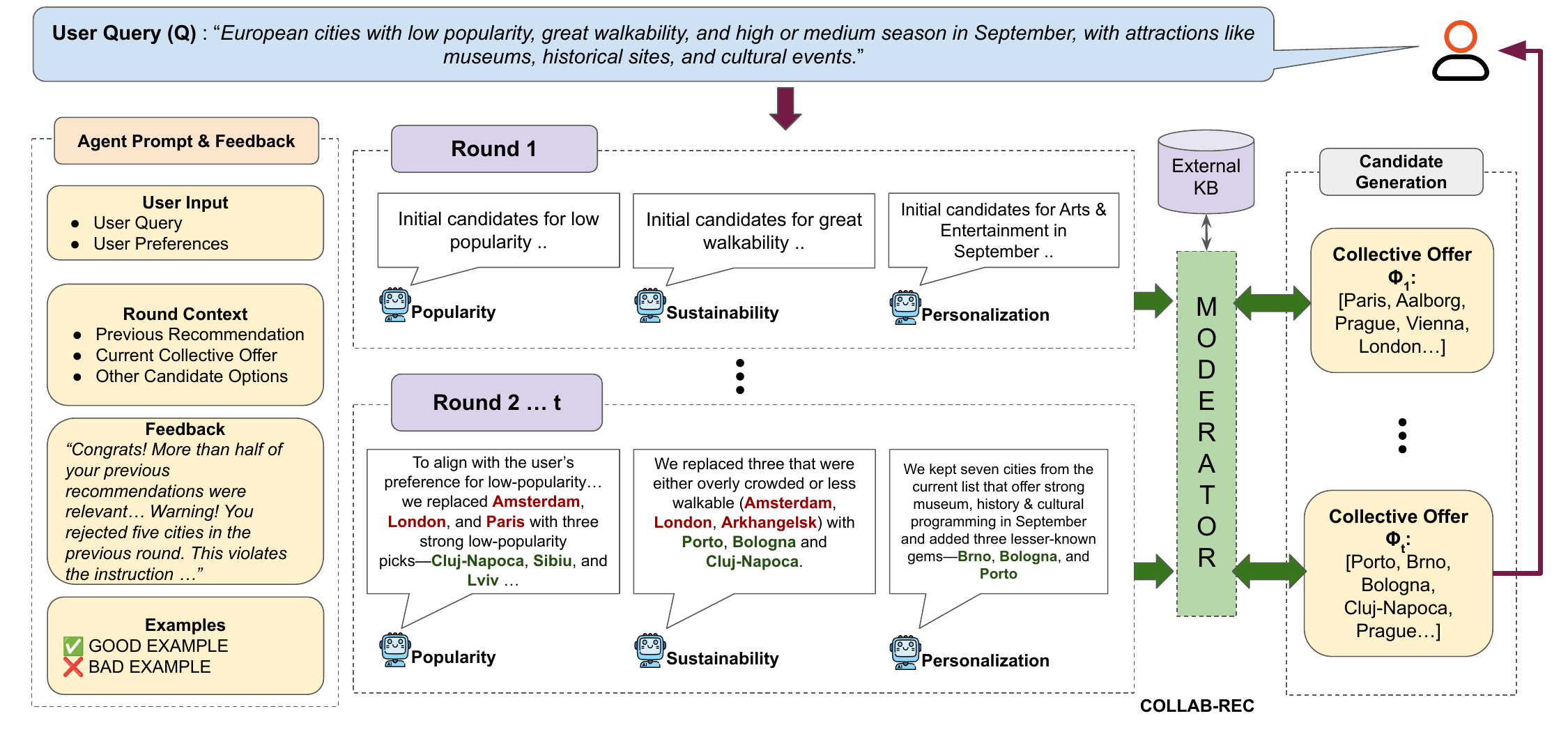}
    \caption{Overview of the \sysname{} workflow to generate city trip recommendations using multiple LLM agents. The non-LLM \textbf{Moderator} evaluates and combines the agent proposals, iteratively refining the final recommendation set, which is then communicated to the user.}
    \label{fig: workflow}
\end{figure*}

We propose \sysname{}, a moderator-mediated, multi-round \change{refinement} framework for \emph{grounded}
multi-stakeholder tourism recommendation (\autoref{fig: workflow}).
Given a user query $Q$, three specialist LLM agents generate candidate destination sets from complementary perspectives:
a \emph{Personalization} agent that emphasizes user constraints and interests, a \emph{Popularity} agent that reasons about
mainstream appeal and the popularity preference expressed in the query, and a \emph{Sustainability} agent that promotes
alternatives that mitigate concentration and seasonality effects.
The moderator then (i) grounds candidates to a structured city catalog, (ii) scores candidates using transparent per-objective diagnostics
(success, reliability, hallucination penalties), and (iii) computes rejection decisions and broadcasts diagnostic feedback that explicitly constrain agents' subsequent proposals. \change{Agents do not directly reason about each other's outputs; instead, they regenerate lists in response to deterministic moderator signals.}
This design makes the system modular: additional stakeholder agents (e.g., safety, accessibility) can be integrated without retraining.

\paragraph{Efficiency-aware multi-round \change{refinement}.}
Multi-round coordination increases latency and cost.
To make the framework more practical, we introduce a patience-based early stopping protocol that dynamically terminates coordination
when improvements stagnate.
Empirically, across models, recommendation quality \change{plateaus by $\sim$4--5 rounds} (\autoref{section: RQ1}); early stopping captures most gains while reducing inference time substantially for API-served models.



\paragraph{Empirical study.}
We conduct a large-scale \change{offline} evaluation on 900 stratified tourism queries derived from SynthTRIPs~\cite{banerjee2025synthtrips} and a grounded catalog of 200 European cities.
We benchmark six LLM backbones spanning proprietary and open-source families \change{(\texttt{claude-sonnet-4-5}, \texttt{gemini-2.5-flash}, \texttt{gpt-oss-20b}, \texttt{gemma-3-12b}, \texttt{olmo3-7b-instruct}, \texttt{gemma-3-4b})}, and compare against (i) non-LLM baselines (\textsc{RandRec}, \textsc{TopPop}, and a structured \textsc{MILP} optimizer), (ii) a single-agent baseline
(\textsc{SASI}), and (iii) a single-round multi-agent baseline (\textsc{MASI}).
We evaluate \emph{grounded recommendation quality} primarily using moderator success (constraint satisfaction under catalog validation),
complemented by diversity and concentration metrics (Gini, entropy, and catalog coverage) and agent behavior metrics (reliability and
hallucination tendency).
\change{Across model families, \sysname{} consistently improves grounded success over \textsc{SASI} and \textsc{MASI}, while also reducing recommendation concentration relative to them, leading to more diverse recommendations (\autoref{section: results}). Although the \MILP{} reference achieves the highest moderator success score, it operates on fully structured catalog inputs with a hand-specified utility function. We therefore treat \MILP{} as a structured-input reference rather than evidence that \sysname{} outperforms all non-LLM recommenders.}

\paragraph{Contributions.}
Our main contributions are as follows:
\begin{itemize}
  \item \textbf{Problem formulation.} We formalize multi-stakeholder, multi-constraint tourism recommendation as a
  \emph{grounded} multi-objective ranking problem, where feasibility is defined with respect to an explicit destination catalog
  and constraint satisfaction (\autoref{section: preliminaries}).
  \item \textbf{Framework design.} We introduce \sysname{}, a modular multi-agent architecture with a transparent,
  deterministic moderator that scores candidates under multiple objectives and iteratively conditions agent outputs via structured feedback (\autoref{section: architecture}--\ref{section: grounding}).
  \item \textbf{Efficiency-aware moderation.} We propose and empirically validate a patience-based early stopping protocol and analyze the
  effect of two rejection strategies (\textit{Aggressive} vs.\ \textit{Majority}) on convergence, stability, and compute cost (\autoref{section: termination}).
  \item \textbf{Large-scale evaluation and analysis.} We evaluate 900 queries across six LLM families, report statistical testing and
  convergence behavior, and analyze the relevance--diversity--cost trade-off induced by multi-round \change{refinement} (\autoref{section: experiments}--\ref{section: results}).
  \item \textbf{Reproducibility.} We release code, prompts, and evaluation artifacts to enable reproduction and extension.
\end{itemize}

\paragraph{Scope and limitations.}
Our goal is to provide a reproducible, controllable blueprint for balanced tourism recommendations grounded in catalogs\change{, not a production-ready system}.
We do not claim to have developed a new foundational multi-agent learning algorithm.
Rather, we study how role specialization, multi-round \change{refinement}, and grounded moderation affect relevance, diversity,
hallucination behavior, and efficiency in a \torsRThree{controlled research} setting.

\paragraph{Paper organization.}
\autoref{section: related} reviews related work on LLM-based agents, multi-agent recommender systems, and multi-objective recommendation.
\autoref{section: framework} presents the \sysname{} framework and moderator design.
\autoref{section: experiments} describes the experimental setup and evaluation protocol.
\autoref{section: results} reports results and discussion organized around our research questions, including efficiency and robustness analysis.
Finally,~\autoref{section: conclusion} concludes with limitations and future directions. 

\section{Related Work}
\label{section: related}

This section reviews (i) LLM-based agents and multi-agent interaction protocols (\autoref{subsection: llm_agents}), (ii) multi-agent recommender systems (\autoref{subsection: mars}),
(iii) classical multi-objective and multi-stakeholder recommendation (\autoref{subsection: moo}), and \tors{(iv) grounding and hallucination control in LLM recommenders (\autoref{subsection: hallucination_rw})}.
We then position \sysname{} within this landscape and clarify how it complements (rather than replaces) established
optimization and reranking approaches in~\autoref{subsection: positioning}.

\subsection{LLM-based Agents and Multi-Agent Interaction} \label{subsection: llm_agents}
LLM-based agents --- LLMs augmented with role instructions, memory, and (optionally) tools---have been studied as a mechanism
to decompose complex tasks into interacting components.
Recent surveys summarize common agentic architectures and workflows, emphasizing communication protocols, evaluation, and
application domains such as web search and scientific question answering~\cite{guo2024large, wu2023autogen, yehudai2025survey,peng2025survey,zhang2025survey}.
Empirical studies also highlight limitations: agent success can be brittle, and scaling the number of agents does not automatically
improve outcomes unless the interaction protocol is carefully designed~\cite{jiang2025beyond}.
From a recommender-systems perspective, recent work synthesizes definitions and open challenges for agentic RSs and multi-agent RSs,
highlighting controllability, trustworthiness, and efficiency as central barriers to deployment~\cite{maragheh2025future}.

A prominent family of interaction protocols is \emph{multi-agent debate} and round-table consensus, where agents critique each other’s
responses over multiple rounds to reach an agreement~\cite{du2023improving, tran2025multi}.
Such debate-feedback mechanisms can improve performance on reasoning and summarization tasks without additional training data~\cite{chen2025debate, chen2024reconcile, zhang2024exploring, chun2025multi}.
However, forcing explicit consensus can reduce diversity; role differentiation and implicit agreement mechanisms have been proposed as
a way to preserve diversity while maintaining broad consistency~\cite{wu2025hidden}.
These observations are directly relevant to tourism recommendation, where a system must often balance competing objectives rather than
optimize a single notion of ``correctness.''

\subsection{Multi-Agent Recommender Systems} \label{subsection: mars}
Multi-Agent Recommender Systems (MARS) leverage agent specialization to support recommendation-related subtasks
(e.g., user preference interpretation, item understanding, search, explanation).
Recent LLM-based conversational recommendation frameworks employ agents with memory and external tools to elicit preferences and
generate recommendations without large-scale training~\cite{wang2023recmind}.
Architectures such as MACRS~\cite{fang2024multi} and MACRec~\cite{wang2024macrec} adopt a centralized manager coordinating specialized sub-agents, often supported by a search agent querying external sources~\cite{guangtao2024hybrid}.
MATCHA extends this pattern with safeguard and explanation agents for video game recommendation~\cite{hui2025matcha}.
More broadly, surveys examine the bidirectional relationship between LLM agents and recommender systems, documenting how agents can
enhance recommendation pipelines and how recommendation concepts (e.g., user modeling, feedback loops) can inform agent design~\cite{zhu2025recommender}.

Most existing MARS work, however, focuses on improving personalization or interaction quality in a single objective setting.
Tourism-specific multi-agent designs that explicitly balance user preferences, popularity-driven platform incentives, and sustainability
constraints remain underexplored.
Moreover, many frameworks rely on LLM-based management or open-world retrieval; while powerful, this can make factual grounding and
reproducibility harder to guarantee when the system must adhere to a fixed catalog.

\subsection{Multi-Objective and Multi-Stakeholder Recommendations} \label{subsection: moo}
Balancing competing objectives has long been studied in RSs.
Multi-objective RSs optimize multiple criteria (e.g., accuracy, diversity, novelty, fairness) via scalarization (weighted sums),
constrained optimization, or Pareto-based approaches~\cite{zheng2022survey,deb2002fast}.
Multi-stakeholder recommendation generalizes this view by explicitly modeling utilities for multiple parties (e.g., users, providers,
platforms) and analyzing trade-offs among them~\cite{amigo2023unifying}.

A widely adopted practical strategy is \emph{post-processing reranking}: given a candidate set, rerank items to improve diversity or reduce
concentration using rank-discounted trade-offs (e.g., MMR~\cite{carbonell1998use}, xQuAD~\cite{santos2010exploiting}) or topic-based diversification in recommendation lists~\cite{ziegler2005improving}.
These approaches are highly effective when item features and objective scores are readily available.

In open-ended tourism queries, however, the first challenge is to produce a feasible candidate set that satisfies natural-language
constraints and is grounded in an explicit inventory.
Our work is therefore complementary: \sysname{} uses specialist generation plus catalog validation to create feasible candidates,
and then applies a transparent scalarization in the moderator to aggregate objectives (\autoref{section: framework}). The resulting pipeline can be seen as an \emph{agentic front-end} that enables multi-objective list construction in a setting where constraint satisfaction and grounding are
first-order requirements.

\subsection{Grounding and Hallucination Control in LLM Recommenders} \label{subsection: hallucination_rw}
Hallucination and factuality concerns have motivated substantial work on grounding LLM outputs, including retrieval-augmented generation,
structured output constraints, and post-hoc verification~\cite{mohammadabadi2025survey,kazlaris2025illusion,anh2025survey,arslan2024survey,e2025rag}.
In recommendation settings, recent research also argues that evaluation should go beyond utility and incorporate aspects such as factual
validity, reasoning quality, and robustness, particularly when LLMs are used to generate items or item attributes~\cite{jiang2025beyond,deldjoo2025toward}.
Tourism recommendation amplifies these concerns because out-of-catalog or incorrectly attributed destinations can lead to poor user experience and safety risks.

\subsection{Positioning \sysname{}} \label{subsection: positioning}
\sysname{} is positioned at the intersection of agentic RSs and multi-objective recommendation:
it targets open-ended, multi-constraint travel queries; decomposes stakeholder objectives into role-specific generation agents; and enforces
catalog grounding and transparent aggregation via a deterministic moderator.
Compared to manager-centric multi-agent recommenders (e.g., MACRec~\cite{wang2024macrec}, MATCHA~\cite{hui2025matcha}), \sysname{} emphasizes
\emph{multi-stakeholder balancing} and provides per-objective diagnostics (success, reliability, hallucination penalties) that are directly
measurable and ablatable.
Compared to classical reranking and multi-objective optimization methods, \sysname{} addresses the upstream challenge of generating and repairing feasible
candidates under natural-language constraints, while still retaining a transparent scalarization layer that can incorporate established
multi-objective principles.

\paragraph{Comparison summary.}
In short, \sysname{} differs from the closest lines of work along four axes:
\begin{itemize}
  \item \textbf{Problem focus:} multi-stakeholder \emph{tourism} recommendation with explicit attention to popularity concentration and sustainability-oriented balancing;
  \item \textbf{Mechanism:} moderator-mediated \emph{multi-round \change{refinement}} with structured feedback, rather than single-shot prompting or a manager-only architecture;
  \item \textbf{Grounding:} explicit catalog validation and hallucination-aware rejection policies to enforce inventory compliance; and
  \item \textbf{Evaluation:} large-scale analysis across six LLM families with statistical testing, convergence/early-stopping behavior, and explicit relevance--diversity--cost trade-offs (\autoref{section: exp_setup}--\ref{section: results}).
\end{itemize}

\section{Agentic Recommendation Framework}
\label{section: framework}

This section formalizes \sysname{}, a multi-agent recommendation framework for city-trip recommendations that explicitly balances multiple stakeholder objectives. Given a natural-language user query, three specialized LLM agents propose ranked candidate cities from a fixed catalog. A deterministic moderator then grounds, evaluates, and aggregates the proposals through a repeated coordination loop. The loop continues until an online termination criterion indicates convergence or stagnation, after which the final ranked recommendation list is returned.

\begin{figure}[htbp]
    \centering
    \includegraphics[width=0.9\linewidth,trim={0 0 2.5cm 0},clip]{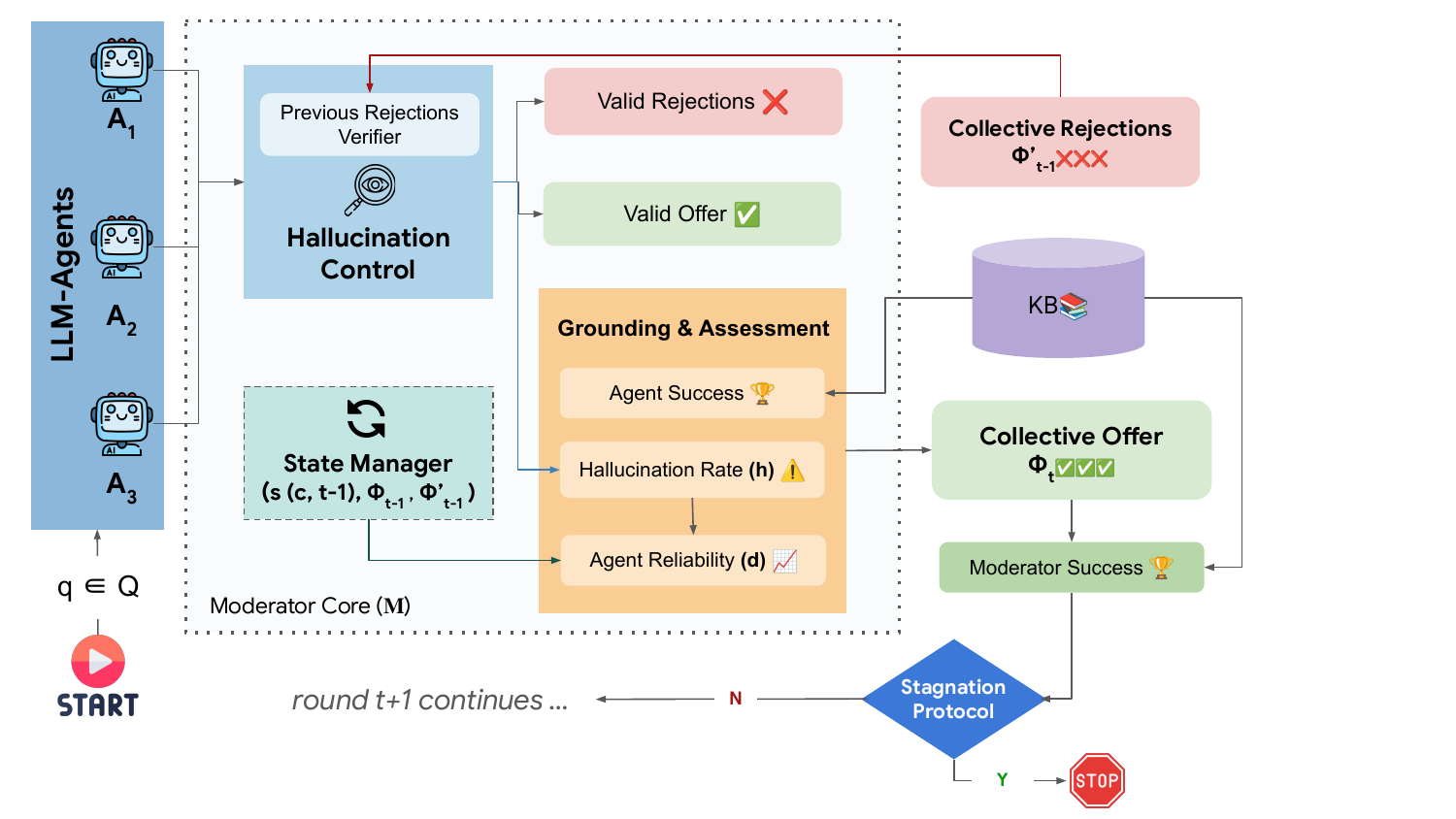}
    \caption{Overview of \sysname{}. Three specialist LLM agents independently propose ranked city candidates. A deterministic moderator validates proposals against a knowledge base, computes grounded diagnostics (constraint satisfaction, stability across rounds, and hallucination/invalid-output rate), aggregates candidates into a collective offer, and broadcasts structured feedback and a rejection set. Iteration continues until an online termination protocol is satisfied.}
    \label{fig: moderator-core}
\end{figure}

\subsection{Preliminaries and System Goal}
\label{section: preliminaries}

\paragraph{Problem setup.}
We consider a city-trip recommendation task in which the input is a natural-language user query $\Query$. The query encodes (i) textual preferences (for example, ``quiet coastal destinations''), and (ii) an explicit set of structured filters
$\filters=\{f_1,f_2,\dots,f_m\}$ (for example, budget range, travel month, activity categories, or sustainability preferences).
Recommendations are drawn from a closed catalog of candidate cities $\mathcal{C}$, where each city $c \in \mathcal{C}$ is associated with structured attributes stored in an external knowledge base (\autoref{section: kb}).

\yashar{In our experiments, the natural-language query is accompanied by benchmark-provided structured filters used for validation and scoring. Thus, \sysname{} should not be interpreted as a fully structure-free raw-text recommender. Its language-model component is responsible for candidate generation and revision conditioned on the natural-language request, while the moderator uses the structured filters and closed catalog for deterministic grounding and constraint checks.}
The goal is to output a ranked list $\collectiveoffer$ of exactly $k$ cities that balances multiple stakeholder objectives under the constraints expressed by $\Query$ and $\filters$. In \sysname{}, $k=10$ in all experiments.

\paragraph{Notation.}
\torsRTwo{We denote the set of agents by $\mathcal{A}=\{a_1,a_2,a_3\}$, and the round index by $t \in \{1,\dots,T\}$, where each round corresponds to a step in an iterative constrained refinement process.}
At round $t$, agent $a_i$ produces a ranked list of $k$ candidate cities, denoted $\candidate$.
The moderator maintains two shared state variables:
(i) the \emph{collective offer} $\collectiveoffer$, a ranked list of $k$ cities representing the current best aggregated proposal; and
(ii) the \emph{collective rejection set} $\Phi'_t$, containing cities that are disallowed in subsequent rounds under the active rejection strategy (\autoref{section: rejections}).

\paragraph{System goal.}
Let $s(c, t)$ denote the cumulative moderator score assigned to city $c$ after round $t$.
At termination round $T$, \sysname{} returns the size-$k$ ranked list that maximizes the cumulative score:
\begin{equation}
\Phi_T \;=\; \arg\max_{L \subseteq \mathcal{C},\,|L|=k}\;\; \sum_{c\in L} s(c,T).
\label{eq:objective}
\end{equation}
In practice, $\collectiveoffer$ is constructed online via repeated aggregation and scoring (\autoref{section: protocol}--\ref{section: scoring}) rather than by exhaustive search over $\binom{|\mathcal{C}|}{k}$ combinations.

\subsection{Architecture and Moderator-Mediated Multi-Round Coordination}
\label{section: architecture}

\subsubsection{System components}
\label{section: components}
The system consists of three specialist agents and a deterministic moderator, as illustrated in \autoref{fig: moderator-core}.

\paragraph{Specialist Agents.}
\sysname{} instantiates three specialist agents, each representing a distinct stakeholder objective commonly studied in tourism recommender systems:
\begin{itemize}

    \item \textbf{Personalization \change{Agent}} $a_1$: focuses on the user-centric perspective, prioritizing explicit filters and query-specific preferences such as budget, travel month, and interests.
    \item \textbf{Popularity \change{Agent}} $a_2$: emphasizes the popularity dimension and is configured to mitigate short-head concentration by proposing less popular cities when the query suggests a preference for less crowded destinations.
    \item \textbf{Sustainability \change{Agent}} $a_3$: prioritizes sustainability-related attributes such as walkability, seasonality, and air-quality indicators, and defaults to environmentally preferable cities when the query does not specify sustainability constraints, consistent with multi-stakeholder tourism perspectives~\cite{banerjee2023review}.
\end{itemize}
\change{Each agent is powered by an LLM backbone and is }constrained to output exactly $k$ catalog city names in a structured schema (\autoref{section: structured_output}). To ensure fair comparisons across models, the three agents share the same LLM backbone within each experimental run (\autoref{section: exp_settings}).

\paragraph{Moderator.}
The moderator $M$ is a deterministic (non-LLM) controller (\autoref{fig: moderator-core}) that (i) validates proposals against the catalog and query constraints, (ii) computes grounded diagnostic scores, (iii) aggregates proposals into an updated collective offer $\Phi_t$, and (iv) broadcasts structured feedback and the updated rejection set $\Phi'_t$.
The moderator has access to a structured knowledge base that provides attributes for every $c \in \mathcal{C}$, enabling catalog grounding, constraint checks, and objective measurements.

\subsubsection{What ``\torsRTwo{iterative constrained refinement}'' means in \sysname{}}
\label{section: negotiation_definition}

We use \emph{\torsRTwo{iterative constrained refinement}} to denote a moderator-mediated coordination loop, not a debate protocol in which agents directly message, strategize, or bargain. Agents do not exchange messages \change{or coordinate with each other}; instead, \change{each agent iteratively refines its own proposals in response to the moderator's structured feedback and the shared evolving state.}
\change{We use the terms ``rounds'' and ``iterations'' interchangeably to refer to steps in the refinement loop.}

Each round consists of four steps:
\begin{enumerate}
    \item \textbf{Proposal:} each agent $a_i$ proposes a ranked list $L_{a_i,t}$ of $k$ destinations aligned with its role.
    \item \textbf{Grounding and assessment:} the moderator validates items, computes grounded diagnostics (constraint satisfaction, stability, and invalid-output rate), and updates city scores using a transparent multi-objective policy (\autoref{section: scoring}--\ref{section: grounding}).
    \item \textbf{Feedback broadcast:} the moderator publishes the collective offer $\collectiveoffer$ and the collective rejection set $\Phi'_t$, and generates structured feedback that summarizes each agent's behavior (for example, invalid items, excessive churn, or insufficient alignment with role-specific filters).
    \item \textbf{Revision:} agents condition on $\collectiveoffer$ and $\Phi'_t$ and revise their proposals in the next round by repairing invalid suggestions, exploring alternatives, and adapting to moderator feedback.
\end{enumerate}

Multi-round execution can improve outcomes through three empirically testable mechanisms:
(i) \emph{repair} (iterative removal and replacement of invalid or constraint-violating candidates),
(ii) \emph{feedback-driven exploration} (moving away from over-recommended short-head destinations when they conflict with the objectives), and
(iii) \emph{stabilization} (convergence of the collective offer as marginal improvements diminish), which motivates early stopping (\autoref{section: termination}).

\subsection{Interaction Protocol and Operational Design}
\label{section: protocol}

\subsubsection{Agent prompting and controlled revision}
\label{section: prompting}

At $t=1$, each agent receives the user query $\Query$ and the relevant filters for its role (\autoref{section: components}), and returns a ranked list of $k$ catalog cities.
For rounds $t>1$, each agent additionally receives:
(i) the current collective offer $\Phi_{t-1}$,
(ii) a role-specific feedback message generated by the moderator, and
(iii) a set of disallowed cities $\Phi'_{t-1}$.
Agents are instructed to produce a \emph{new} ranked list of exactly $k$ cities at every round and to \change{retain at least $k-3$ cities from the current collective offer $\Phi_{t-1}$, modifying at most three relative to it}, which operationalizes limited ``offer revision'' inspired by iterative offer-and-veto style protocols~\cite{erlich2018negotiation}. The limited replacement budget encourages continuity and makes reliability measurable (\autoref{section: RQ3}).

\subsubsection{Hallucination control via structured output constraints}
\label{section: structured_output}

LLM recommenders may produce non-grounded items, including destinations outside the catalog or contradicting explicit constraints~\cite{jiang2025beyond}.
To mitigate this risk, \sysname{} enforces a structured output schema that requires:
(i) exactly $k$ ranked city names,
(ii) JSON-serializable fields, and
(iii) fixed field names and types for downstream validation.
\change{While structured output constraints reduce hallucinations by enforcing syntactic validity, they do not fully guarantee catalog membership. Final grounding is enforced by the moderator through post-generation validation.}

Within \sysname{}, an output is treated as \emph{invalid} if it recommends a city that is either
(i) not in the catalog $\mathcal{C}$, or
(ii) present in the collective rejection set $\collectiverejection$ under the active rejection strategy.
We quantify invalidity via the hallucination metric defined in~\autoref{section: hallucination}.

\subsubsection{Constructing the collective offer}
\label{section: offer}

After scoring candidate cities at round $t$, the moderator produces the collective offer $\collectiveoffer$ by selecting the top-$k$ cities under min--max normalized scores \cite{patro2015normalization}.

Let
\[
\operatorname{norm}(s(c, t))=\frac{s(c, t)-\min_{c'\in\mathcal{C}} s(c', t)}{\max_{c'\in\mathcal{C}} s(c', t)-\min_{c'\in\mathcal{C}} s(c', t)}.
\]
Then the collective offer is defined as:

\begin{equation}
\collectiveoffer 
= 
\operatorname{arg\,top}_k 
\left[ 
\operatorname{norm}(s(c, t)) 
\right], 
\quad c \in \mathcal{C}.
\label{eq:collective_offer}
\end{equation}

where $\operatorname{arg\,top}_k$ returns the $k$ cities with the largest normalized scores, ordered from highest to lowest. 
\subsubsection{Aggregating rejections}
\label{section: rejections}

The collective rejection set $\collectiverejection$ removes cities that are deemed unacceptable under a voting policy based on agent omissions.
Let $\mathbb{I}[\cdot]$ be the indicator function, and define the number of ``omit votes'' for city $c$ at round $t$ as


\[
v_t(c)
\;=\;
\sum_{a_i \in \mathcal{A}}
\mathbb{I}\!\left[\, c \notin L_{a_i,t} \right],
\qquad \forall\, c \in \Phi_{t-1}.
\]

We consider two rejection strategies:
\begin{itemize}
    \item \textbf{\textit{Aggressive} rejection:} reject $c$ if \emph{any} agent omits it, that is, $v_t(c)\ge 1$.
    \item \textbf{\textit{Majority} rejection:} reject $c$ if \emph{at least two} agents omit it, that is, $v_t(c)\ge 2$.
\end{itemize}
The collective rejection set is updated as:
\begin{equation}
\Phi'_t \;=\; \Phi'_{t-1} \;\cup\; \{\,c \in \Phi_{t-1} \;:\; v_t(c)\ge \tau\,\},
\label{eq:rejection_update}
\end{equation}
where $\tau=1$ for \textit{Aggressive} rejection\change{,} and $\tau=2$ for \textit{Majority} rejection.
Cities in $\Phi'_t$ are disallowed in subsequent rounds for both the agents and the moderator.

\subsection{Scoring and Decision Policy}
\label{section: scoring}

\torsRTwo{The moderator's role is to combine individual agent diagnostics into a collective score that reflects the system's multi-objective optimization goals. This aggregation is a canonical example of multi-criteria decision-making~\cite{aruldoss2013survey} and social choice theory~\cite{sen1986social}, where diverse agent evaluations must be fused into a single collective decision.
For each agent $a_i$ and round $t$, the moderator aggregates three grounded diagnostics: 
(i) agent success (relevance) $r_{a_i,t}$, measuring constraint alignment, 
(ii) agent reliability $d_{a_i,t}$, capturing temporal stability across rounds, and 
(iii) hallucination rate $h_{a_i,t}$, indicating the fraction of invalid outputs. 
These diagnostics are combined via a transparent linear scalarization with inverse-rank discounting, providing both interpretability and the ability to perform ablation analyses (\autoref{section: RQ5}).}

\paragraph{Rank-discounted scalarization.}

For a candidate city $c$ proposed by agent $a_i$ at round $t$, let $\operatorname{rank}_{a_i,t}(c)\in\{1,\dots,k\}$ denote its rank position in $L_{a_i,t}$.
We define the incremental contribution of agent $a_i$ to city $c$ at round $t$ as:
\begin{equation}
\Delta s_{a_i}(c,t)
\;=\;
\frac{1}{\operatorname{rank}_{a_i,t}(c)}
\Bigl(
\lambda_r\, r_{a_i,t}
+
\lambda_d\, d_{a_i,t}
-
\lambda_h\, h_{a_i,t}
\Bigr),
\label{eq:incremental_score}
\end{equation}
where $\lambda_r\ge 0$, $\lambda_d\ge 0$, and $\lambda_h\ge 0$ are weighting coefficients.

\paragraph{Scope of the diagnostic terms.}
\yashar{The terms $r_{a_i,t}$, $d_{a_i,t}$, and $h_{a_i,t}$ are agent or list-level diagnostics rather than item-specific utility estimates. They measure, respectively, how well an agent's complete list aligns with its assigned filters, how stable that list is across rounds, and how often the agent produces invalid outputs. Consequently, every city proposed by the same agent in the same round receives the same diagnostic multiplier, modulated only by rank. Item-level differentiation enters through catalog feasibility, the reciprocal-rank discount, repeated endorsement by multiple agents and rounds, and rejection constraints. This design is intentionally transparent and auditable, but it can still assign a diagnostic boost to a weak item appearing in an otherwise strong list. We therefore interpret the score as a heuristic coordination signal rather than an item-level utility model.}


\yashar{The term $1/\operatorname{rank}_{a_i,t}(c)$ is a reciprocal-rank positional weight. It is related to positional aggregation in spirit but is not a standard linear Borda count, which would assign scores such as $k-\operatorname{rank}$ for a list of length $k$. Reciprocal-rank discounting attenuates lower-ranked items more strongly, thereby concentrating influence on top-ranked proposals. We adopt this parameter-free discount for interpretability; alternative positional functions, including linear Borda-style or logarithmic DCG-style discounts, remain natural variants for future work.
}



\paragraph{Cumulative scoring across rounds.}
Scores accumulate across rounds, reflecting repeated endorsement by agents and persistence under moderator grounding:
\begin{equation}
s(c, t)
\;=\;
s(c,t-1)
+
\sum_{a_i\in\mathcal{A}}
\mathbb{I}[c \in L_{a_i,t}]\;\Delta s_{a_i}(c,t),
\label{eq:cumulative_score}
\end{equation}
with $s(c,0)=0$ for all $c\in\mathcal{C}$.
Unless stated otherwise, we use $\lambda_r=\lambda_d=\lambda_h=1$ to preserve interpretability and to avoid tuning on the evaluation set. 
\torsRThree{Ablation analyses over scoring \emph{components} (i.e., removing $r$, $d$, or $h$ terms entirely) are reported in~\autoref{section: RQ5}.}

\paragraph{Positioning of the moderator score.}
\change{\autoref{eq:incremental_score} is a transparent heuristic scalarization of grounded diagnostics, not a learned, calibrated, or theoretically optimal objective. We fix $\lambda_r=\lambda_d=\lambda_h=1$ \emph{a priori} to preserve interpretability and to avoid post-hoc tuning on the evaluation set.}
\change{The ablations in~\autoref{section: RQ5} remove scoring components under these fixed weights; they should not be interpreted as a sensitivity analysis over the $\lambda$ parameters.}

\subsection{Grounding and Assessment}
\label{section: grounding}

In each round, the moderator computes grounded diagnostics for each agent based on the knowledge base attributes and the catalog constraints.

\subsubsection{Agent success}
\label{section: success}

Agent success measures how well an agent aligns its recommendations with the filters assigned to its role.
Let $f_{a_i}\subseteq \filters$ denote the subset of query filters assigned to agent $a_i$.
Let $\mathcal{M}(c)$ denote the set of filters satisfied by city $c$ under the knowledge base metadata (for example, a city satisfies a budget filter if its cost attribute lies in the requested range).
Then \change{the} agent's success is:
\begin{equation}
r_{a_i,t}
\;=\;
\frac{1}{|L_{a_i,t}|}
\sum_{c \in L_{a_i,t}}
\frac{|\mathcal{M}(c)\cap f_{a_i}|}{|f_{a_i}|},
\label{eq:agent_success}
\end{equation}
where $r_{a_i,t}\in[0,1]$.

\change{We distinguish two success metrics: \textit{agent success} (\autoref{eq:agent_success}) measures constraint satisfaction for an agent's proposed list against its assigned filters $f_{a_i}$, while \textit{moderator success} (\autoref{eq: moderator_success}) measures average filter satisfaction of the collective offer $\Phi_t$ against the full filter set $\mathcal{F}$. We use the terms ``success'' and ``relevance'' interchangeably when referring to the agent-level metric.}

\subsubsection{Agent reliability}
\label{section: reliability}

Agent reliability quantifies how stable an agent's ranked list is across consecutive rounds.
Let $A=L_{a_i,t-1}$ and $B=L_{a_i,t}$.
We define a rank-deviation operator:
\begin{equation}
\Delta(A,B)
=
\sum_{x \in A\cap B}\bigl|\operatorname{rank}_{A}(x)-\operatorname{rank}_{B}(x)\bigr|
+
|A\setminus B|\cdot \mu_1
+
|B\setminus A|\cdot \mu_2,
\label{eq:rank_deviation}
\end{equation}
where $\mu_1$ penalizes dropped candidates and $\mu_2$ penalizes newly introduced candidates.

\change{We set both penalties to the same scalar:}
\begin{equation}
\mu_1 = \mu_2 = |L_{a_i,t}|,
\label{eq: mu1}
\end{equation}
\change{so that $|L_{a_i,t-1}|\cdot(\mu_1+\mu_2)$ is a conservative upper bound on $\Delta(A,B)$, guaranteeing $d_{a_i,t}\in[0,1]$.}

Reliability is then:
\begin{equation}
d_{a_i,t}
=
\max\!\left(
0,\;
1-\frac{\Delta(L_{a_i,t-1},L_{a_i,t})}{|L_{a_i,t-1}|\cdot(\mu_1+\mu_2)}
\right),
\label{eq: reliability}
\end{equation}
with $d_{a_i,t}\in[0,1]$.

\subsubsection{Hallucination rate}
\label{section: hallucination}

The hallucination rate measures the fraction of invalid recommendations in $L_{a_i,t}$, where ``invalid'' means either out-of-catalog or rejected by the moderator.
Let the currently feasible set be $\mathcal{C}\setminus \Phi'_t$.
Then:
\begin{equation}
h_{a_i,t}
=
\frac{1}{k}
\sum_{j=1}^{k}
\mathbb{I}\!\left[\change{L_{a_i,t}[j]} \notin (\mathcal{C}\setminus \Phi'_t)\right],
\label{eq: hallucination_rate}
\end{equation}
where $\change{L_{a_i,t}[j]}$ denotes the city at rank position $j$ in $L_{a_i,t}$.
By construction, $h_{a_i,t}\in[0,1]$, and it is subtracted in the score update through~\autoref{eq:incremental_score}.

\subsection{Termination Criteria and Complexity}
\label{section: termination}

A fixed-round budget can be wasteful when improvements plateau early, but greedy stopping can terminate prematurely when scores fluctuate across rounds.
\sysname{} therefore uses an online termination protocol based on the moderator's success score for the collective offer.

\paragraph{Moderator success for termination.}
Let $\filters$ be the full set of query filters. We define moderator success as the average fraction of satisfied filters over the current collective offer:
\begin{equation}
S(\Phi_t)
=
\frac{1}{|\Phi_t|}
\sum_{c\in \Phi_t}
\frac{|\mathcal{M}(c)\cap \filters|}{|\filters|},
\label{eq: moderator_success}
\end{equation}
so that $S(\Phi_t)=1$ indicates that all recommended cities satisfy all query constraints under knowledge-base grounding.

\paragraph{Online termination protocol.}
The process terminates if either of the following holds:
\begin{enumerate}
    \item \textbf{Ideal convergence:} if $S(\Phi_t)=1$, the process stops immediately at round $t$.
    \item \textbf{Patience-based stagnation:} after a minimum exploration phase of $T_{\min}$ rounds, the process stops if improvements are below a threshold $\epsilon$ over a sliding patience window of length $p$:
    \begin{equation}
    t \ge T_{\min}
    \;\wedge\;
    \left(
    \max_{i\in\{0,\dots,p\}} S(\Phi_{t-i}) - S(\Phi_{t-p})
    < \epsilon
    \right).
    \label{eq: early_stop}
    \end{equation}
\end{enumerate}

\paragraph{Complexity considerations.}
Each round requires $|\mathcal{A}|$ LLM generations and one deterministic moderator pass over at most $|\mathcal{A}|\cdot k$ proposed items.
The dominant cost is LLM inference. In our implementation, agents are executed in parallel, so the wall-clock time per round is close to the slowest agent call plus moderator overhead.
The early stopping criterion reduces the expected number of rounds, and~\autoref{section: RQ4} quantifies the resulting latency and token savings.

\section{Experiments}
\label{section: experiments}

This section describes the experimental design used to evaluate \sysname{}. We detail the dataset and knowledge base, implementation and orchestration settings, baselines, model backbones, evaluation metrics, and the research questions guiding the analysis.

\subsection{Setup}
\label{section: exp_setup}

\subsubsection{Dataset}
\label{section: dataset}

We evaluate on a stratified sample of $900$ queries from SynthTRIPs~\cite{banerjee2025synthtrips}, a knowledge-grounded benchmark for personalized tourism recommendation. The benchmark contains over $4{,}000$ synthetic, natural-language queries designed to reflect diverse user travel intents and sustainability preferences.

To obtain balanced coverage while keeping computational cost tractable, we stratify by two axes provided by the benchmark:
(i) \emph{popularity preference level} (low, medium, high) and
(ii) \emph{query complexity tier} (medium, hard, sustainable).
We uniformly sample $100$ queries from each of the $3\times 3=9$ strata, yielding $900$ total queries.
The resulting queries range from broad requests (for example, planning a short, budget-friendly trip to a less crowded coastal destination) to more constrained prompts that include multiple filters (for example, budget constraints, month constraints, and explicit sustainability requirements).
\tors{Notably, SynthTRIPs queries stem from LLMs themselves, but we
exclude those generated by \gemini{} (a model family included among our evaluated backbones) to reduce the risk of model-specific stylistic advantages. Instead, we rely solely on queries generated using \textit{llama-3.2-90B}, thereby ensuring a clearer separation between query generation and model evaluation.}


\subsubsection{External knowledge base}
\label{section: kb}

Grounding and validation are performed using the SynthTRIPs knowledge base~\cite{banerjee2025synthtrips}, which contains a closed catalog of $200$ European cities. Each city is annotated with attributes relevant to the objectives of this paper, including indicators for popularity, budget, seasonality, and sustainability. The moderator uses this knowledge base to:
(i) validate that recommended items belong to the catalog,
(ii) check satisfaction of structured filters,
and (iii) compute the grounded diagnostics used in scoring (\autoref{section: grounding}).

\subsection{Experimental Settings}
\label{section: exp_settings}

\subsubsection{Implementation details}
\label{section: implementation}

\paragraph{Primary configuration.}
Our main evaluation focuses on the multi-agent, multi-round configuration of \sysname{}, referred to as the \emph{\change{Multi-Agent Multi-Iteration}} (MAMI) configuration. For every query, the system instantiates the three specialist agents (\textit{Personalization}, \textit{Popularity}, \textit{Sustainability}) and runs the moderator-mediated coordination loop described in~\autoref{section: protocol}. Each agent produces a ranked list of length $k=10$ at each round.

\paragraph{Decoding parameters and structured output.}
For all models that expose sampling controls, we set temperature to $0.5$ and top-$p$ to $0.95$. These parameters are chosen to allow controlled exploration while maintaining stability across rounds. All agents are constrained to produce structured outputs, and the moderator validates outputs against the catalog and the rejection set (\autoref{section: structured_output}).

\paragraph{Round budget and termination.}
The system is allowed up to $T_{\max}=10$ rounds, but it typically stops earlier via the online termination protocol in~\autoref{section: termination}. In our experiments, we use:
$T_{\min}=3$ (minimum exploration rounds),
$p=2$ (patience window length),
and $\epsilon=0.005$ (stagnation threshold in moderator success).
We report results for both (i) early-stopped runs (\change{\MAMIearly{}}) and (ii) full $10$-round runs (\MAMIfull{}), \change{to separate early-stopped behavior from the full round-budget setting.}

\paragraph{Orchestration.}
We implement \sysname{} with the Google Agent Development Kit (ADK)\footnote{\url{https://google.github.io/adk-docs}} to support modular role separation, looping, and parallel agent execution. Agents execute in parallel with an independent local state, while the deterministic moderator maintains global state, applies the rejection strategy, performs grounding checks, and generates structured feedback.

\paragraph{Initialization.}
At $t=1$, the first proposals are generated without a prior collective state. For the reliability and hallucination diagnostics, we initialize $d_{a_i,0}=1$ and $h_{a_i,0}=0$ and then compute true values from round $t=1$ onward.

\paragraph{Additional configuration for ablations.}
To study ablations (RQ5), we run an additional set of experiments on $150$ queries with two representative model backbones (\gemini{} and \olmo{}), using the \textit{Aggressive} rejection strategy and a maximum of five rounds under the same early-stopping mechanism (see~\autoref{section: RQ5}).

\subsubsection{Baselines}
\label{section: baselines}

We compare \sysname{} against a diverse set of baselines designed to isolate different aspects of recommendation performance, including popularity bias, single-agent reasoning, and classical multi-objective optimization.

\begin{itemize}
     \item \textbf{Random recommender (\textsc{RandRec}).}
    A non-language-model baseline that ignores the query and returns a reproducible random set of $k=10$ cities~\cite{luctan2024literature}.
    \item \textbf{Top popularity recommender (\textsc{TopPop}).}
    A non-personalized baseline that returns the globally most popular cities, independent of user constraints~\cite{cremonesi2011looking}.

    \item \yashar{\textbf{Constraint Optimizer (\MILP{}) baseline.}
To provide a controlled optimization-based reference, we introduce a Mixed-Integer Linear Programming (MILP) baseline that selects exactly $k=10$ cities from the catalog:
\begin{equation}
\max_{x_1,\ldots,x_n} \sum_{i=1}^{n} x_i u_i
\quad \text{s.t.} \quad x_i \in \{0,1\}, \quad \sum_{i=1}^{n}x_i=k .
\end{equation}
The utility of city $i$ is defined as
\begin{equation}
u_i = r_i + s_i + \tilde{p}_i,
\end{equation}
where $r_i$ and $s_i$ denote the relevance and sustainability scores computed from the same filter-based signals used by the moderator, and $\tilde{p}_i$ is a conditional popularity score. Specifically, $\tilde{p}_i=p_i$ when the city's categorical popularity level matches the popularity preference expressed in the query, and $\tilde{p}_i=0$ otherwise. Thus, the term rewards alignment with the requested popularity level; it is not an inverse-popularity penalty and does not uniformly discourage popular cities.}

\yashar{This MILP should be interpreted as a structured-input reference rather than a complete non-LLM recommender baseline. It assumes that the query has already been converted into structured filters and that all candidate utilities are available. It also optimizes additive utility under a cardinality constraint and does not model iterative candidate repair, explanation traces, or diversification objectives such as MMR/xQuAD-style coverage, Pareto/weighted reranking, or constrained diversification. Stronger catalog-only baselines of this kind are important future comparators, and we do not claim practical superiority over them in this paper.}

    \item \textbf{Single-Agent Single-Iteration (\textsc{SASI}).}
    A single LLM is prompted with the full query and returns one ranked list of $k=10$ cities without iterative refinement or a specialist decomposition.
    \item \textbf{Multi-Agent Single-Iteration (\textsc{MASI}).}
    The three specialist agents each propose an initial list; the moderator grounds and aggregates once, without any additional coordination rounds.
    \change{For brevity in tables, we also write $\mathsf{M}_1$ for \textsc{MASI}, as a single-round \textsc{MAMI} run is equivalent to \textsc{MASI}.}
\end{itemize}

\torsRTwo{While open-ended conversational queries do not provide exhaustive ground-truth labels or a verifiable set of globally optimal recommendations, the \textsc{MILP} baseline serves as a controlled and interpretable reference point, complementing \textsc{RandRec} and \textsc{TopPop}. 
Together, these baselines enable us to quantify how \sysname{} balances multiple objectives --- relevance, diversity, and fairness in practice, even when the true optimal recommendation set is unobservable.}

\subsubsection{Large language model backbones}
\label{section: models}

We evaluate six reasoning-capable LLMs spanning proprietary and open-source families and a range of parameter scales:
\begin{itemize}
    \item \textbf{Large proprietary models:} Gemini (\texttt{gemini-2.5-flash})~\cite{comanici2025gemini}, and Claude (\texttt{\change{claude-sonnet-4-5}})~\cite{anthropic2025claude_sonnet_4_5}.
    \item \textbf{Medium-scale open models:} \texttt{gpt-oss-20b}~\cite{agarwal2025gpt}, and \texttt{gemma-3-12b}~\cite{team2025gemma}.
    \item \textbf{Small-scale open models:} \texttt{gemma-3-4b}~\cite{team2025gemma}, and \texttt{olmo3-7b-instruct}~\cite{olmo2025olmo}.
\end{itemize}
\change{Throughout the paper, we drop version suffixes and refer to these models as \gemini{}, \claude{}, \gptOss{}, \gemmaTwelve{}, \gemmaFour{}, and \olmo{} respectively; these short forms are used consistently in all tables, figures, and body text.}

\subsection{Evaluation Metrics}
\label{section: metrics}

We evaluate \sysname{} from three complementary perspectives: \textit{final recommendation quality}, \textit{per-round agent behavior}, and \textit{computational overhead}.

\subsubsection{Final recommendation quality}

\paragraph{\change{Grounded relevance via moderator success (\autoref{eq: moderator_success}).}}
Since open-ended tourism queries lack explicit interaction logs and canonical relevance labels, we measure \change{moderator success} \change{as defined in~\autoref{eq: moderator_success}}, \change{the fraction of recommended cities that satisfy all query constraints under catalog validation}. A value of $1$ indicates that all recommended cities satisfy all structured constraints under knowledge base validation.

\paragraph{Popularity-bias and diversity.}
To quantify concentration on short-head destinations, we compute the Gini index~\cite{gastwirth1972estimation} and normalized entropy~\cite{jost2006entropy} over the distribution of recommended cities aggregated across queries.
Lower Gini indicates less concentration, and higher normalized entropy indicates a more even distribution.

\paragraph{Catalog coverage.}
We additionally report coverage, defined as the fraction of the $200$-city catalog that appears at least once in the recommendations. Coverage complements Gini and entropy by directly capturing long-tail breadth.

\subsubsection{Agent behavior (per round)}

\paragraph{Reliability.}
We report agent reliability $d_{a_i,t}$ (\autoref{eq: reliability}), which measures stability of an agent's ranked list across consecutive rounds.

\paragraph{Invalid-output rate.}
We report hallucination rate $h_{a_i,t}$ (\autoref{eq: hallucination_rate}), defined as the fraction of an agent's recommendations that are out-of-catalog or violate the collective rejection constraint.

\subsubsection{Computational overhead}

We measure (i) wall-clock time per query and (ii) total token usage. Both are aggregated over all agent calls and moderator operations across all executed rounds. These metrics quantify the practical cost of multi-round \change{refinement} and motivate early stopping (\autoref{section: termination}).

\subsection{Research Questions}
\label{sec:research_questions}

The experiments are structured around five research questions:
\begin{description}
    \item[\textbf{RQ1}] Does multi-agent, multi-round \change{refinement} improve grounded recommendation quality compared to single-agent prompting, single-round aggregation, and non-language-model baselines?
    \item[\textbf{RQ2}] Does multi-round \change{refinement} reduce popularity concentration and increase long-tail coverage of destinations?
    \item[\textbf{RQ3}] How do specialist agents evolve across rounds in terms of reliability and invalid-output behavior under moderator feedback?
    \item[\textbf{RQ4}] What latency and token overhead does multi-round \change{refinement} introduce, and how effectively does online early stopping mitigate these costs?
    \item[\textbf{RQ5}] How sensitive is the method to its scoring components and design choices, as assessed by ablation studies?
\end{description}

\autoref{section: results} reports results and discussion for RQ1--RQ5, with supplementary plots for additional rejection strategies provided in the appendix.

\section{Results and Discussion}
\label{section: results}

We evaluate whether enabling multiple specialist agents to coordinate over multiple rounds (\textsc{MAMI}) yields tangible
benefits over (i) traditional non-LLM baselines (\textsc{RandRec}, \textsc{TopPop}), (ii) a single-agent single-iteration baseline
(\textsc{SASI}), and (iii) a multi-agent single-iteration baseline (\textsc{MASI}).
Unless stated otherwise, results are reported over 900 stratified queries (\autoref{section: dataset}) and averaged across queries.
For \textsc{MAMI}, we report two operational modes: a \textit{patience-based early-stopped} variant (\MAMIearly) and a full 10-round
run (\MAMIfull).
We further compare two rejection strategies in the moderator: \textit{Aggressive} (discard any flagged city) and \textit{Majority}
(discard only if at least two agents flag a city\change{).}


Unless noted otherwise, figures in the main text report \textsc{MAMI} with the \textit{Aggressive} rejection strategy (\autoref{fig: agent_adaptability} shows both variants).  Tables include results for both \textit{Aggressive} and \textit{Majority}. Detailed results for the \textit{Majority} strategy are provided in the~\autoref{appendix: relevance results}–\ref{appendix: complexity results}.

\textbf{Interpretation of ``multi-round \torsRTwo{iterative constrained refinement}''}
Throughout this section, we use ``coordination'' or \torsRTwo{``refinement''}in a pragmatic sense: agents do not communicate directly or
strategize against each other; instead, they iteratively \emph{adapt} to structured moderator feedback.
Mechanistically, \textsc{MAMI} can be understood as an iterative constrained search/optimization loop that alternates between
candidate generation (agents) and feasibility-aware aggregation (moderator), with multi-round execution providing additional opportunities
to repair constraint violations and escape short-head popularity modes.

\subsection{RQ1: System-level impact on grounded recommendation quality}
\label{section: RQ1} 

\noindent\textbf{RQ1 asks:} \emph{Does the multi-agent, multi-round (\textsc{MAMI}) approach improve final recommendation quality
compared to \textsc{SASI}, \textsc{MASI}, \torsRThree{and the simple non-LLM baselines (\textsc{RandRec}, \textsc{TopPop})?}} 
\torsRThree{To contextualize the results, we also compare against the \textsc{MILP} reference, which operates over fully structured inputs and is not a direct competitor in the natural-language setting addressed by \sysname{}.}
We operationalize system-level quality primarily via \emph{moderator success}, i.e., the fraction of recommended cities that satisfy
query constraints under catalog validation (\autoref{section: metrics}).

\paragraph{Overall effectiveness across models and strategies.}
\autoref{tab: relevance_results} shows that multi-round \change{refinement} improves grounded quality for all evaluated model families and for both rejection policies.
The improvements are substantial when comparing \textsc{MAMI} to \textsc{SASI}.
For example, under \textit{Aggressive} rejection, \claude{} increases from 0.465 (\textsc{SASI}) to 0.657 (\MAMIearly{}),
and \olmo{} increases from 0.536 to 0.666.
Even compared to the stronger single-round multi-agent baseline (\textsc{MASI}), \textsc{MAMI} remains consistently better:
e.g., under \textit{Aggressive} rejection, \gemini{} improves from 0.616 (\textsc{MASI}) to 0.648 (\MAMIearly{}),
and \gemmaTwelve{} improves from 0.618 to 0.647.

\begin{table}[htbp]
\centering
\small
\caption{\tors{Performance comparison across model sizes and rejection strategies.
Average moderator success scores are reported for SASI, $\mathsf{M}_1$ (=MASI), \MAMIearly{}, and \MAMIfull{} (full 10-round run without early stopping). The higher the score, the better.
Best scores per model and rejection strategy (Rej. Strat.) are highlighted in \textbf{bold}.
\change{\textbf{Succ. Rate (\%)} = fraction of queries where \MAMIearly{} moderator success strictly exceeds that of $\mathsf{M}_1$ (=\textsc{MASI}).
\textbf{Odds Ratio} = $\text{Wins}_{\MAMIearly{}\,\text{vs}\,\mathsf{M}_1}$\,/\,$\text{Losses}_{\MAMIearly{}\,\text{vs}\,\mathsf{M}_1}$, i.e., the ratio of queries where \MAMIearly{} strictly outperforms $\mathsf{M}_1$ to queries where it strictly underperforms, excluding ties, computed using a Fisher's exact test.}}}
\label{tab: relevance_results}
\setlength{\tabcolsep}{4pt}
\resizebox{\textwidth}{!}{
\begin{tabular}{lllcccccccc}
\toprule
\multirow{3}{*}{\textbf{\begin{tabular}[c]{@{}l@{}}Model \\ Size\end{tabular}}}& \multirow{3}{*}{\textbf{Model}} & \multirow{3}{*}{\textbf{\begin{tabular}[c]{@{}l@{}}Rej.\\ Strat.\end{tabular}}}
& \multicolumn{4}{c}{\moderator{} \textbf{Success Scores}\maxSign{}} 
& \multicolumn{4}{c}{\textbf{Early-Stopping Metrics}} \\
\cmidrule(lr){4-7} \cmidrule(lr){8-11}
& & 
& \multicolumn{2}{c}{\textbf{LLM-Baselines}} 
& \multicolumn{2}{c}{\textbf{Proposed}} 
& \multirow{2}{*}{\begin{tabular}[c]{@{}c@{}}\textbf{Succ. Rate (\%)}\\ \textbf{\MAMIearly $>$ \MASI}\end{tabular}}
& \multirow{2}{*}{\begin{tabular}[c]{@{}c@{}}\textbf{Avg. Conv.}\\ \textbf{Round}\end{tabular}}
& \multirow{2}{*}{\begin{tabular}[c]{@{}c@{}}\textbf{Odds}\\ \textbf{Ratio}\end{tabular}}
& \multirow{2}{*}{\begin{tabular}[c]{@{}c@{}}\textbf{Ties (\%)}\\ \textbf{\MAMIearly $=$ \MASI}\end{tabular}} \\
\cmidrule(lr){4-5} \cmidrule(lr){6-7}
& & 
& \textbf{SASI} & \textbf{\MASI} 
& \textbf{\MAMIearly} & \textbf{\MAMIfull} 
& & & & \\
\midrule

\multirow{4}{*}{\textbf{Big}}
& \multirow{2}{*}{Claude} 
& A & \multirow{2}{*}{0.465} & \multirow{2}{*}{0.594} & \maxHighlight{0.657} & 0.620 & 59.6 & 3.72 & 17.54 & 26.2 \\
& & M &  &  & \maxHighlight{0.647} & 0.632 & 50.7 & 3.82 & 16.58 & 36.9 \\
\cmidrule(lr){2-11}
& \multirow{2}{*}{Gemini}
& A & \multirow{2}{*}{0.520} & \multirow{2}{*}{0.616} & \maxHighlight{0.648} & 0.617 & 58.4 & 3.73 & 15.58 & 26.8 \\
& & M &  &  & \maxHighlight{0.636} & 0.621 & 49.1 & 3.96 & 12.50 & 37.0 \\
\midrule

\multirow{4}{*}{\textbf{Mid-sized}}
& \multirow{2}{*}{GPT-OSS-20b} 
& A & \multirow{2}{*}{0.632} & \multirow{2}{*}{0.640} & \maxHighlight{0.661} & 0.625 & 51.9 & 3.73 & 4.89 & 24.7 \\
& & M &  &  & \maxHighlight{0.655} & 0.634 & 47.4 & 3.87 & 5.62 & 32.6 \\
\cmidrule(lr){2-11}
& \multirow{2}{*}{Gemma-12b} 
& A & \multirow{2}{*}{0.616} & \multirow{2}{*}{0.618} & \maxHighlight{0.647} & 0.617 & 57.9 & 3.79 & 18.85 & 28.8 \\
& & M &  &  & \maxHighlight{0.637} & 0.621 & 49.4 & 4.00 & 13.52 & 37.1 \\
\midrule

\multirow{4}{*}{\textbf{Small}}
& \multirow{2}{*}{Olmo-7b} 
& A & \multirow{2}{*}{0.536} & \multirow{2}{*}{0.597} & \maxHighlight{0.666} & 0.624 & 70.1 & 3.84 & 17.46 & 13.1 \\
& & M &  &  & \maxHighlight{0.662} & 0.628 & 67.0 & 3.88 & 16.82 & 16.7 \\
\cmidrule(lr){2-11}
& \multirow{2}{*}{Gemma-4b} 
& A & \multirow{2}{*}{0.615} & \multirow{2}{*}{0.617} & \maxHighlight{0.650} & 0.618 & 61.9 & 3.78 & 23.05 & 25.2 \\
& & M &  &  & \maxHighlight{0.636} & 0.620 & 48.3 & 4.01 & 16.84 & 39.9 \\
\bottomrule
\end{tabular}}
\vspace{1mm}
\footnotesize
\textit{Non-LLM baselines: RandRec (RR) = 0.501, TopPop (TP) = 0.676, \torsRTwo{\textsc{MILP} = 0.730}; reported for reference and omitted from the table body for clarity.}

\end{table}

\begin{table}[htbp]
\centering
\footnotesize
\caption{Pairwise statistical significance comparisons across models and rejection strategies (\textsc{MAMI} with early stopping, i.e., \MAMIearly{}). P-values are Bonferroni-corrected over $m=3$ planned comparisons ($\alpha_{\mathrm{corr}} = 0.05/3 \approx 0.017$); the null hypothesis ($H_0$) assumes equal performance between the compared methods. Significant results ($p_{\mathrm{corr}} < 0.05$, equivalently raw $p < 0.017$) are highlighted in green. Each cell reports: $t$-stat $(p \mid p_{\mathrm{corr}})$.}
\label{tab: statistical_significance}
\begin{tabular}{l l c c c}
\toprule
\multirow{2}{*}{\textbf{Model}} &
\multirow{2}{*}{\begin{tabular}[c]{@{}l@{}}\textbf{Rejection}\\ \textbf{Strategy}\end{tabular}} &
\multicolumn{3}{c}{\textbf{Groups}} \\
\cmidrule(lr){3-5}
& & \textbf{\MAMIearly vs \MASI} & \textbf{\MAMIearly vs SASI} & \textbf{\MASI vs SASI} \\
\midrule

\multirow{2}{*}{Claude}
& A & \sigcell 3.99 (0.00 | 0.00) & \sigcell 17.10 (0.00 | 0.00) & \sigcell 14.28 (0.00 | 0.00) \\
& M & \sigcell 2.85 (0.00 | 0.01) & \sigcell 16.04 (0.00 | 0.00) & \sigcell 14.00 (0.00 | 0.00) \\

\midrule

\multirow{2}{*}{Gemini}
& A & \sigcell 3.63 (0.00 | 0.00) & \sigcell 11.39 (0.00 | 0.00) & \sigcell 8.60 (0.00 | 0.00) \\
& M & 2.22 (0.03 | 0.08) & \sigcell 10.32 (0.00 | 0.00) & \sigcell 8.61 (0.00 | 0.00) \\

\midrule

\multirow{2}{*}{GPT-OSS-20b}
& A & \sigcell 2.53 (0.01 | 0.03) & \sigcell 3.55 (0.00 | 0.00) & 1.08 (0.28 | 0.85) \\
& M & 1.87 (0.06 | 0.19) & \sigcell 2.81 (0.01 | 0.01) & 0.97 (0.33 | 0.99) \\

\midrule

\multirow{2}{*}{Gemma-12b}
& A & \sigcell 3.50 (0.00 | 0.00) & \sigcell 3.71 (0.00 | 0.00) & 0.23 (0.82 | 1.00) \\
& M & 2.29 (0.02 | 0.07) & \sigcell 2.56 (0.01 | 0.03) & 0.27 (0.78 | 1.00) \\

\midrule

\multirow{2}{*}{Olmo-7b}
& A & \sigcell 7.07 (0.00 | 0.00) & \sigcell 13.21 (0.00 | 0.00) & \sigcell 7.79 (0.00 | 0.00) \\
& M & \sigcell 6.47 (0.00 | 0.00) & \sigcell 12.81 (0.00 | 0.00) & \sigcell 7.85 (0.00 | 0.00) \\

\midrule

\multirow{2}{*}{Gemma-4b}
& A & \sigcell 3.92 (0.00 | 0.00) & \sigcell 4.11 (0.00 | 0.00) & 0.22 (0.83 | 1.00) \\
& M & 2.14 (0.03 | 0.10) & 2.41 (0.02 | 0.05) & 0.29 (0.77 | 1.00) \\

\bottomrule
\end{tabular}
\end{table}

\yashar{The non-LLM baselines help contextualize the trade-offs. \textsc{MILP} achieves the highest grounded success ($0.73$), outperforming all \sysname{} variants on the moderator metric. This outcome is expected because \textsc{MILP} formulates recommendations as a closed-form optimization problem over fully structured catalog inputs, predefined constraints, and a hand-specified utility function. In contrast, \sysname{} operates directly on natural-language queries and uses iterative multi-agent refinement with deterministic catalog validation to generate and repair recommendations. \change{Despite this harder setting, \sysname{} achieves $\sim$90\% of \textsc{MILP}'s grounded success (0.66 vs.\ 0.73) while also reducing recommendation concentration and improving diversity (\autoref{section: RQ2}).} We therefore treat \textsc{MILP} as a structured-input upper bound on grounded success rather than a direct competitor in a fully conversational recommendation setting.}

\yashar{\sysname{}'s complementary advantages lie in its ability to handle open-ended natural-language requests, iteratively repair constraint violations, expose transparent per-agent diagnostics, and support modular stakeholder objectives without retraining. By contrast, \textsc{TopPop} attains relatively high grounded success ($0.676$) largely by repeatedly recommending the most popular cities and ignoring personalization, while \textsc{RandRec} performs substantially worse ($0.501$), underscoring the difficulty of satisfying multi-constraint queries without reasoning or grounding.}

We additionally observe robustness limitations in the \textsc{SASI} pipeline for \claude{} and \gptOss{}, which \change{fail} to produce valid outputs for $\sim$157/900~\change{and}~$\sim$17/900 queries respectively, further motivating multi-agent moderation.

\begin{figure*}[htbp]
    \centering

    \begin{subfigure}{0.32\textwidth}
        \includegraphics[width=\linewidth]{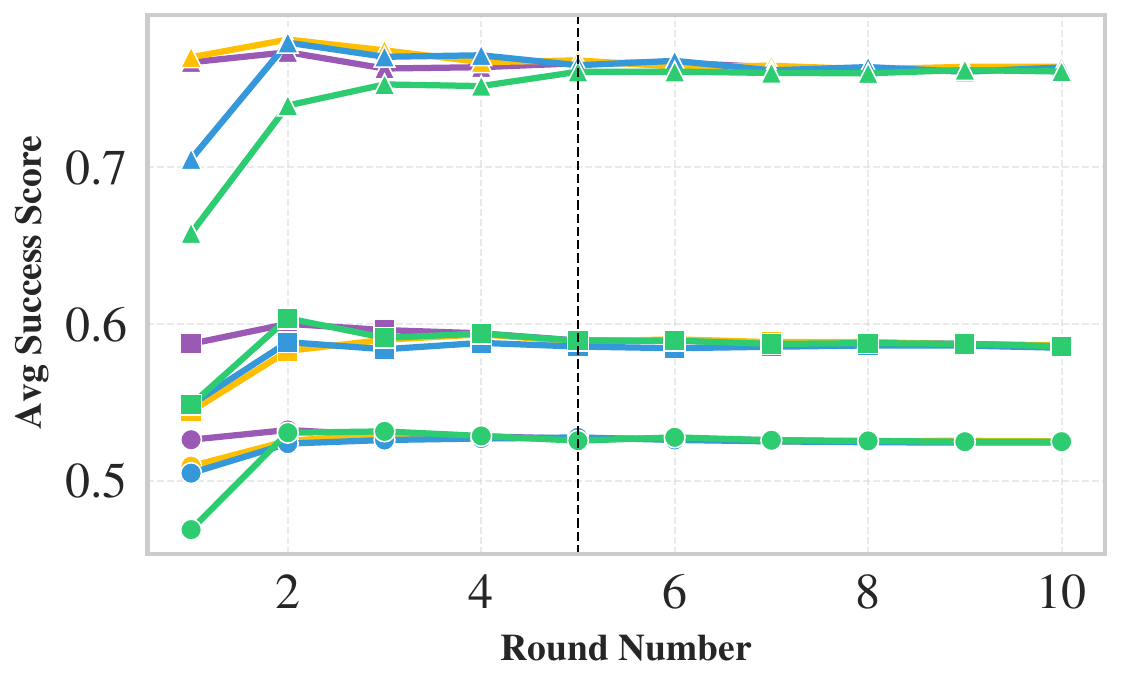}
        \caption{Claude}
    \end{subfigure}
    \begin{subfigure}{0.32\textwidth}
        \includegraphics[width=\linewidth]{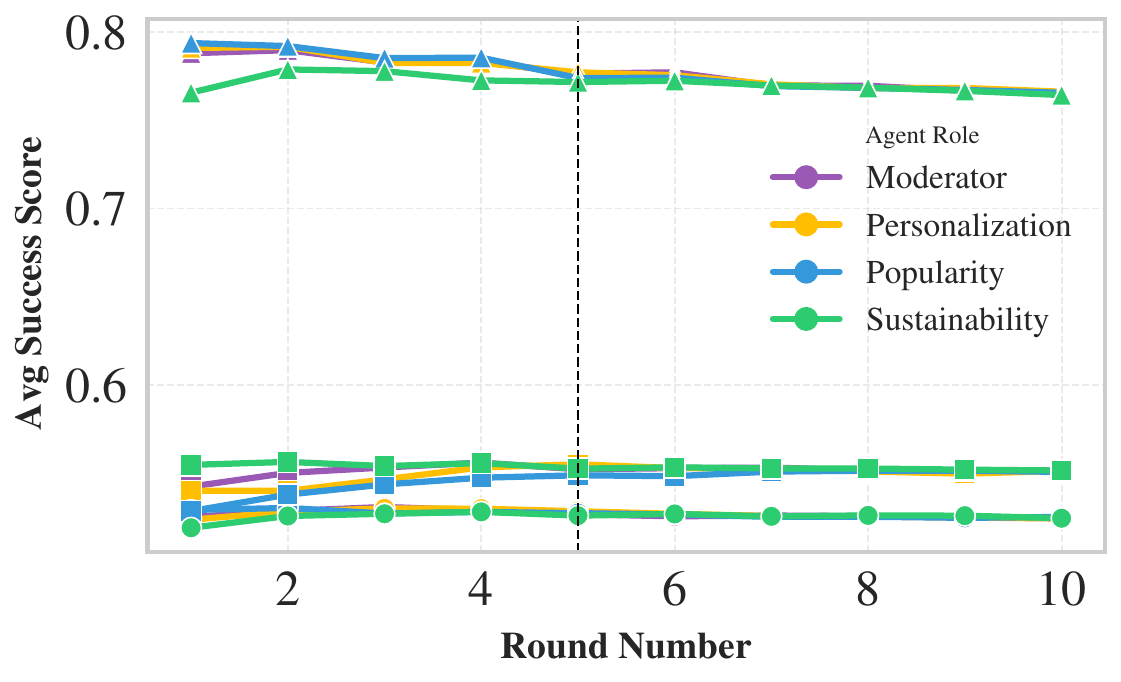}
        \caption{Gemini}
    \end{subfigure}
    \begin{subfigure}{0.32\textwidth}
        \includegraphics[width=\linewidth]{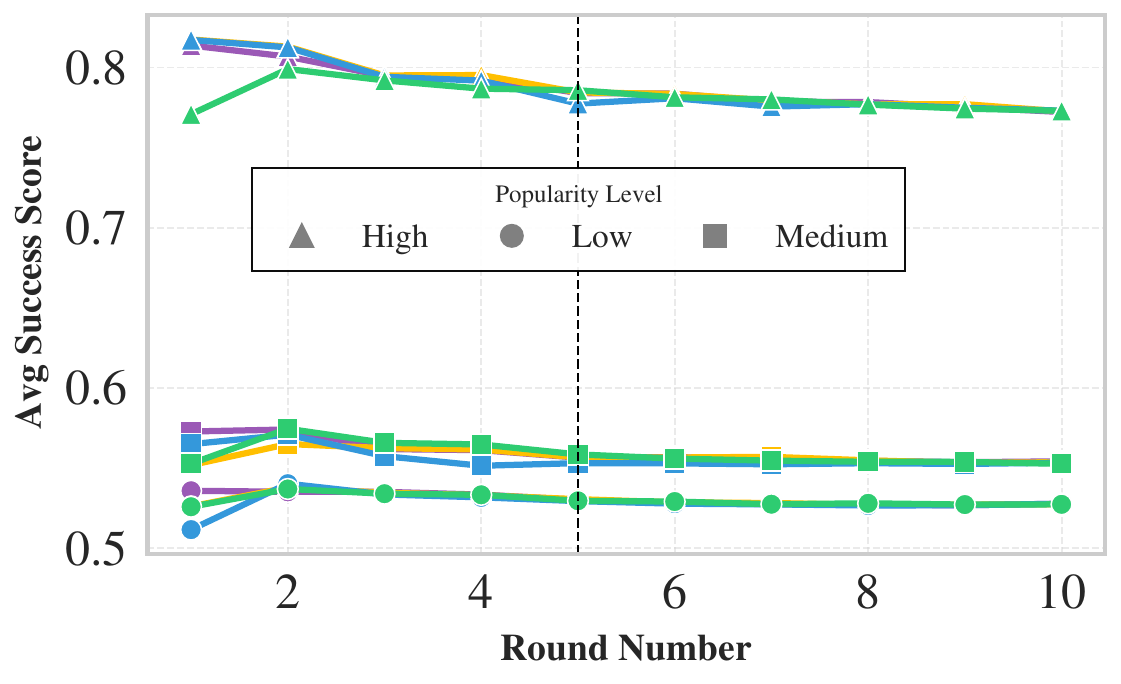}
        \caption{GPT-OSS-20B}
    \end{subfigure}

    \vspace{0.5em}

    \begin{subfigure}{0.32\textwidth}
        \includegraphics[width=\linewidth]{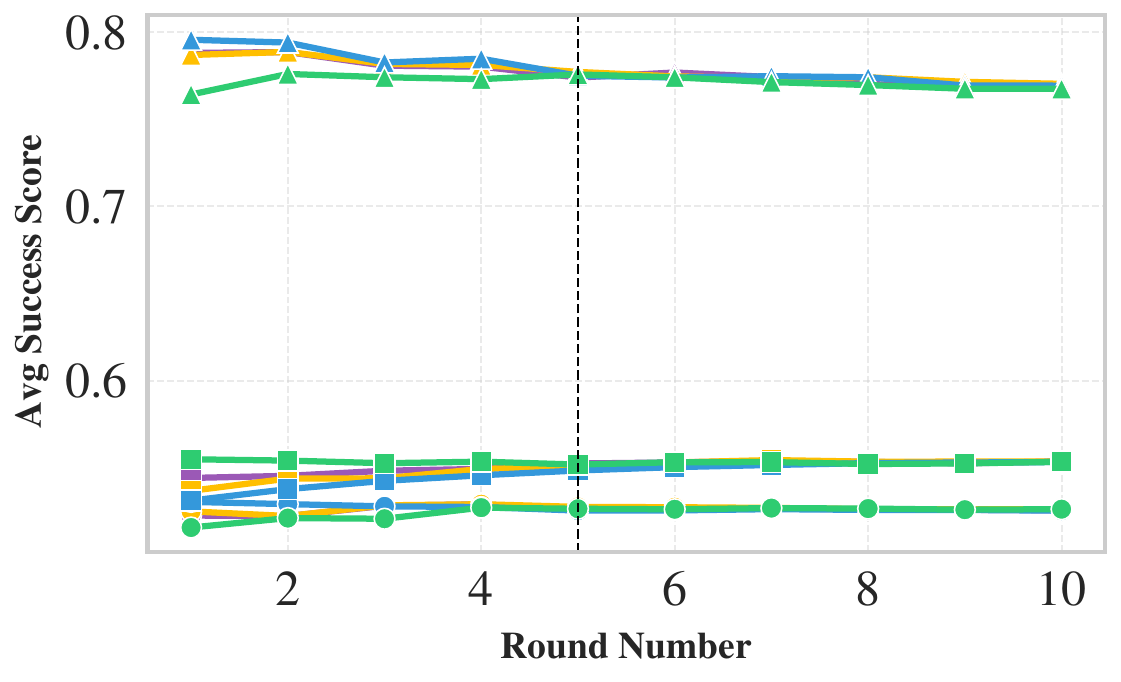}
        \caption{Gemma-12b}
    \end{subfigure}
    \begin{subfigure}{0.32\textwidth}
        \includegraphics[width=\linewidth]{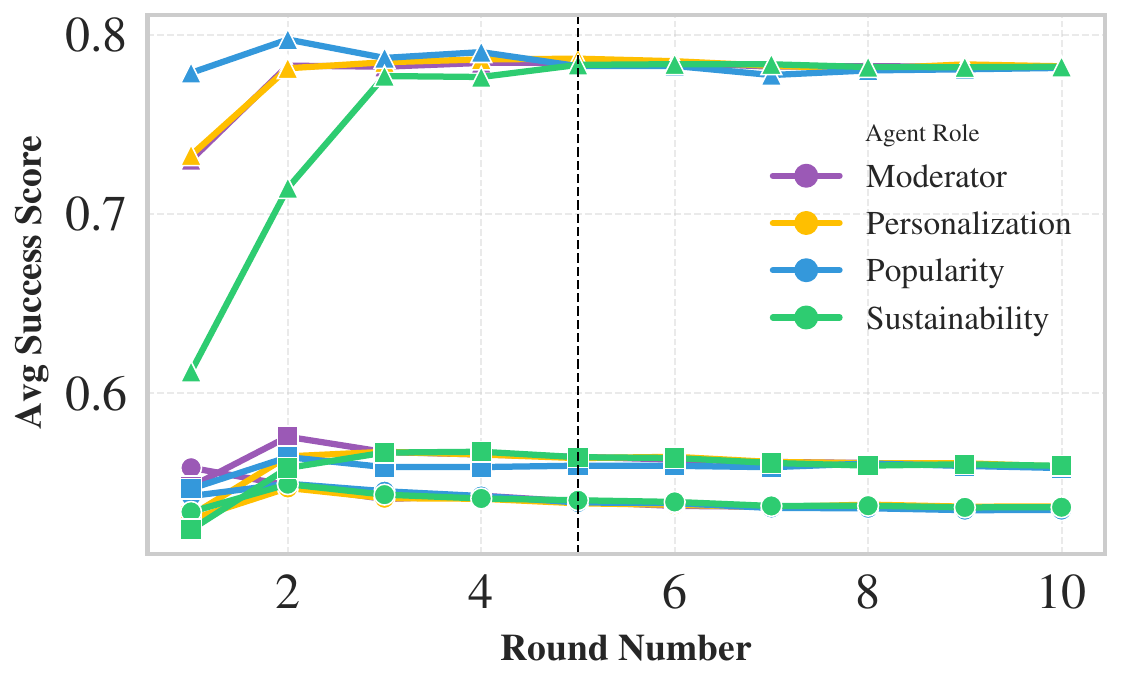}
        \caption{Olmo-7b}
    \end{subfigure}
    \begin{subfigure}{0.32\textwidth}
        \includegraphics[width=\linewidth]{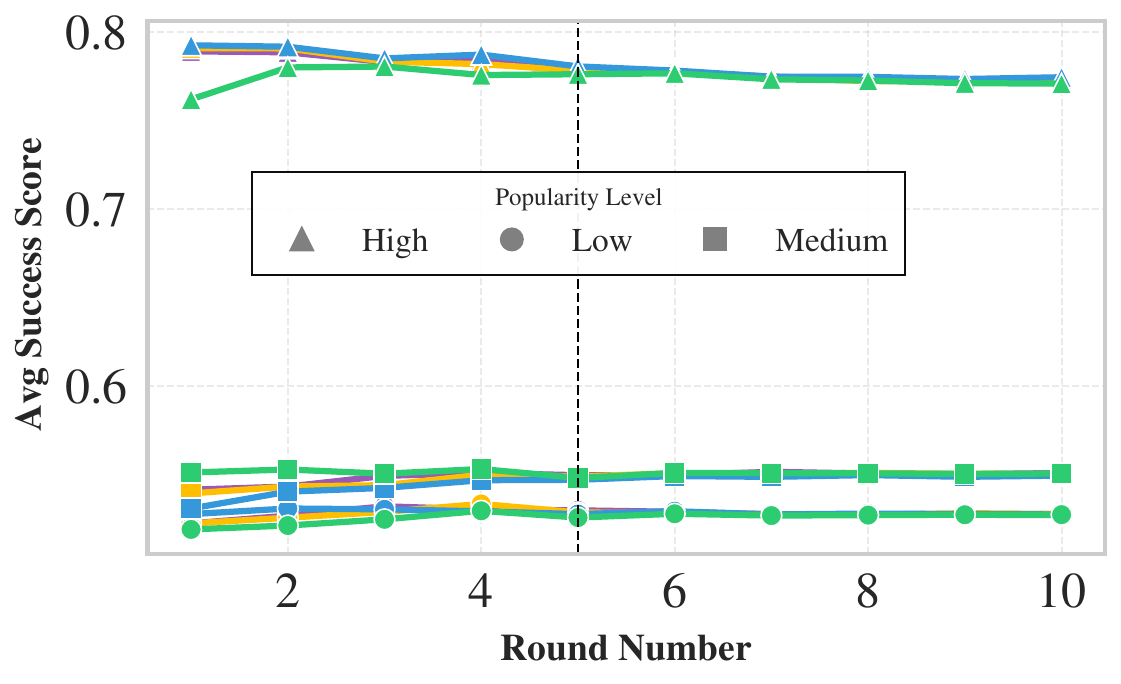}
        \caption{Gemma-4b}
    \end{subfigure}

    \caption{Average agent success scores over refinement rounds under the \textit{Aggressive} rejection strategy. The plots track performance for the \textcolor{personalization}{\textit{Personalization}}, \textcolor{popularity}{\textit{Popularity}}, \textcolor{sustainability}{\textit{Sustainability}}, and \textcolor{moderator}{\textit{Moderator}} agents across LLM backbones. Results are stratified by query popularity level (low, medium, high).
    The dotted black line denotes the convergence plateau \change{typically reached by rounds 4--5}. This stabilization validates the patience-based early stopping protocol, as relevance gains generally diminish thereafter.
    }
    \label{fig: relevance_plots_aggressive}
\end{figure*}

\paragraph{Early stopping and convergence dynamics.}
A central practical question is whether multi-round gains require running the full 10-round budget.
\autoref{tab: relevance_results} reports that \textsc{MAMI} typically converges by $\sim$4 rounds (average convergence 3.72--4.01). \change{As shown in \autoref{fig: relevance_plots_aggressive}, which plots average agent success scores over refinement rounds,} success rises quickly in the first 3--4 rounds and then reaches a plateau around rounds 4--5.
This pattern is consistent across models and indicates diminishing returns for longer runs.
\tors{Extended runs up to 20 rounds (\gemmaFour{}, \claude{}, \olmo{}; see \autoref{fig: relevance_plots_20_rounds} in~\autoref{appendix: relevance results}) confirm convergence within 3–4 rounds, supporting our 10-round cap.}

Importantly, early stopping does \emph{not} sacrifice quality: across models, \MAMIearly{} is often the best or near-best variant.
Under \textit{Aggressive} rejection, \MAMIearly{} is higher than \MAMIfull{} for every model in~\autoref{tab: relevance_results} \change{(e.g., \claude{} 0.657 vs.\ 0.620, \gemini{} 0.648 vs.\ 0.617)}, indicating that later rounds can add churn without improving grounded satisfaction once the system has already
settled into a feasible region.

\textit{Statistical Significance.}
\autoref{tab: statistical_significance} reports paired comparisons using Bonferroni correction (\change{$\alpha_{\mathrm{corr}} = 0.05/3 \approx 0.017$}) over query-level distributions.
The strongest and most consistent effect is \textsc{MAMI} vs.\ \textsc{SASI}, which is significant across nearly all models and strategies.
\change{The sole borderline case is \gemmaFour{} under \textit{Majority} rejection, where \MAMIearly{} vs.\ \textsc{SASI} lands exactly at the correction threshold ($p_{\mathrm{corr}} = 0.05$), leaving this comparison inconclusive under our strict criterion.}
\change{\textsc{MAMI} vs.\ \textsc{MASI} is also significant for most settings, with exceptions under \textit{Majority} rejection for \gemini{}, \gptOss{}, \gemmaTwelve{}, and \gemmaFour{} --- where the comparison does not reach $\alpha_{\mathrm{corr}}$ --- indicating that gains over \textsc{MASI} are most reliable when the moderator applies the stricter \textit{Aggressive} rejection strategy.}

\vspace{2.0mm}
\RQBox{RQ1 Summary.}{Multi-round, multi-agent \change{refinement} improves \change{moderator success} over single-agent and single-round
baselines across all evaluated LLM families. Quality improves rapidly in the first 3--4 rounds and typically plateaus by rounds 4--5.
Patience-based early stopping, therefore, captures most gains while avoiding unnecessary computation.}

\subsection{RQ2: Popularity Bias and Diversification}
\label{section: RQ2}

\begin{figure*}[htbp]
    \centering

    \begin{subfigure}{0.32\textwidth}
        \includegraphics[width=\linewidth]{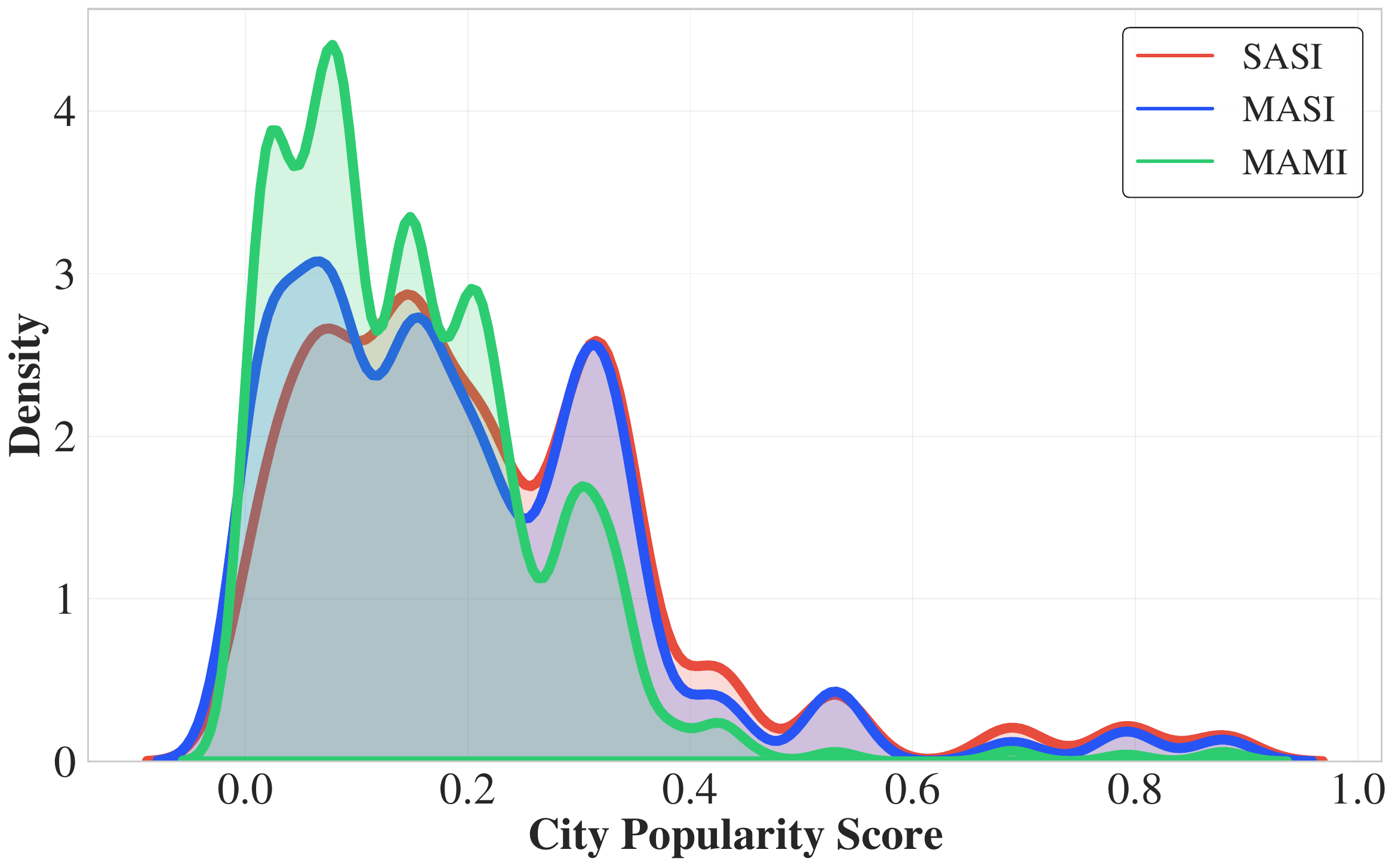}
        \caption{Claude}
    \end{subfigure}
    \begin{subfigure}{0.32\textwidth}
        \includegraphics[width=\linewidth]{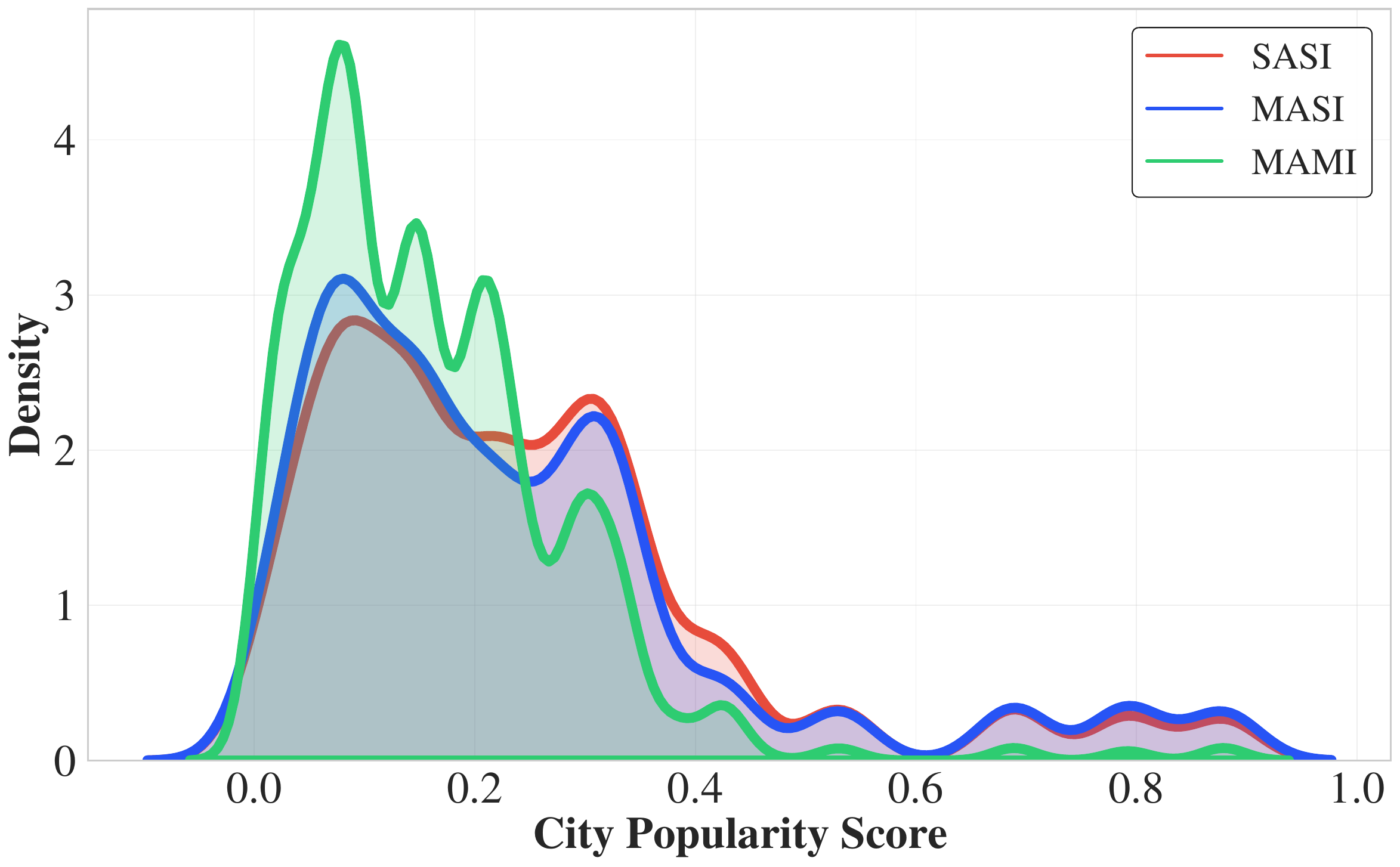}
        \caption{Gemini}
    \end{subfigure}
    \begin{subfigure}{0.32\textwidth}
        \includegraphics[width=\linewidth]{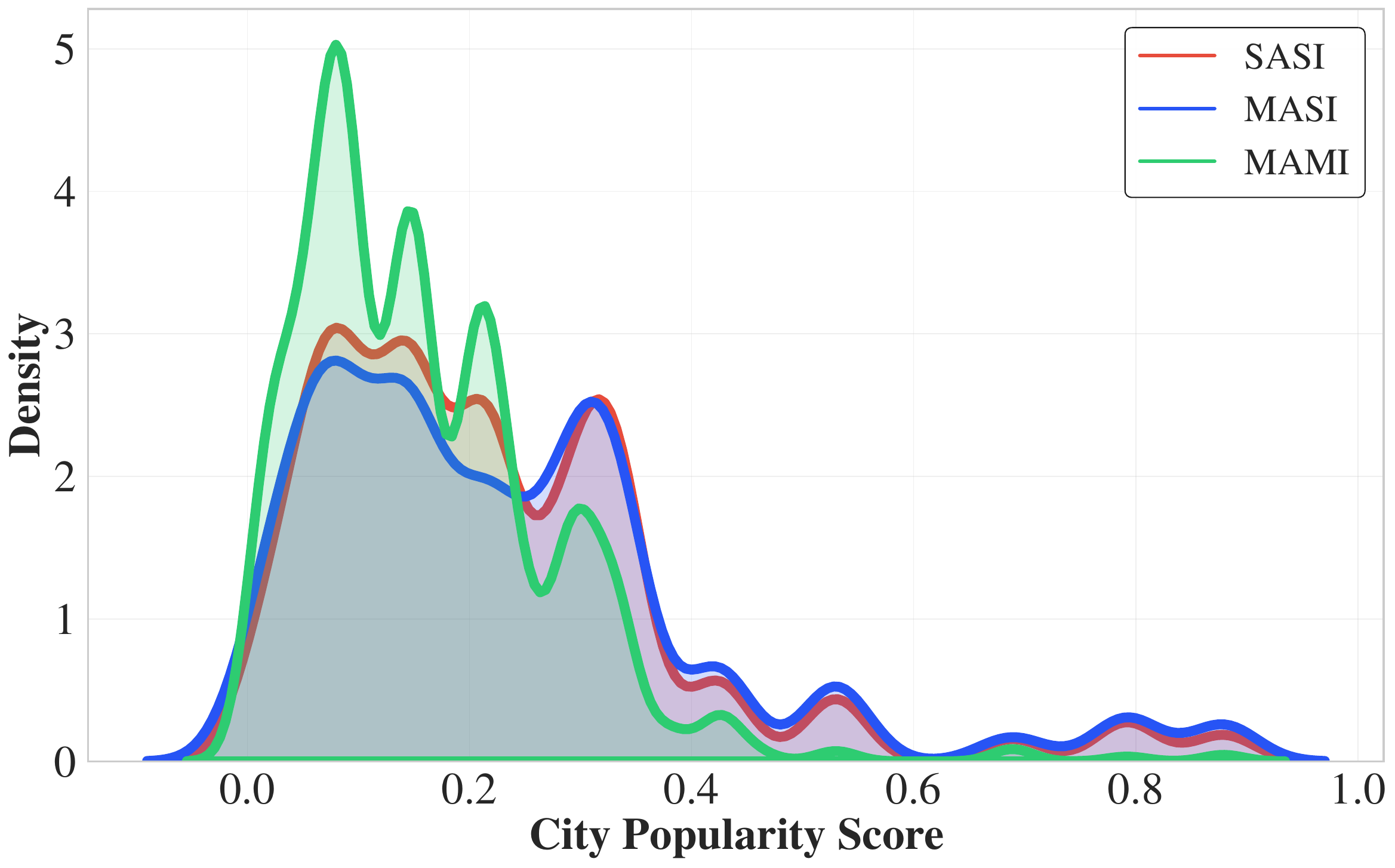}
        \caption{GPT-OSS-20B}
    \end{subfigure}

    \vspace{0.5em}

    \begin{subfigure}{0.32\textwidth}
        \includegraphics[width=\linewidth]{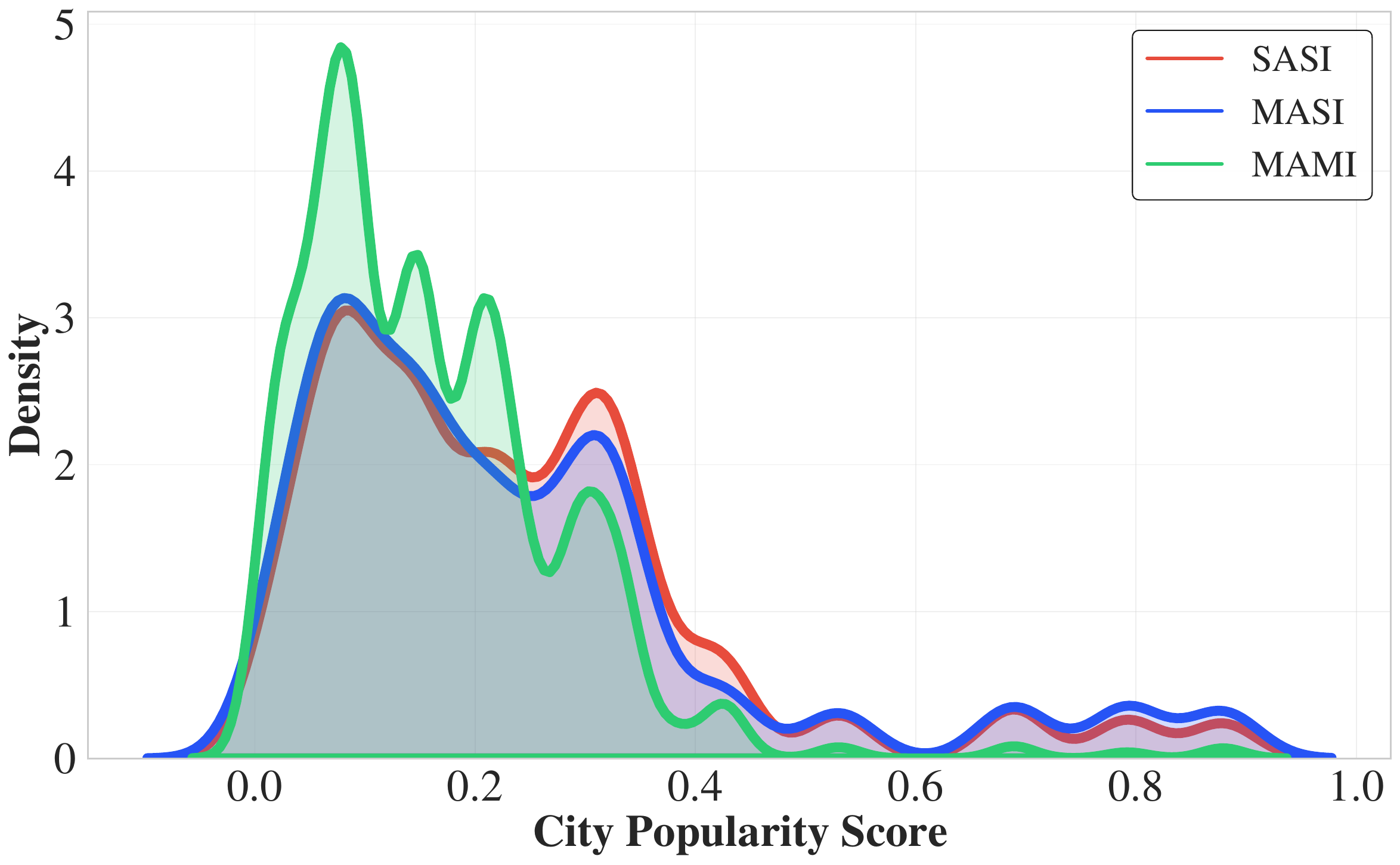}
        \caption{Gemma-12b}
    \end{subfigure}
    \begin{subfigure}{0.32\textwidth}
        \includegraphics[width=\linewidth]{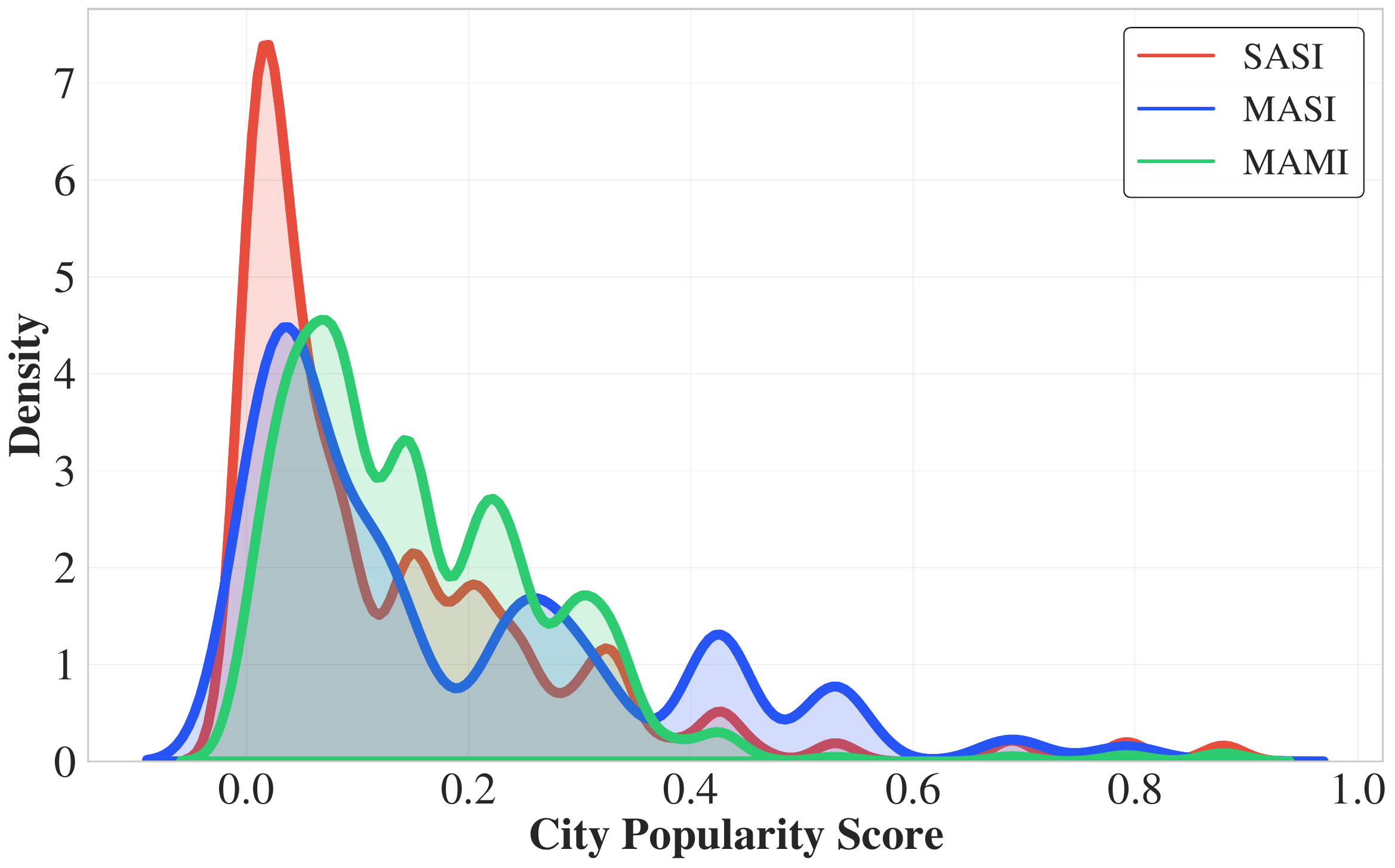}
        \caption{Olmo-7b}
    \end{subfigure}
    \begin{subfigure}{0.32\textwidth}
        \includegraphics[width=\linewidth]{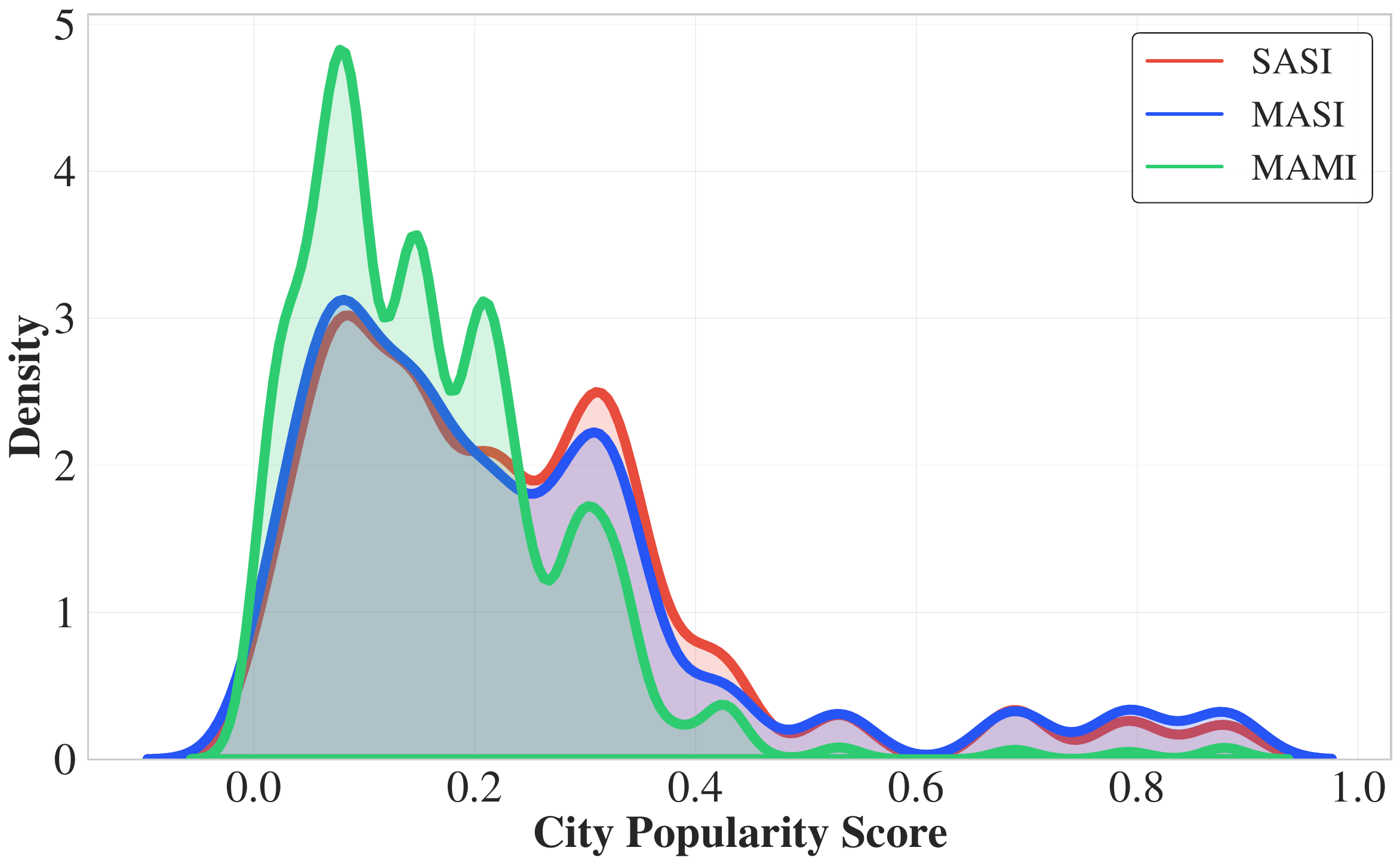}
        \caption{Gemma-4b}
    \end{subfigure}

    \caption{Kernel Density Estimation (KDE) of city popularity distributions across methods and models for the aggressive rejection strategy. The filled regions illustrate the probability density of recommended cities according to their popularity scores.
    While \textcolor{SASI}{\textsc{SASI}} and \textcolor{MASI}{\textsc{MASI}} show high-amplitude peaks concentrated on popular hubs, \textcolor{MAMI}{\textsc{MAMI}} (Round 10) displays a flatter, broader curve. This visual shift indicates a significant reduction in popularity bias and an increased coverage of lesser-known \change{``long-tail''} destinations.
    }
    \label{fig: kde_diversity_aggressive}
\end{figure*}

\begin{figure*}[htbp]
    \centering

    \begin{subfigure}{0.32\textwidth}
        \includegraphics[width=\linewidth]{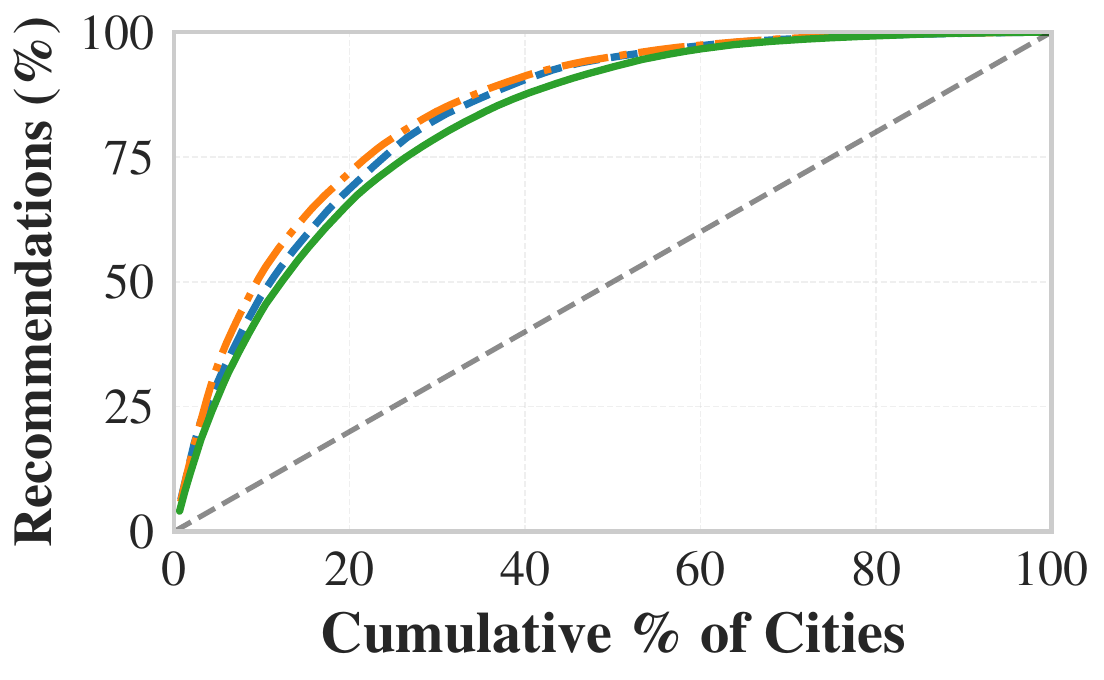}
        \caption{Claude}
    \end{subfigure}
    \begin{subfigure}{0.32\textwidth}
        \includegraphics[width=\linewidth]{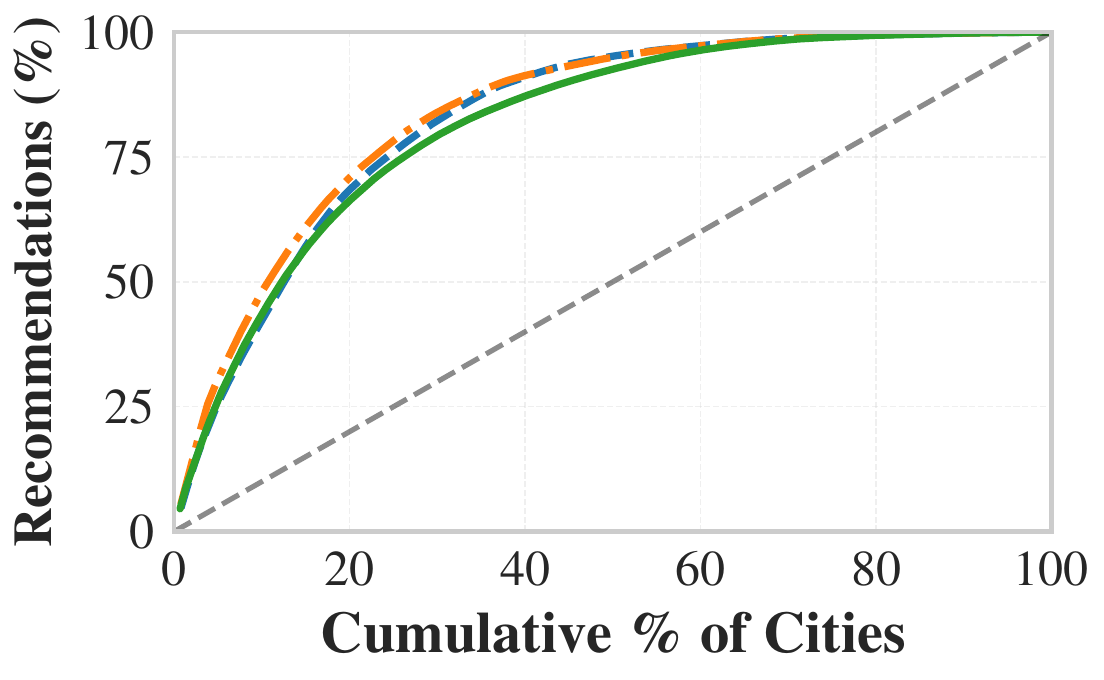}
        \caption{Gemini}
    \end{subfigure}
    \begin{subfigure}{0.32\textwidth}
        \includegraphics[width=\linewidth]{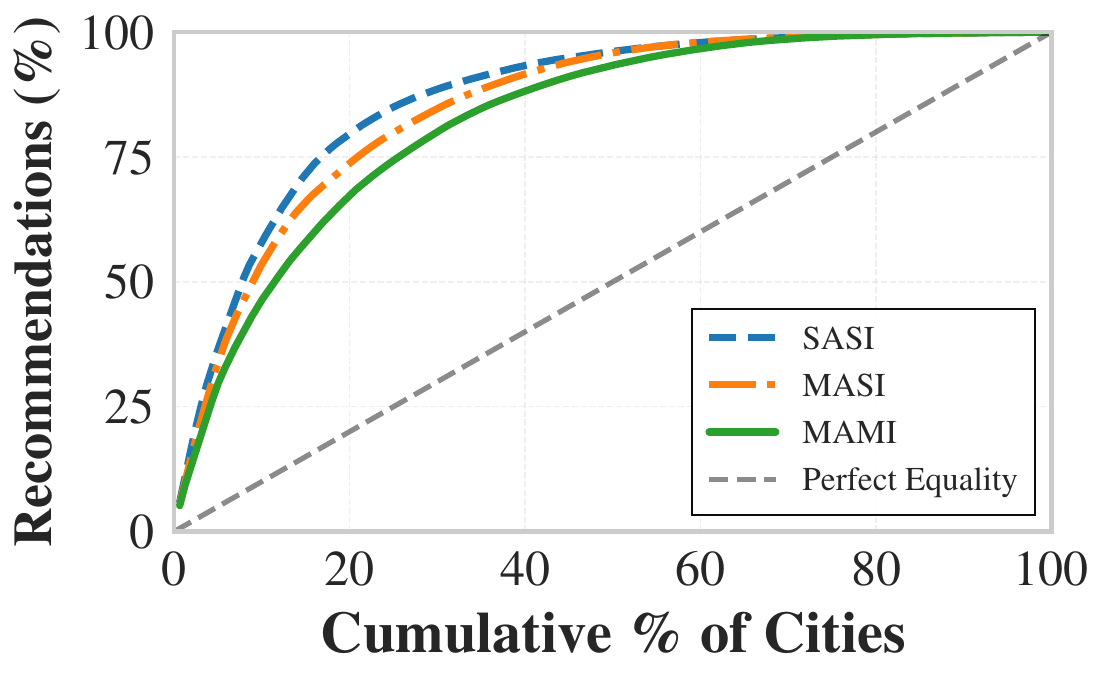}
        \caption{GPT-OSS-20B}
    \end{subfigure}

    \vspace{0.5em}

    \begin{subfigure}{0.32\textwidth}
        \includegraphics[width=\linewidth]{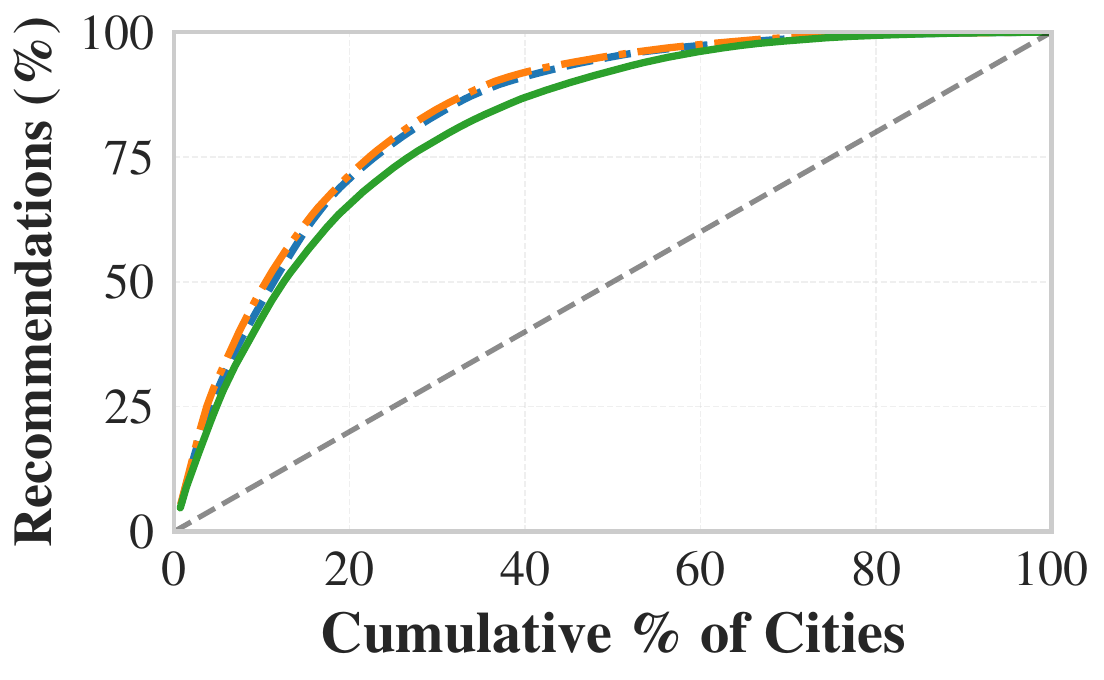}
        \caption{Gemma-12b}
    \end{subfigure}
    \begin{subfigure}{0.32\textwidth}
        \includegraphics[width=\linewidth]{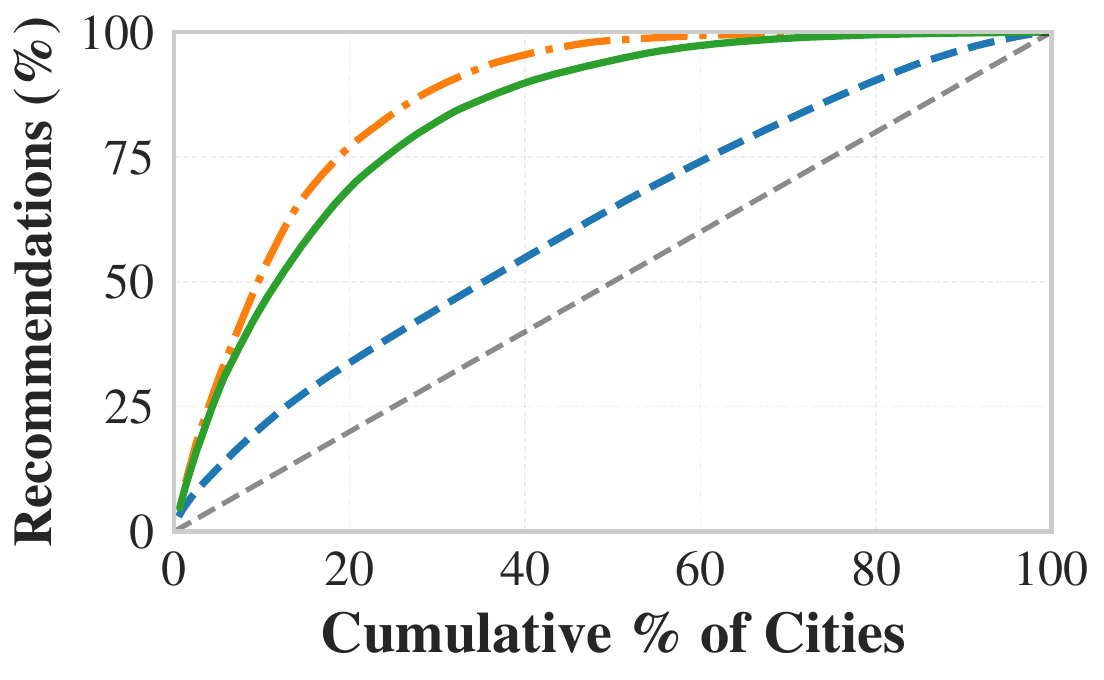}
        \caption{Olmo-7b}
    \end{subfigure}
    \begin{subfigure}{0.32\textwidth}
        \includegraphics[width=\linewidth]{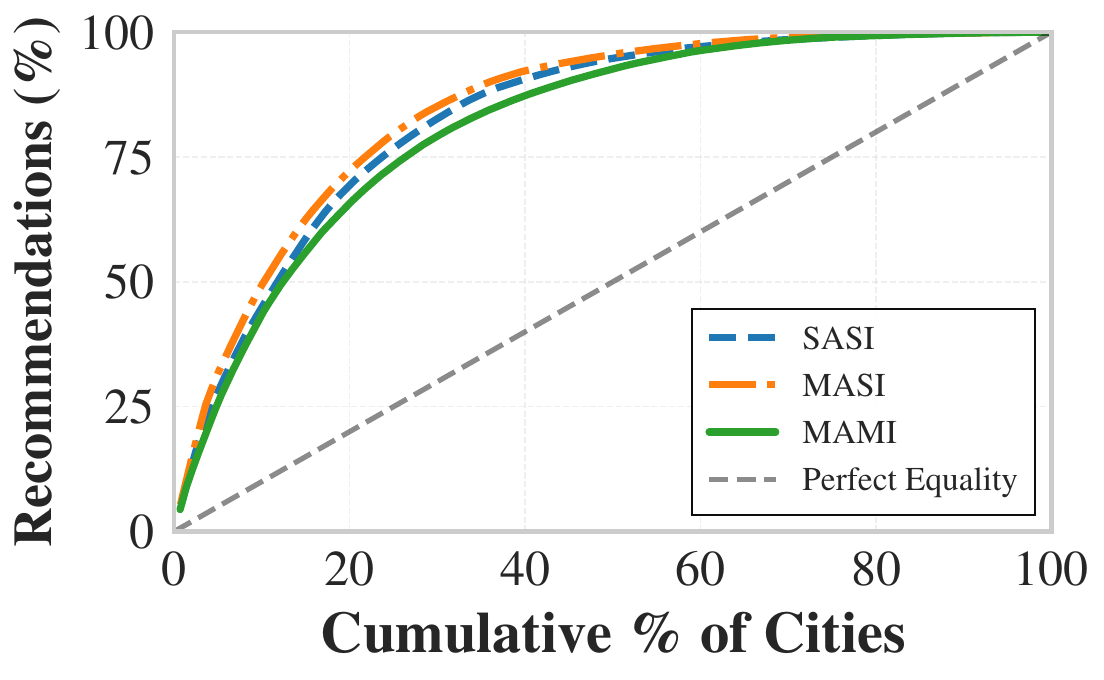}
        \caption{Gemma-4b}
    \end{subfigure}

    \caption{\tors{Lorenz curves showing recommendation concentration across the 200-city catalog. The x-axis represents the cumulative percentage of cities, and the y-axis represents the cumulative percentage of recommendations. The diagonal ($y=x$) indicates perfect equality. Curves that bow further below the diagonal indicate higher concentration, with a few \change{``short-head''} cities dominating recommendations. MAMI/\MAMIfull{} (solid) consistently bows less than SASI (dashed) and MASI/\MASI{} (dot-dashed), indicating reduced popularity bias and a more equitable, long-tail distribution of recommended destinations.}
    }
    \label{fig: lorenz_aggressive}
\end{figure*}

\noindent\textbf{RQ2 asks:} \emph{Does multi-round \torsRTwo{refinement} reduce popularity concentration and increase long-tail coverage?}
We evaluate popularity bias using (i) distributional shifts in recommended popularity scores (KDE in~\autoref{fig: kde_diversity_aggressive}), (ii) concentration in the
catalog (Lorenz curves in~\autoref{fig: lorenz_aggressive}), and (iii) summary metrics: Gini and entropy (lower/higher is better) as well as catalog coverage
(\autoref{tab: diversity}).

\paragraph{Distributional shift toward the long tail.}
Across models, \textcolor{MAMI}{\textsc{MAMI}} systematically shifts recommendation mass away from the highest-popularity range and toward the mid/long tail
(\autoref{fig: kde_diversity_aggressive}).
This effect is strongest under \textit{Aggressive} rejection, where the moderator more readily discards repeated high-traffic suggestions and
forces agents to explore alternatives.
In contrast, \textcolor{SASI}{\textsc{SASI}} and \textcolor{MASI}{\textsc{MASI}} frequently exhibit sharper peaks in the high-popularity region, reflecting the well-known
tendency of LLMs to default to canonical destinations.

\begin{table}[htbp]
\centering
\resizebox{\textwidth}{!}{
\begin{tabular}{llcccccccc}
\toprule
\multirow{2}{*}{\textbf{Model}} &
\multicolumn{1}{l}{\multirow{2}{*}{\textbf{\begin{tabular}[c]{@{}l@{}}Rej.\\ Strat.\end{tabular}}}}&
\multicolumn{4}{c}{\textbf{Gini \minSign{} (Entropy \maxSign{})}} &
\multicolumn{4}{c}{\textbf{Coverage (\%) (Avg \#recs/city) \maxSign{}}} \\
\cmidrule(lr){3-6} \cmidrule(lr){7-10}
& &
\textbf{SASI} &
\textbf{\MASI} &
\textbf{\MAMIearly} &
\textbf{\MAMIfull} &
\textbf{SASI} &
\textbf{\MASI} &
\textbf{\MAMIearly} &
\textbf{\MAMIfull} \\
\midrule

\multirow{2}{*}{Claude}
& A & \multirow{2}{*}{0.71 (0.82)} & \multirow{2}{*}{0.69 (0.82)} & 0.65 (0.85) & \maxHighlight{0.64} (0.86)
& \multirow{2}{*}{66.0 (49.9)} & 67.0 (67.2) & \maxHighlight{81.5} (55.2) & \maxHighlight{81.5} (55.2) \\
& M &  & & 0.67 (0.84) & \maxHighlight{0.67} (0.84)
& & 69.0 (65.2) & 75.0 (60.0) & \maxHighlight{77.0} (58.4) \\
\midrule

\multirow{2}{*}{Gemini}
& A & \multirow{2}{*}{0.66 (0.84)} & \multirow{2}{*}{0.68 (0.83)} & \maxHighlight{0.63} (0.86) & 0.64 (0.86)
& \multirow{2}{*}{64.5 (57.9)} & 65.5 (68.6) & 73.5 (61.2) & \maxHighlight{74.5} (60.3) \\
& M &  & & 0.67 (0.84) & \maxHighlight{0.66} (0.84)
& & 66.5 (67.7) & \maxHighlight{73.0} (61.8) & \maxHighlight{73.0} (61.6) \\
\midrule

\multirow{2}{*}{GPT-OSS-20b}
& A & \multirow{2}{*}{0.76 (0.78)} & \multirow{2}{*}{0.71 (0.81)} & 0.65 (0.85) & \maxHighlight{0.65} (0.85)
& \multirow{2}{*}{81.5 (53.4)} & 76.0 (59.2) & 80.0 (56.2) & 79.5 (56.6) \\
& M &  & & 0.68 (0.84) & \maxHighlight{0.67} (0.84)
& & 76.5 (58.8) & 83.0 (54.2) & \maxHighlight{85.0} (52.9) \\
\midrule

\multirow{2}{*}{Gemma-12b}
& A & \multirow{2}{*}{0.69 (0.82)} & \multirow{2}{*}{0.69 (0.82)} & 0.64 (0.86) & \maxHighlight{0.63} (0.86)
& \multirow{2}{*}{63.5 (70.5)} & 67.0 (67.2) & 71.5 (62.9) & \maxHighlight{72.0} (62.5) \\
& M &  & & 0.67 (0.84) & \maxHighlight{0.64} (0.85)
& & 64.5 (69.8) & 68.0 (66.2) & \maxHighlight{69.0} (65.2) \\
\midrule

\multirow{2}{*}{Olmo-7b}
& A & \multirow{2}{*}{0.67 (0.84)} & \multirow{2}{*}{0.73 (0.77)} & 0.67 (0.84) & \maxHighlight{0.66} (0.84)
& \multirow{2}{*}{\maxHighlight{100} (242.5)} & 38.5 (116.9) & 77.0 (58.4) & 79.5 (56.6) \\
& M &  & & 0.70 (0.82) & \maxHighlight{0.68} (0.84)
& & 39.0 (115.4) & 77.0 (58.4) & 79.0 (57.0) \\
\midrule

\multirow{2}{*}{Gemma-4b}
& A & \multirow{2}{*}{0.68 (0.83)} & \multirow{2}{*}{0.70 (0.82)} & 0.64 (0.86) & \maxHighlight{0.63} (0.86)
& \multirow{2}{*}{61.0 (73.3)} & 69.5 (64.7) & 70.0 (64.3) & \maxHighlight{74.0} (60.8) \\
& M &  & & 0.66 (0.84) & \maxHighlight{0.65} (0.85)
& & 66.0 (68.2) & 71.0 (63.4) & \maxHighlight{71.5} (62.9) \\

\bottomrule
\end{tabular}}
\vspace{1mm}
\footnotesize{\textbf{Note:} Non-LLM baselines: Gini (Entropy): \textbf{RR} = 0.08 (0.99), \textbf{TP} = 0.95 (0.43), \torsRTwo{\textsc{MILP} = 0.78 (0.77)}. Coverage: \textbf{RR} = 99.5\%, \torsRThree{\textbf{TP} = 5\%}, \torsRTwo{\textsc{MILP} = 48.0\%.}}
\vspace{2mm}
\caption{\textbf{Gini (Entropy)} and \textbf{Coverage metrics} across models and rejection strategies.
The table reports concentration bias and long-tail coverage for each LLM model under \textit{Aggressive} \textbf{(A)} and \textit{Majority} \textbf{(M)} rejection strategies. Lower Gini indicates reduced concentration bias, and higher entropy reflects a more equitable recommendation distribution. \change{\textbf{Coverage (\%)} is the fraction of the 200-city catalog appearing at least once in validated recommendations; \textbf{Avg \#recs/city} is the number of valid city appearances divided by the number of unique catalog cities covered.} Row-wise maximum coverage is highlighted using \maxHighlight{bold}. MAMI variants consistently achieve higher coverage and more balanced distributions than SASI and MASI baselines, except in \olmo{}. \change{All counts are computed after parsing, catalog validation, removal of invalid outputs, and duplicate handling; for methods with failed or invalid outputs, total city appearances may therefore fall below $900 \times 10$.}}
\label{tab: diversity}
\end{table}

\change{\olmo{} is a notable exception. Its \textsc{SASI} baseline shows pronounced density in very low popularity bins and 100\% catalog coverage. \olmo{} routinely emits substantially more than $k=10$ cities per query in unmoderated mode, trivially inflating both coverage and the apparent long-tail skew in the KDE. The high Avg \#recs/city of 242.5, computed after duplicate handling as all metrics in this table, indicates that \olmo{} in unmoderated \textsc{SASI} mode emits far more than $k=10$ unique cities per query, making the 100\% coverage figure an artifact of over-generation rather than a signal of genuine diversity. \textsc{MAMI} corrects this by enforcing exactly $k$ cities per round via the moderator, so the apparent drop from 100\% to $\sim$77--79\% reflects constraint enforcement rather than a coverage regression. On the concentration metrics unaffected by output size (Gini, Lorenz curves in \autoref{fig: lorenz_aggressive}), \textsc{MAMI} yields only a marginal improvement (Gini: 0.67 $\to$ 0.66 under \MAMIfull{}), consistent with \olmo{} not exhibiting strong hub concentration to begin with.}

\paragraph{Gini, entropy, and coverage.}
\autoref{tab: diversity} provides a compact numerical view of the same phenomenon.
Across most models, multi-round \torsRTwo{refinement} reduces concentration and increases dispersion.
For example, for \gptOss{} under \textit{Aggressive} rejection, Gini decreases from 0.76 (\textsc{SASI}) to 0.65 (\MAMIearly{}/\MAMIfull{}),
while entropy increases from 0.78 to 0.85.
For \claude{} under \textit{Aggressive} rejection, Gini decreases from 0.71 to 0.64 and coverage increases from 66.0\% to 81.5\%.
For \gemini{} under \textit{Aggressive} rejection, Gini decreases from 0.66 to 0.63 and coverage increases from 64.5\% to 74.5\%.

A key trade-off emerges between relevance and diversity across rounds.
While relevance improvements tend to plateau after rounds 4--5 (RQ1:~\autoref{section: RQ1}), diversity continues to increase modestly with additional rounds: in \autoref{tab: diversity}, \MAMIearly{} yields the best diversity metrics for five of six models, whereas \MAMIfull{} is slightly preferable for Gemini.
This suggests that extra rounds can be used as a ``diversity budget'' when coverage is prioritized, albeit with additional cost (RQ4:~\autoref{section: RQ4}).

The non-LLM baselines illustrate the extremes of the relevance–diversity trade-off. \textsc{RandRec} achieves near-perfect equity (Gini 0.08, entropy 0.99, coverage 99.5\%) but produces low-quality recommendations, whereas \textsc{TopPop} is highly concentrated (Gini 0.95, entropy 0.43, coverage \torsRThree{5\%}) due to repeatedly suggesting the most popular cities\torsRThree{, yielding only 10 unique cities out of the 200-city catalog.}



\change{The \textsc{MILP} baseline occupies a middle ground in catalog concentration (Gini 0.78, entropy 0.77, coverage 48\%), achieving lower concentration than \textsc{TopPop} while still exhibiting notable popularity bias. This highlights the limitations of static scalarized objectives in capturing nuanced diversity preferences. In contrast, \sysname{} operates directly on natural-language queries and uses iterative multi-agent constrained refinement to interpret intent and repair constraint violations. Across SynthTRIPs, \sysname{} variants generally achieve broader catalog coverage and lower recommendation concentration than both single-agent/single-round baselines and the structured \textsc{MILP} reference, surfacing more lesser-visited destinations (\autoref{tab: diversity}). These results should be interpreted as controlled evidence of diversity and bias mitigation on the SynthTRIPs benchmark, rather than as proof of superiority over stronger real-world constrained-diversification or reranking approaches; validation with user studies and logged interaction data remains future work.}

\vspace{2.0mm}
\RQBox{RQ2 Summary.}{Yes. Multi-round \torsRTwo{refinement} consistently mitigates popularity concentration and increases long-tail coverage.
Diversity improves across rounds even after relevance plateaus, exposing an explicit relevance--diversity--cost trade-off: more rounds modestly improve diversity, while early stopping preserves relevance and reduces latency.}

\subsection{RQ3: Agent reliability and hallucination behavior}
\label{section: RQ3}

\noindent\textbf{RQ3 asks:} \emph{How do specialist agents behave across rounds in terms of stability (reliability) and hallucination
tendency?}
We focus on two behavioral signals: (i) \emph{reliability} (recommendation stability across rounds) and (ii) \emph{hallucinations},
operationalized as out-of-catalog or repeatedly rejected cities (\autoref{section: hallucination}).

\begin{figure*}[htbp]
    \centering

    \begin{subfigure}[t]{0.3\textwidth}
        \includegraphics[width=\textwidth]{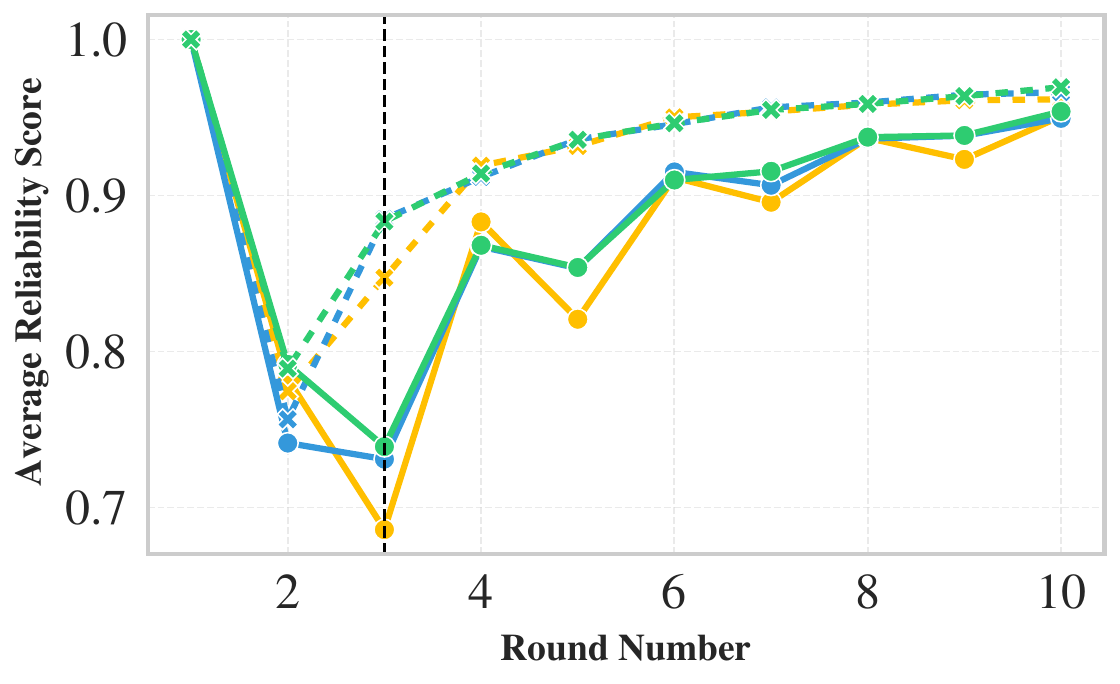}
        \caption{Claude}
    \end{subfigure}\hfill
    \begin{subfigure}[t]{0.3\textwidth}
        \includegraphics[width=\textwidth]{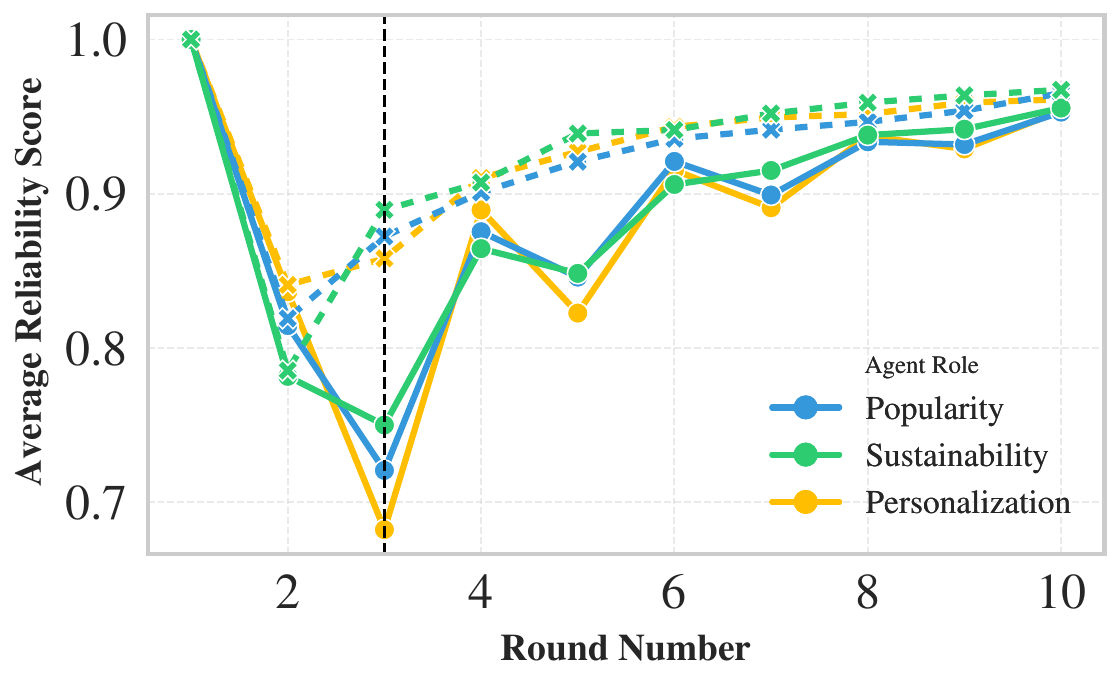}
        \caption{Gemini}
    \end{subfigure}\hfill
    \begin{subfigure}[t]{0.3\textwidth}
        \includegraphics[width=\textwidth]{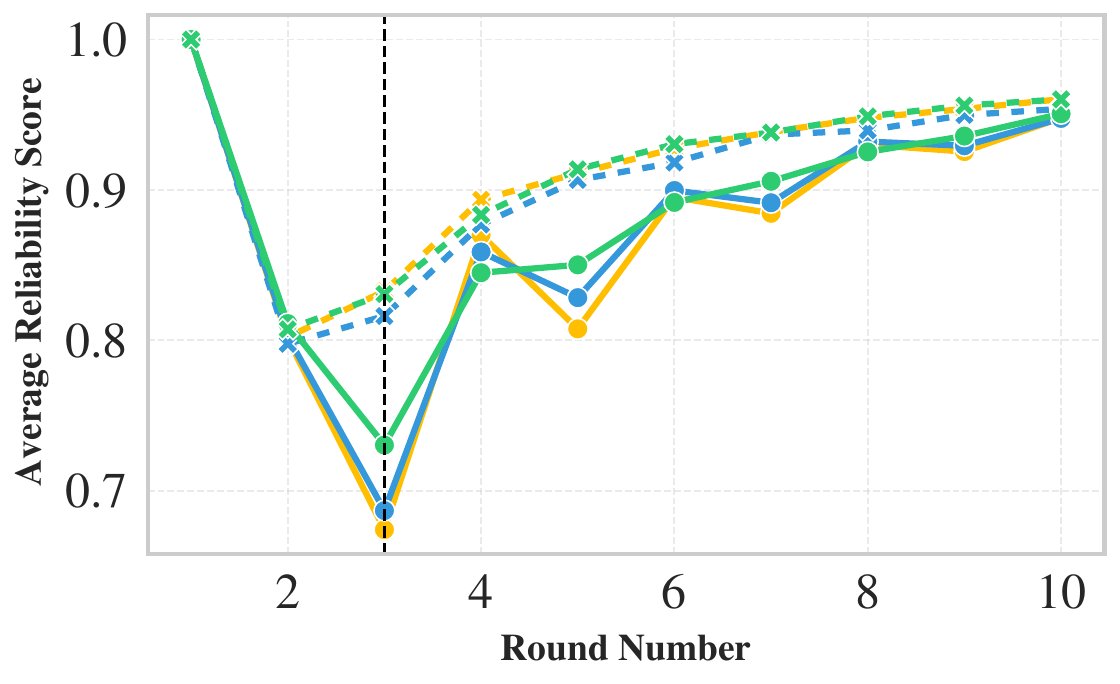}
        \caption{GPT-OSS-20B}
    \end{subfigure}

    \vspace{0.8em}

    \begin{subfigure}[t]{0.3\textwidth}
        \includegraphics[width=\textwidth]{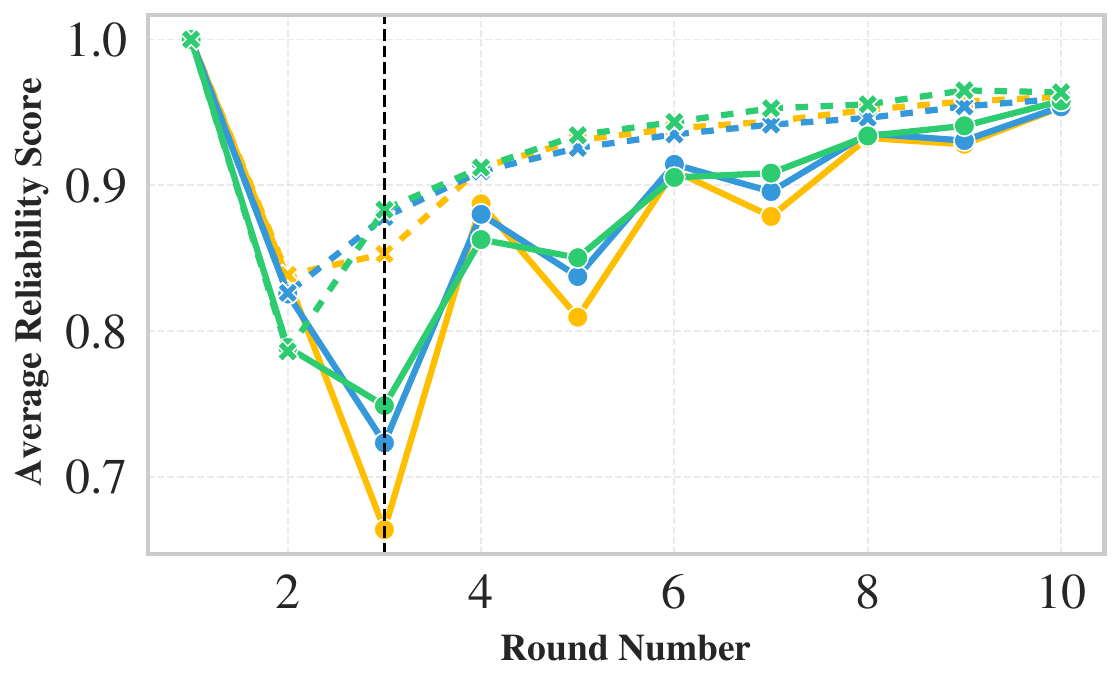}
        \caption{Gemma-12b}
    \end{subfigure}\hfill
    \begin{subfigure}[t]{0.3\textwidth}
        \includegraphics[width=\textwidth]{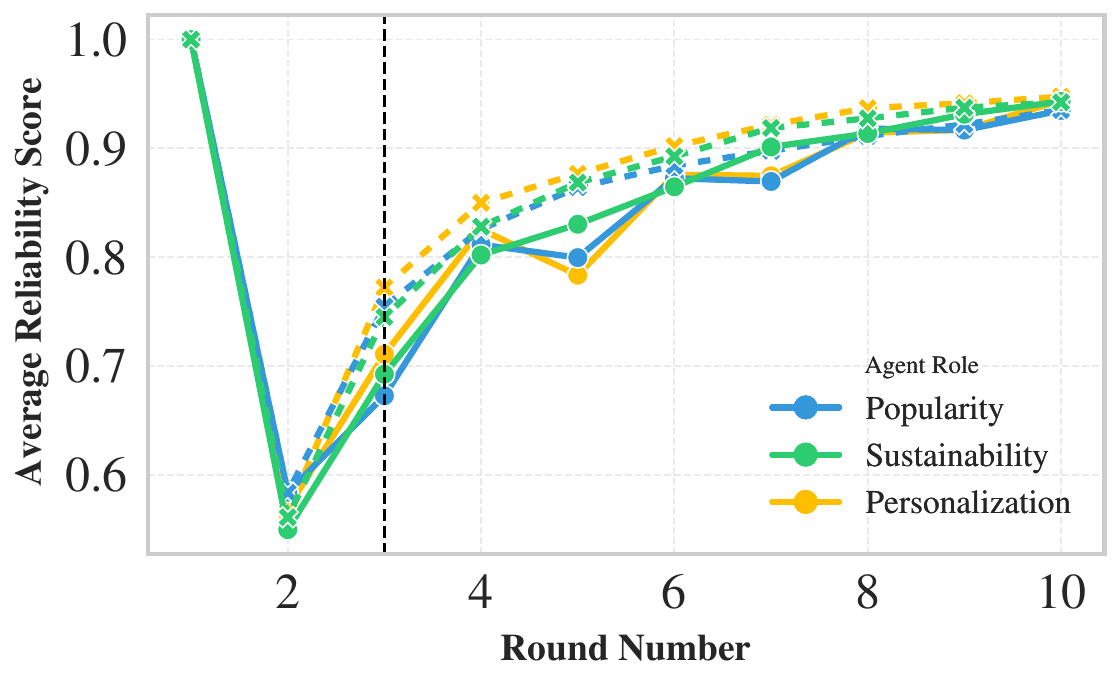}
        \caption{Olmo-7b}
    \end{subfigure}\hfill
    \begin{subfigure}[t]{0.3\textwidth}
        \includegraphics[width=\textwidth]{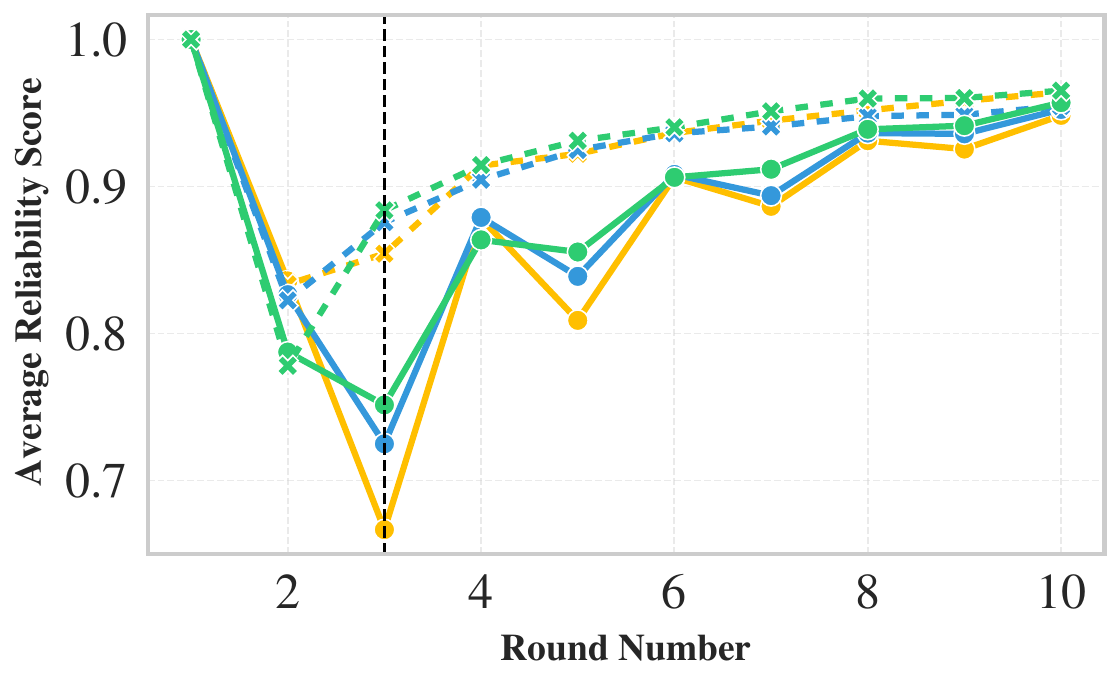}
        \caption{Gemma-4b}
    \end{subfigure}

    \caption{Agent behavior metrics showing agents' reliability scores over multiple rounds. Solid lines represent results from the \textit{Aggressive} rejection strategy, while dotted lines represent results from the \textit{Majority} voting strategy. \change{The black-dashed vertical line marks round 3 ($T_{\min}$), after which the patience criterion becomes active.} The early-round dip in reliability reflects agents adapting their proposals in response to moderator feedback; as they align on a shared candidate set, reliability scores increase and stabilize in later rounds.}
    \label{fig: agent_adaptability}
\end{figure*}

\paragraph{Reliability evolves from exploration to stabilization.}
\autoref{fig: agent_adaptability} reveals a consistent pattern across backbones.
In early rounds (typically rounds 2--3), reliability dips as agents explore the candidate space and react to moderator feedback
(i.e., they replace previously suggested cities that were rejected or penalized).
As the collective offer becomes feasible and agents align on a shared set of candidates, reliability increases steadily.
By round 10, reliability for all agents typically exceeds 0.90, indicating that the system has stabilized and that additional rounds are
unlikely to meaningfully change the final list.

The rejection strategy moderates this stability --- exploration trade-off.
\textit{Majority} rejection yields smoother and consistently higher reliability than \textit{Aggressive} rejection, because fewer cities are
discarded and the candidate set churns less.
This improved stability comes at the cost of slightly slower convergence (RQ1: \autoref{section: RQ1}) and often slightly weaker diversity gains (RQ2: \autoref{section: RQ2}).
\paragraph{Hallucinations decrease with grounded feedback.}
\change{Structured output constraints reduce a large portion of out-of-inventory hallucinations by enforcing syntactic validity; remaining catalog-membership violations are caught by the moderator's post-generation validation.}
Residual hallucinations can still occur when agents reintroduce cities that were previously rejected by the moderator.
Across models, we observe that hallucination tendency decreases across rounds: after initial exploration, agents increasingly comply with
moderator feedback and stop proposing invalid or repeatedly rejected items.

\tors{For example, under the \textit{Aggressive} rejection strategy, the average agent-level hallucination rate for \gemini{} is approximately 0.002 in round 1, decreases to 0.000093 by round 2, and converges to \change{0} in subsequent rounds. Similarly, for \gemmaFour{} under the \textit{Majority} strategy, the rate starts at around 0.0017 in round 1, drops to 0.000056 by round 2, and reaches \change{0} thereafter. Comparable dynamics are observed across the remaining backbones; we therefore omit the full numerical breakdown for brevity. By the final rounds, hallucination rates converge toward zero across backbones, showing that iterative feedback can enforce grounding even when logit-level constraints are not natively supported.}

\vspace{2.5mm}
\RQBox{RQ3 Summary.}{Agents initially explore aggressively (lower reliability in early rounds) but become increasingly stable and compliant
as feedback accumulates. \textit{Majority} rejection promotes smoother stability; \textit{Aggressive} rejection promotes stronger churn and faster correction.
Across both policies, hallucination tendency decreases sharply across rounds under catalog grounding and structured feedback.}

\subsection{RQ4: Time and cost complexity}
\label{section: RQ4}

\begin{figure*}[htbp]
    \centering
    \begin{subfigure}[b]{0.45\textwidth}
        \includegraphics[width=\textwidth]{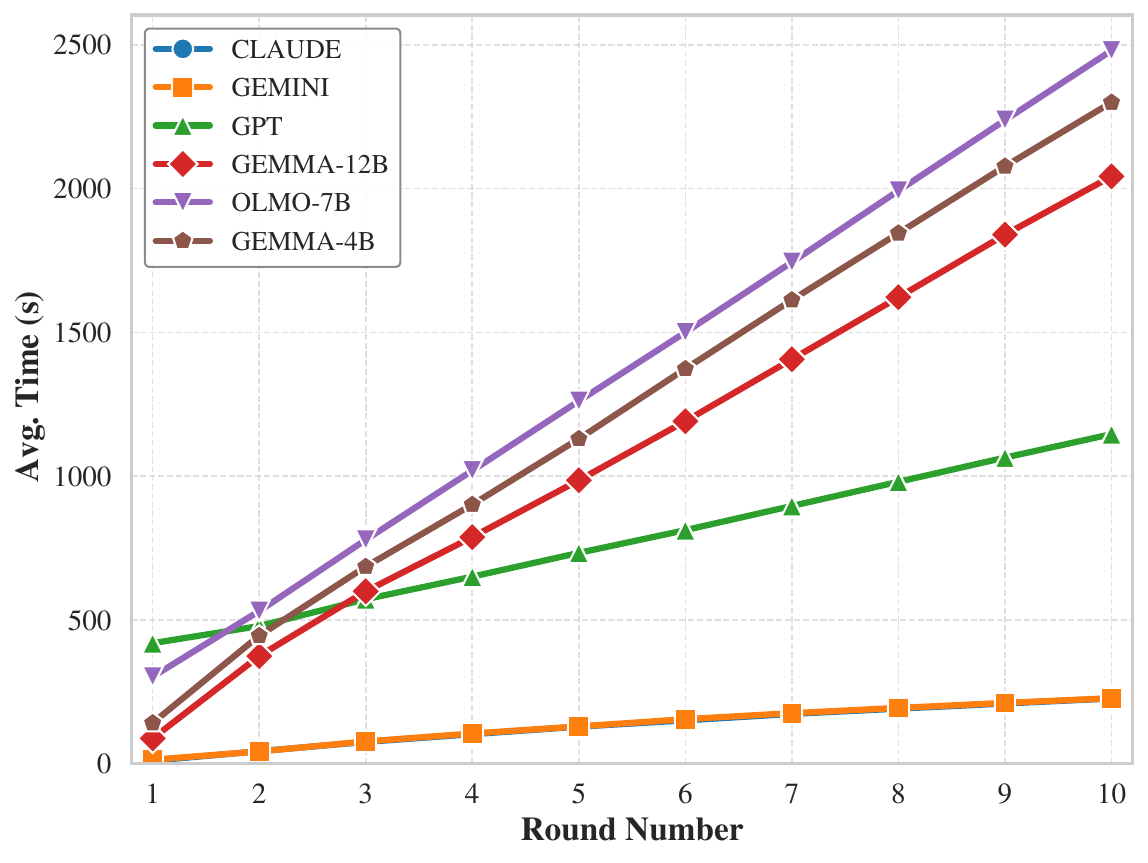}
         \caption{Time Taken}
    \end{subfigure}
    \begin{subfigure}[b]{0.45\textwidth}
        \includegraphics[width=\textwidth]{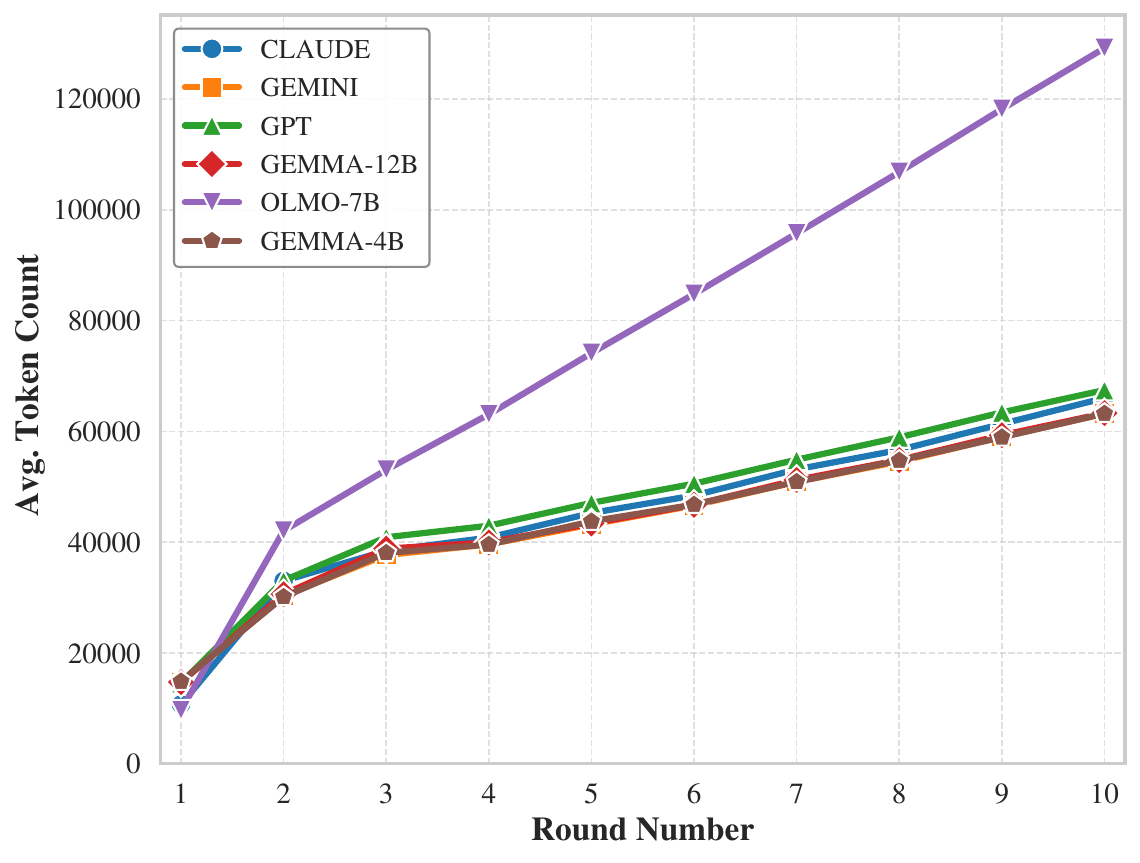}
        \caption{Tokens Used}
    \end{subfigure}
    \caption{Average wall-clock time (left) and token usage (right) as a function of the refinement round for all models using the \textit{Aggressive} rejection strategy. 
    \change{Note that the \textit{Avg.\ Time (s)} axis reflects cumulative wall-clock time across all 10 rounds; for API-served models, the per-round wall-clock time equals the maximum over the 3 parallel agent calls plus moderator overhead, whereas for locally served open-source models, agents execute sequentially on-device without endpoint-level batching.} The token count is the total number of tokens used across all agents and the moderator across all rounds. }
    \label{fig: time taken aggressive}
\end{figure*}

\noindent\textbf{RQ4 asks:} \emph{What time and token overheads does multi-round \change{refinement} introduce, and how can they be mitigated?}
~\autoref{fig: time taken aggressive} summarizes time and token usage over rounds.

\paragraph{Overhead scales approximately linearly with rounds.}
As expected, both inference latency and token usage increase with the number of rounds.
However, absolute costs vary substantially by deployment mode.
For API-served proprietary models, cumulative latency stays below 500 seconds for a full 10-round run:
\change{\gemini{} averages 133.89 s (\textit{Aggressive}) and 211.63 s (\textit{Majority}); \claude{} averages 382.01 s under \textit{Majority} (see \autoref{fig: time taken aggressive} for the \textit{Aggressive} result).}
For locally served open-source models, costs are much higher: under \textit{Aggressive} rejection, \gemmaTwelve{} reaches 2042.61 seconds and
\gemmaFour{} reaches 2299.60 seconds.
This gap reflects the absence of endpoint-level batching/parallelization and the sequential execution of multiple agents and moderation on
local hardware.
\change{\gemmaFour{}'s higher cumulative latency relative to the larger \gemmaTwelve{} is counterintuitive but reflects greater per-round output-validation overhead: the smaller model produces invalid or out-of-catalog outputs more frequently, triggering additional moderator processing and candidate replacement steps rather than a higher per-token generation cost.}

Token usage exhibits a similar pattern under practical deployment.
\change{When utilizing early stopping (which averages $\sim$4 rounds), cumulative token counts across all rounds sum to $\sim$44k for \gemini{} (both strategies), while \claude{} (\textit{Majority})
and \olmo{} (\textit{Majority}) accumulate higher totals (68,578.59 and 75,874.79 tokens respectively), consistent with more verbose
justifications and longer contextual prompts that grow with each additional round (\autoref{fig: time taken majority}).}
\textit{Aggressive} rejection typically increases token use by inducing more frequent candidate replacement.

\paragraph{Early stopping as a practical mitigation.}
The main operational takeaway is that early stopping captures most of the quality gains (RQ1) while substantially reducing costs.
When we stop at round 4, \gemini{} (\textit{Aggressive}) drops to 60.41 seconds, and \claude{} (\textit{Majority}) drops to 87.16 seconds.
In the local cluster setting, early stopping reduces \gemmaTwelve{} (\textit{Aggressive}) to 462.51 seconds and \gemmaFour{} (\textit{Majority})
to 241.51 seconds.
Thus, while multi-round \change{refinement} is more expensive than single-shot baselines, dynamic termination makes the approach considerably more
practical---especially for API-served backbones.

\torsRTwo{Beyond termination strategies, costs can be further mitigated by utilizing "non-reasoning" or distilled models for specific agent roles, which significantly reduces per-token pricing and inference time without compromising the structural grounding of the recommendation~\cite{liu2024agentlite}. 
For production environments, we suggest \textit{agent pruning}~\cite{zhang2026safesieve}, i.e., deactivating specific agents once their proposals stabilize, and \textit{semantic caching}~\cite{bang2023gptcache, mohandoss2024context}, which reuses refinement states for semantically similar user queries to avoid redundant computation. However, we leave the implementation and evaluation of these strategies to future work, as they require additional infrastructure and are not natively supported by all backbones.}


\paragraph{Deployment scope.}
\torsRTwo{Despite the gains from early stopping, the current implementation remains too slow for interactive real-time recommendation. We therefore position \sysname{} as a research prototype suitable for offline or high-latency planning scenarios. Techniques such as smaller role-specific models, batching, agent pruning, and semantic caching may reduce cost in future higher-throughput settings, but we do not claim production readiness in the current system.}



\paragraph{Reproducibility resources.}
\torsRThree{The \sysname{} repository\footnote{https://github.com/ashmibanerjee/collab-rec} provides a complete workflow with dependencies, API / backend setup, example requests, and GPU / cloud instructions to enable replication and experimentation.}

\RQBox{RQ4 Summary.}{Costs scale with the number of rounds and are highly deployment-dependent: API-served models are substantially
faster than locally hosted open-source models in our setup. Early stopping at $\sim$4 rounds is an effective mitigation that preserves most
quality gains while reducing latency and token usage.}

\subsection{RQ5: Ablation Analysis}\label{section: RQ5}

\change{\noindent\textbf{RQ5 asks:} \emph{What is the individual contribution of each moderator scoring component (\textit{Relevance}, \textit{Reliability}, \textit{Hallucination penalty}, and \textit{ranking}) to overall recommendation quality?}}

\begin{figure}[htbp]
    \centering
    
    \includegraphics[width=0.9\textwidth]{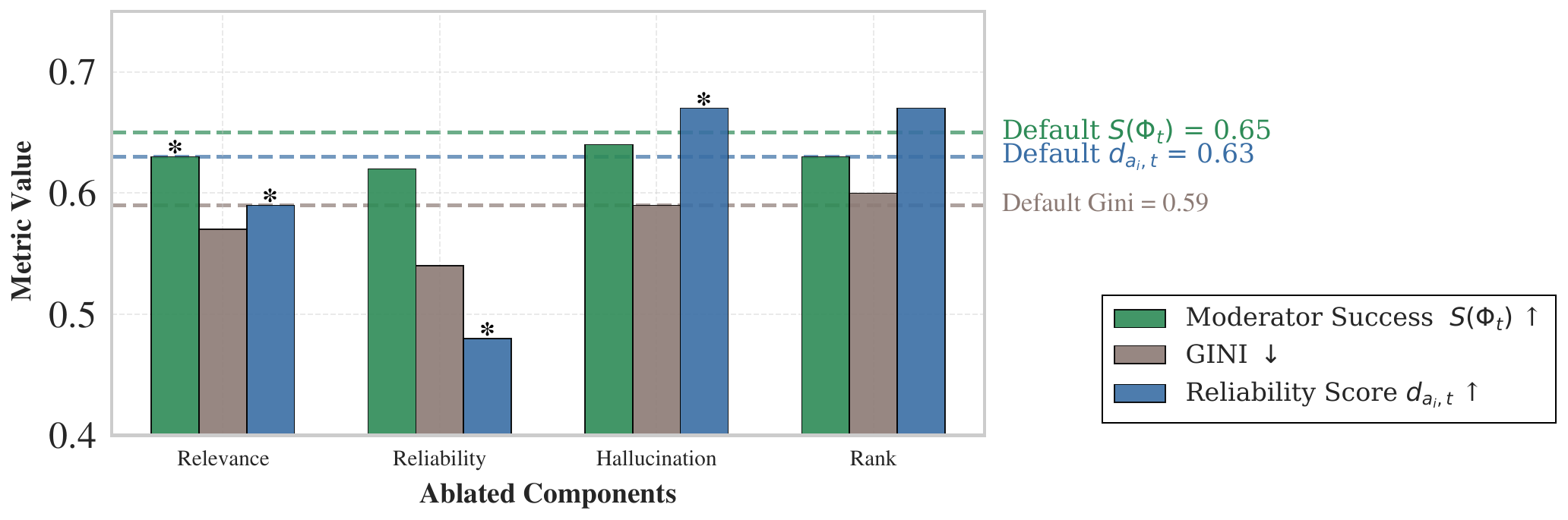}
    \caption{Olmo-7b ablation results with default values (with all components) shown as horizontal dashed lines. Moderator Success is shown in green, Gini in brown, and Reliability Score in blue. The horizontal lines indicate the default values for each metric, so that the effect of removing each component is clearly visible. The stars indicate significant differences from the default setting, with $\ast$ for $p < 0.05$. The results show that removing the Relevance and Reliability components leads to significant drops in Moderator Success and Reliability Score, while removing the Hallucination component has a less pronounced effect. Diversity metrics (Gini) remain relatively stable across ablations, suggesting that popularity bias mitigation is not solely dependent on any single scoring component.}
    \label{fig: olmo_ablation}
\end{figure}

\tors{To assess the contribution of individual moderator components to overall recommendation quality, we conduct an ablation study on two representative models: \gemini{} (a high-capacity closed-source model) and \olmo{} (a representative open-source model). Experiments are performed on 150 stratified queries using the \textit{Aggressive} rejection strategy with five rounds and early stopping, as described in~\autoref{section: termination}. 
We systematically ablate the core components of the moderator’s scoring function—\textit{Relevance} ($\relevancescore$), \textit{Reliability} ($\reliabilityscore$), \textit{Hallucination} ($\hallucinationscore$), and the \textit{ranking mechanism} ($\operatorname{rank}_{a_i,t}(c)$) defined in~\autoref{eq:incremental_score}. 
Each ablated variant is compared against the default $\MAMIearly$ configuration, and we report the mean values of the evaluation metrics (\textit{Moderator Success}, \textit{Diversity (Gini)}, and \textit{Reliability scores}) for each setting.
Statistical significance is evaluated using paired t-tests~\cite{hsu2014paired} $(p < 0.05)$ against the default distribution, with significant differences highlighted using a $\ast$ on the bar plot in~\autoref{fig: olmo_ablation}.}

\tors{Across models, we observe distinct sensitivity patterns. While \gemini{} remains largely stable under component removal, exhibiting only marginal fluctuations across \textit{Moderator Success}, diversity, and reliability, \olmo{} shows substantially greater dependence on explicit moderator signals. Removing \textit{Relevance} reduces Moderator Success from 0.65 to 0.63 and lowers reliability from 0.63 to 0.59 (both significant). Eliminating \textit{Reliability} further decreases reliability to 0.48 and slightly increases diversity (Gini 0.54), underscoring the stabilizing role of structured scoring signals for smaller models. In contrast, removing the hallucination penalty results in minimal changes in \textit{Moderator Success} and \textit{diversity}, suggesting that hallucination control is largely enforced through structured output constraints rather than the moderator’s scoring term. Notably, diversity metrics remain broadly stable across settings (Gini 0.54–0.60), indicating that popularity bias mitigation persists even under component-level modifications.
Since \gemini{} exhibits only minor variation across ablations, we present a detailed visualization of \olmo{} in~\autoref{fig: olmo_ablation}, where the impact of removing individual moderator components is more pronounced and thus more informative.}

\yashar{This ablation study is intentionally limited. It isolates the effect of removing individual scoring components while keeping $\lambda_r=\lambda_d=\lambda_h=1$ fixed, and therefore it is not a sensitivity analysis over the scalarization weights. It also covers two model families, 150 queries, the \textit{Aggressive} rejection strategy, and a five-round budget. The results should consequently be interpreted as indicative evidence about component behavior rather than as a comprehensive robustness claim. Systematic exploration of alternative $\lambda$ configurations, learned weight calibration, and broader model/query settings remains future work.}

\vspace{2.5mm}
\RQBox{RQ5 summary.}{The ablation study indicates that the framework is robust for higher-capacity models, while the moderator’s scoring components are particularly beneficial for stabilizing performance in smaller LLMs. \torsRThree{These results are encouraging but remain based on fixed $\lambda$ values, two model families, and 150 queries. Exploring alternative weight configurations and a broader range of models remains a promising direction for future work.}}


\subsection{Summary of Results} \label{subsection: results_summary}
\tors{To summarize, our results show that multi-agent multi-round \change{refinement} (MAMI) consistently outperforms single-agent (SASI) and single-round (MASI) baselines in \change{moderator success}, with most relevance gains saturating after \change{rounds 4--5} (RQ1). Beyond quality improvements, MAMI also mitigates popularity bias by reducing recommendation concentration (lower Gini, higher entropy) and improving long-tail coverage (RQ2). The iterative \change{refinement} process further stabilizes agent behavior: structured feedback and rejection signals progressively reduce hallucinations and enhance groundedness (RQ3). While additional rounds increase computational cost, early stopping retains the majority of quality improvements while substantially lowering latency for API-based models, though local execution remains resource-intensive (RQ4). Finally, the ablation analysis confirms that the multi-objective scoring behaves as expected: smaller models benefit strongly from explicit relevance and reliability signals, whereas the hallucination penalty term has only a minor effect, since grounding is mostly enforced by the structured-output constraints (RQ5).}
\paragraph{Additional evidence (Appendix).}
To complement the main RQ1 results, \autoref{appendix: relevance results} reports per-round success trajectories under the \textit{Majority} rejection strategy, stratified by query popularity levels. The trajectories show that improvements concentrate in the early rounds and typically stabilize around \change{rounds 4--5}.~\autoref{appendix: relevance results} also includes extended twenty-round runs, confirming that additional rounds yield diminishing returns rather than delayed late-stage improvements, which supports the early-stopping criterion in~\autoref{eq: early_stop}.

\autoref{appendix: diversity results} provides distribution-level diagnostics under the \textit{Majority} rejection strategy. The kernel density plots visualize how multi-round \change{refinement} shifts recommendations away from the highest-popularity region, and the Lorenz curves show reduced concentration across the catalog relative to single-shot and single-round baselines. These plots provide an interpretable complement to the aggregate Gini, entropy, and coverage metrics reported in the main text (\autoref{section: RQ2}).

\autoref{appendix: complexity results} reports round-by-round time and token growth under the \textit{Majority} rejection strategy. The plots make explicit that both wall-clock time and token usage increase approximately monotonically with additional rounds, reinforcing the practical value of early stopping once relevance stabilizes.

\paragraph{Reproducibility details (Appendix).}
The complete prompt templates, role constraints, and \torsRTwo{iterative constrained refinement} context injections used to elicit specialist behavior are documented in~\autoref{appendix: prompts}. This documentation clarifies the exact mechanisms by which moderator feedback constrains revisions, limits churn (to at most 3 replacements), and enforces closed-catalog generation through structured outputs.
The ablation experiments in~\autoref{section: RQ5} rely on the same structured-output constraints and revision rules used in the main experiments.

\section{Conclusion}
\label{section: conclusion}

This paper presents \sysname{}, a large-language-model-based agentic recommendation framework for tourism recommendation under competing stakeholder objectives. The method decomposes the recommendation task into three specialist agents that represent personalization, popularity, and sustainability objectives, and coordinates them through a deterministic moderator. The moderator enforces catalog grounding, validates structured constraints, aggregates proposals using a transparent multi-objective scoring policy, and stabilizes computation through an online termination protocol. \change{While \sysname{} demonstrates the potential of multi-agent coordination for balanced tourism recommendations, it remains a research prototype and is not yet optimized for real-time deployment.}

Across a stratified evaluation of $900$ benchmark queries and six model backbones, the empirical results show that multi-agent multi-round \change{refinement} \change{consistently improves} \change{moderator success} while simultaneously reducing short-head concentration relative to single-shot and single-round variants. The analyses further show that multi-round execution is most beneficial in the first few rounds: improvements typically plateau after a small number of iterations, which motivates early stopping as a practical mechanism for reducing latency and token usage without sacrificing most of the quality gains.
\yashar{\paragraph{Limitations and future work.}
The present evaluation is offline and based on a synthetic benchmark with a closed catalog. Although this design enables controlled catalog grounding and reproducible constraint checks, it does not demonstrate real-world user satisfaction, logged interaction performance, or actual sustainability effects on tourism flows. The moderator score is also a heuristic scalarization of agent- and list-level diagnostics rather than an item-level utility model, and we do not provide a sensitivity analysis over the $\lambda$ weights. Furthermore, the \MILP{} baseline should be viewed as a structured-input reference, not as a substitute for stronger catalog-grounded reranking and constrained-diversification baselines such as deterministic filter-rankers, MMR/xQuAD-style methods, Pareto/weighted rerankers, or explicit diversity-constrained optimizers. Finally, the current latency is too high for interactive deployment. Future work should evaluate the framework with human preference judgments and logged interactions, calibrate or learn the moderator weights, add item-level utility estimates, compare against stronger catalog-only rerankers, and investigate batching, caching, agent pruning, and smaller role-specific models.}

\section*{GenAI Usage Disclosure}

Generative AI tools, including ChatGPT (OpenAI), Claude (Anthropic), and Gemini (Google), were used in supporting roles during this research. These tools assisted with refining code snippets, improving sentence clarity, identifying grammatical inconsistencies, and enhancing readability. They were not used to generate the scientific contributions, analyses, results, or core intellectual content of this manuscript.

All AI-generated suggestions were carefully reviewed, edited wherever necessary, and validated by the authors. The authors take full responsibility for the accuracy, originality, integrity, and final content of this manuscript.

\appendix \label{appendix: supplementary results}
\section{Additional Results for RQ1: Relevance Analysis} \label{appendix: relevance results}

\begin{figure*}[htbp]
    \centering

    \begin{subfigure}{0.32\textwidth}
        \includegraphics[width=\linewidth]{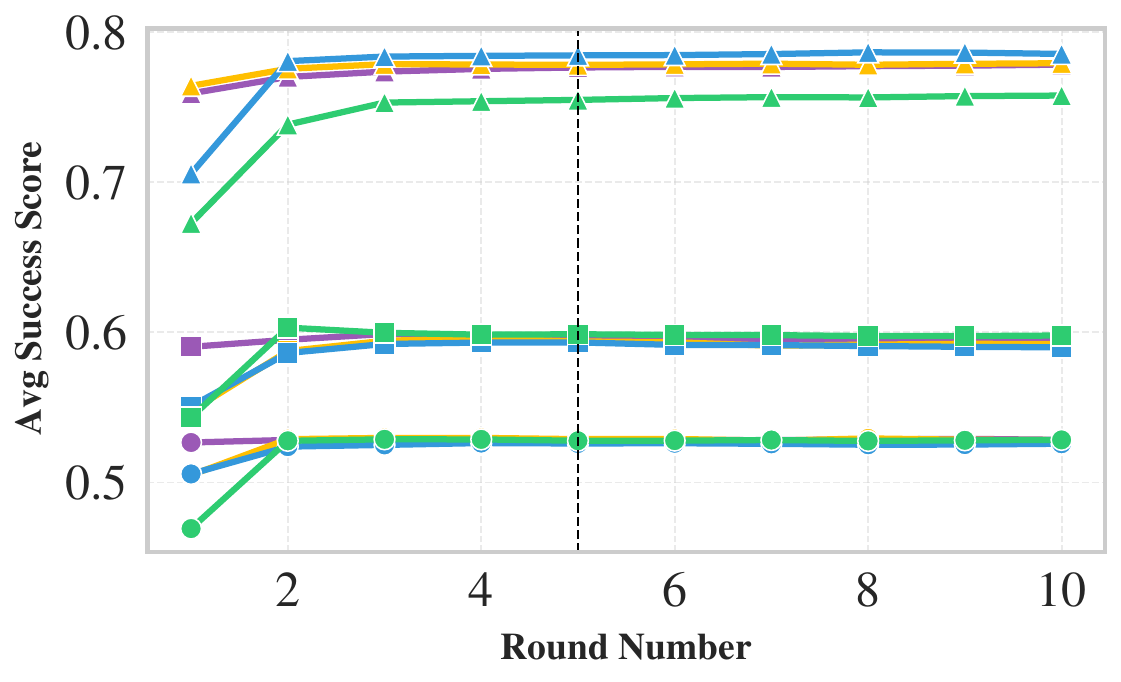}
        \caption{Claude}
    \end{subfigure}
    \begin{subfigure}{0.32\textwidth}
        \includegraphics[width=\linewidth]{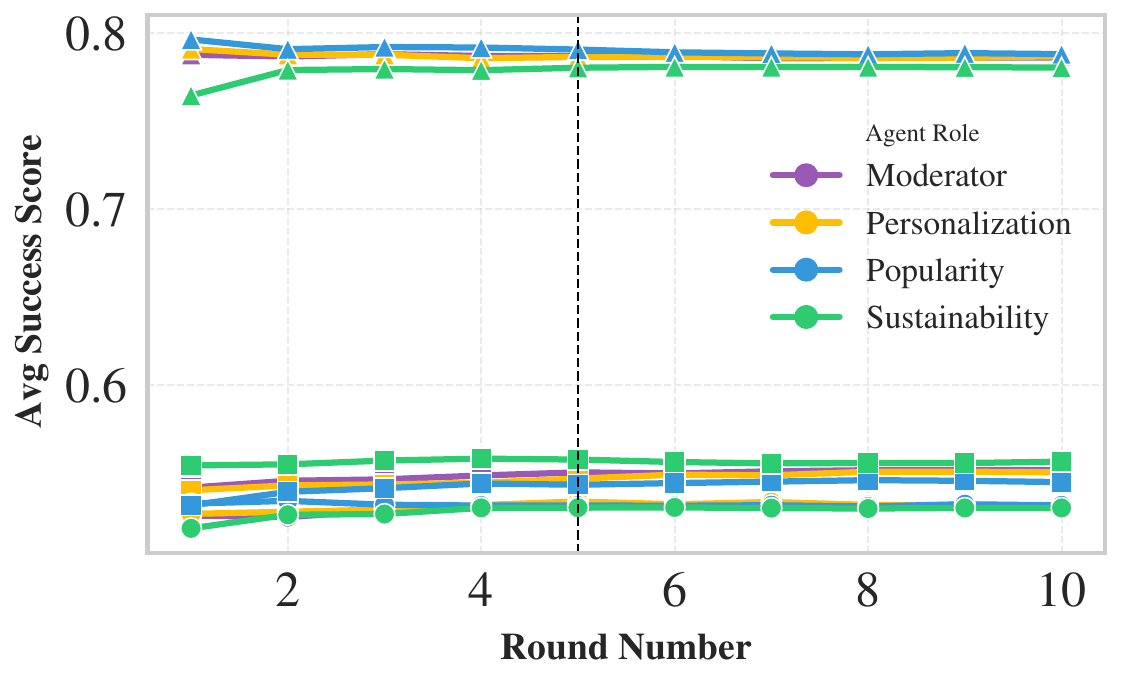}
        \caption{Gemini}
    \end{subfigure}
    \begin{subfigure}{0.32\textwidth}
        \includegraphics[width=\linewidth]{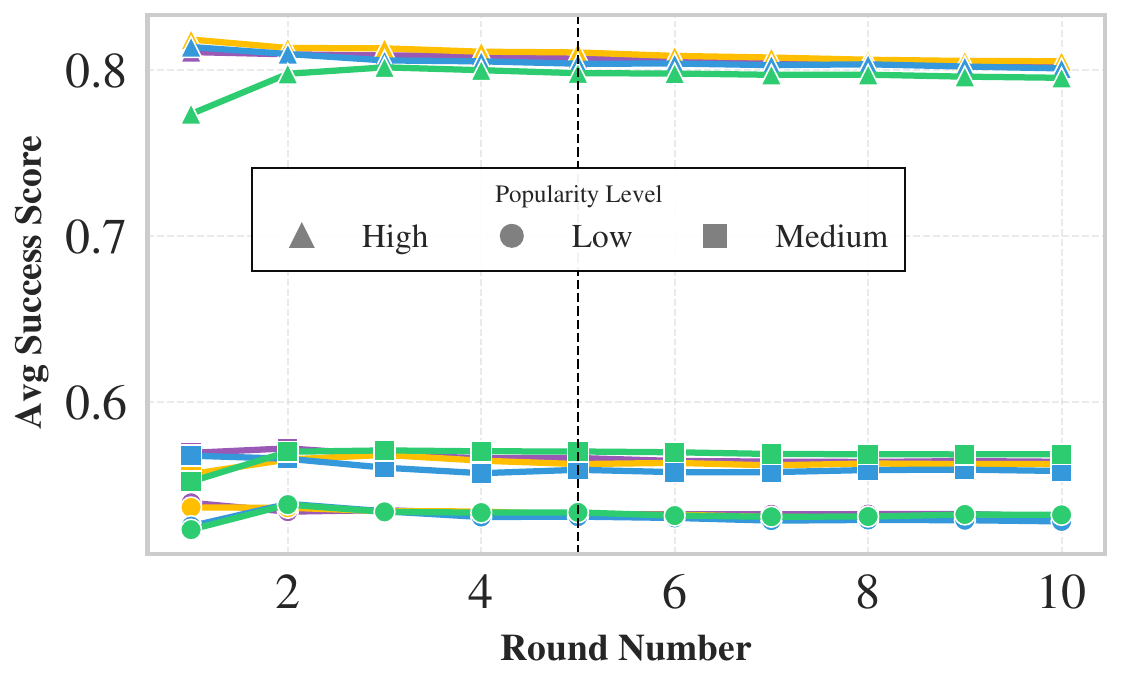}
        \caption{GPT-OSS-20B}
    \end{subfigure}

    \vspace{0.5em}

    \begin{subfigure}{0.32\textwidth}
        \includegraphics[width=\linewidth]{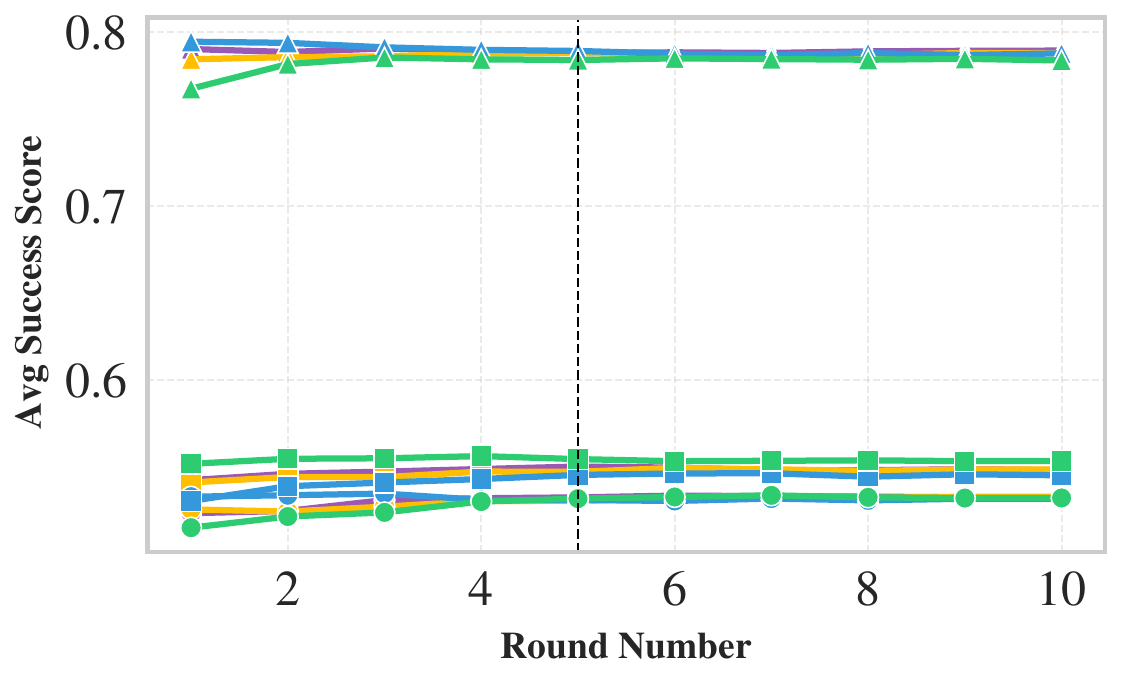}
        \caption{Gemma-12b}
    \end{subfigure}
    \begin{subfigure}{0.32\textwidth}
        \includegraphics[width=\linewidth]{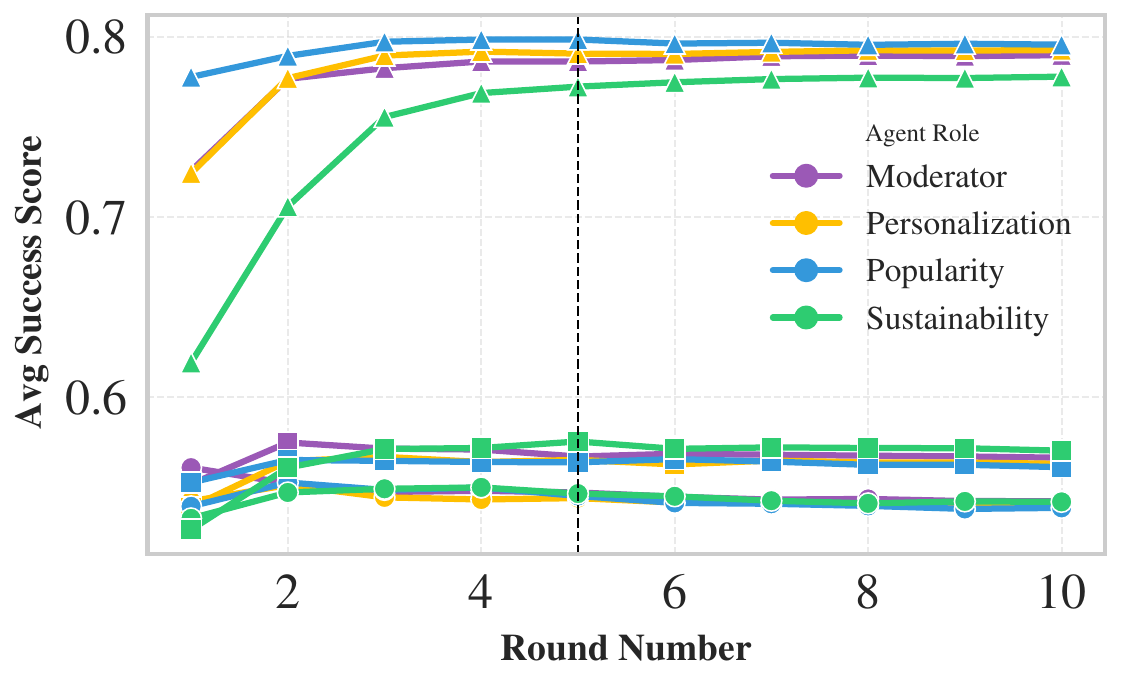}
        \caption{Olmo-7b}
    \end{subfigure}
    \begin{subfigure}{0.32\textwidth}
        \includegraphics[width=\linewidth]{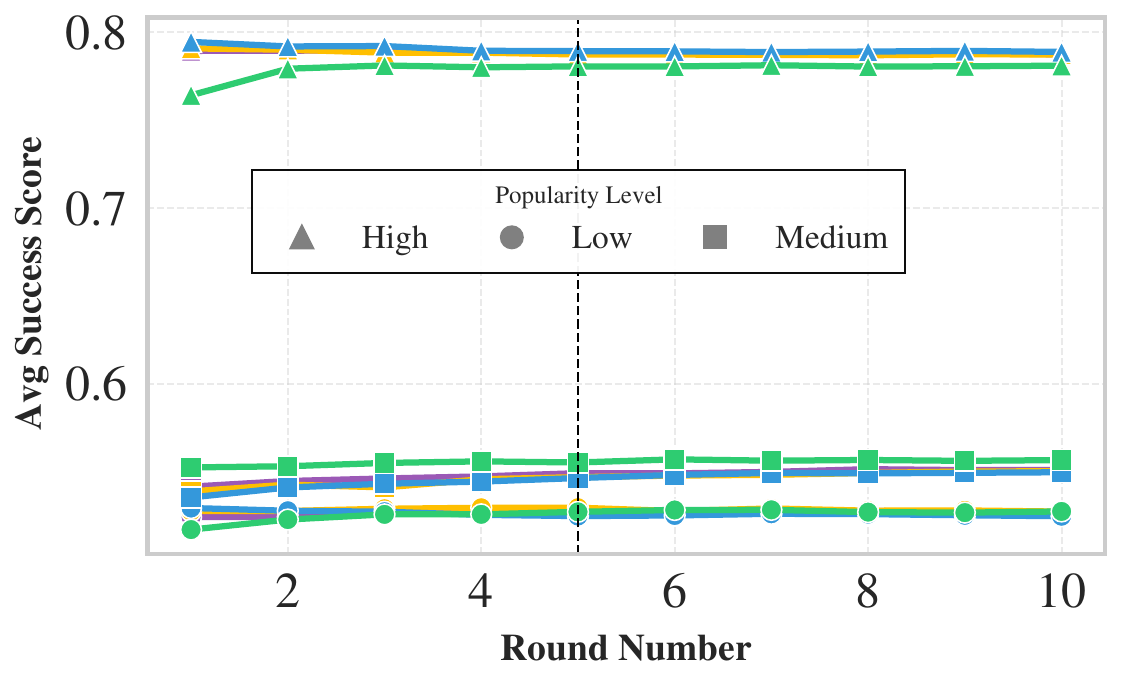}
        \caption{Gemma-4b}
    \end{subfigure}

   \caption{Average agent success scores over \torsRTwo{refinement} rounds under the \textit{Majority} rejection strategy. The plots track performance for the \textcolor{personalization}{\textit{Personalization}}, \textcolor{popularity}{\textit{Popularity}}, \textcolor{sustainability}{\textit{Sustainability}}, and \textcolor{moderator}{\textit{Moderator}} agents across LLM backbones. Results are stratified by query popularity level (low, medium, high). Across models, success improves rapidly in early rounds and stabilizes around \change{rounds 4--5}, supporting patience-based early stopping as a practical approximation to longer runs.
    }
    \label{fig: relevance_plots_majority}
\end{figure*}

\begin{figure*}[htbp]
    \centering

    \begin{subfigure}{0.32\textwidth}
        \includegraphics[width=\linewidth]{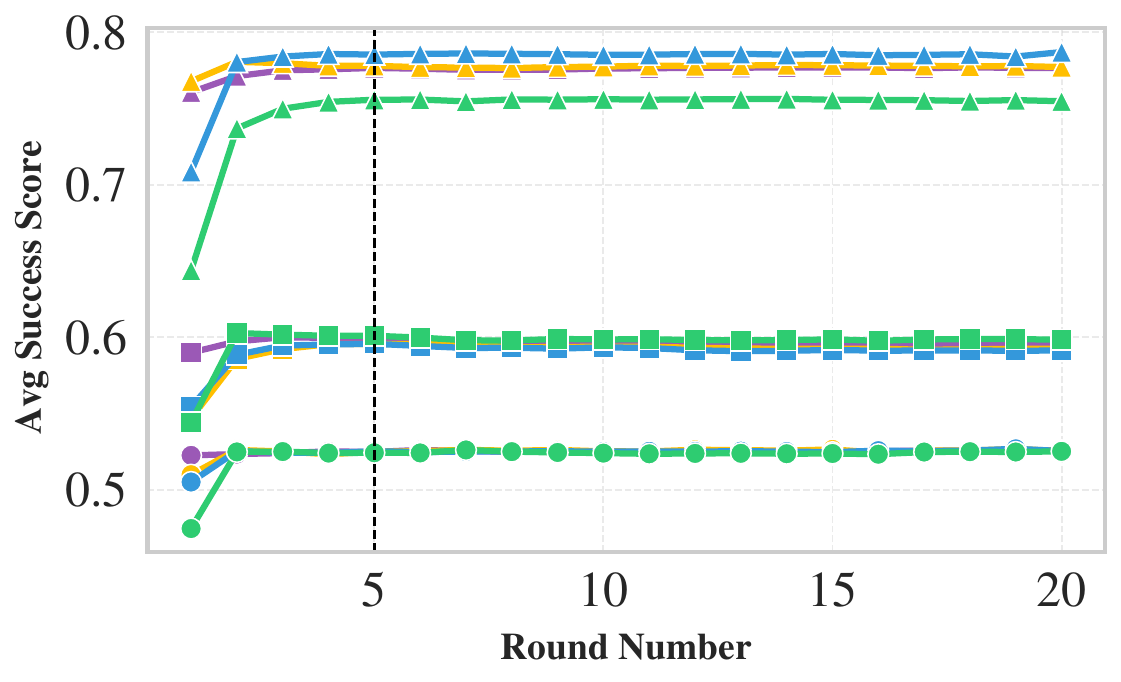}
        \caption{Claude (M)}
    \end{subfigure}
    \begin{subfigure}{0.32\textwidth}
        \includegraphics[width=\linewidth]{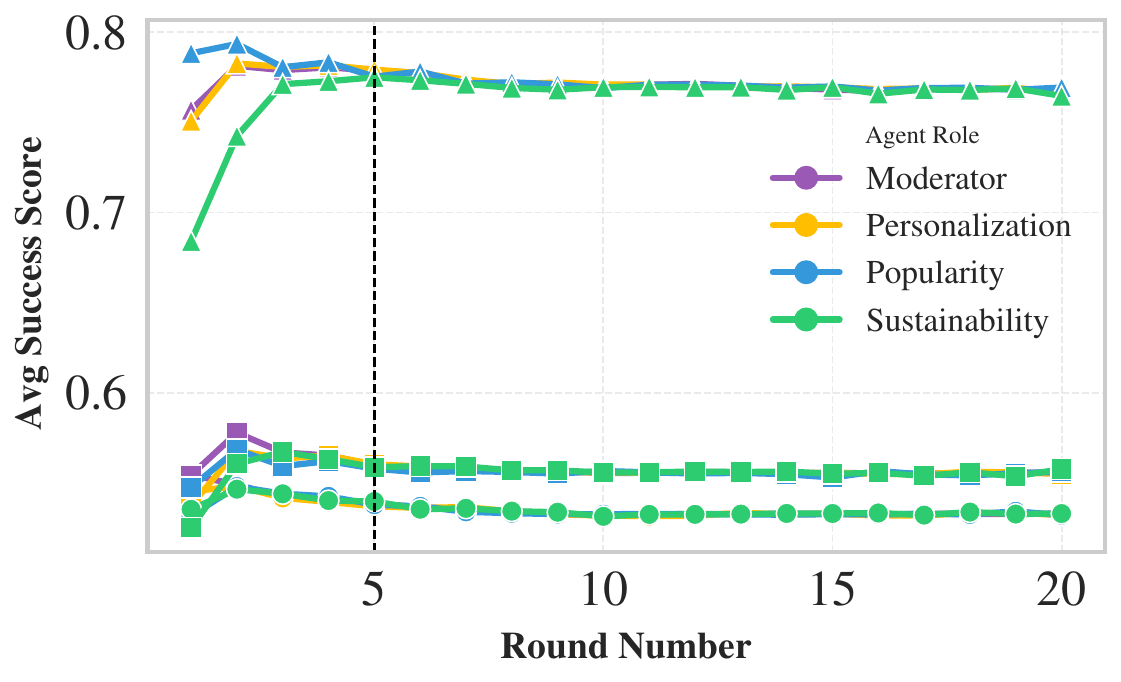}
        \caption{Olmo-7b (A)}
    \end{subfigure}
    \begin{subfigure}{0.32\textwidth}
        \includegraphics[width=\linewidth]{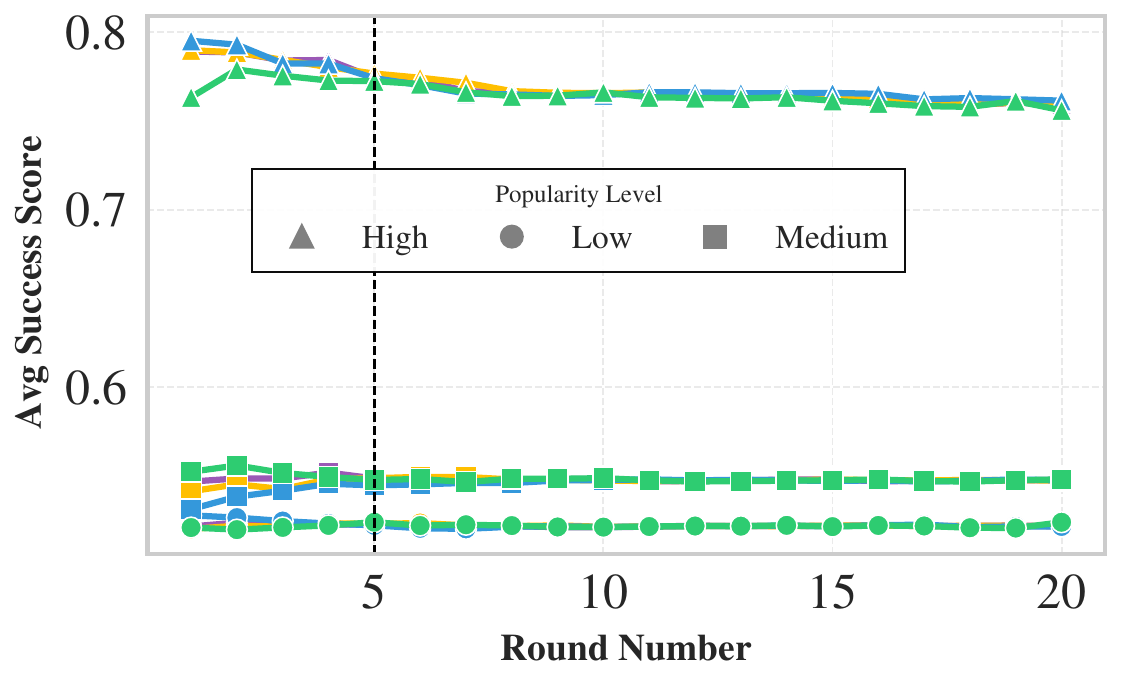}
        \caption{Gemma-4b (A)}
    \end{subfigure}
    \caption{Extended trajectories for selected model backbones over twenty rounds. \change{The plots confirm that convergence occurs at approximately round 4, with quality gains plateauing by round 5, and that this is not a transient local optimum}: additional rounds yield diminishing returns in grounded success, consistent with the early-stopping criterion in~\autoref{eq: early_stop}.
     \textbf{M} = \textit{Majority} voting strategy, \textbf{A} = \textit{Aggressive} rejection strategy.
    }
    \label{fig: relevance_plots_20_rounds}
\end{figure*}

This appendix provides supplementary relevance diagnostics that complement the main RQ1 analysis in~\autoref{section: RQ1}. In the main text, figures emphasize the \textit{Aggressive} rejection strategy for clarity; here, we report additional trajectories under the \textit{Majority} rejection strategy and include extended runs beyond the default ten-round budget to verify that early stopping does not truncate late improvements (\autoref{fig: relevance_plots_majority} -- \ref{fig: relevance_plots_20_rounds}).

\section{Additional Results for RQ2: Diversity Analysis} \label{appendix: diversity results}
\begin{figure*}[htbp]
    \centering

    \begin{subfigure}{0.32\textwidth}
        \includegraphics[width=\linewidth]{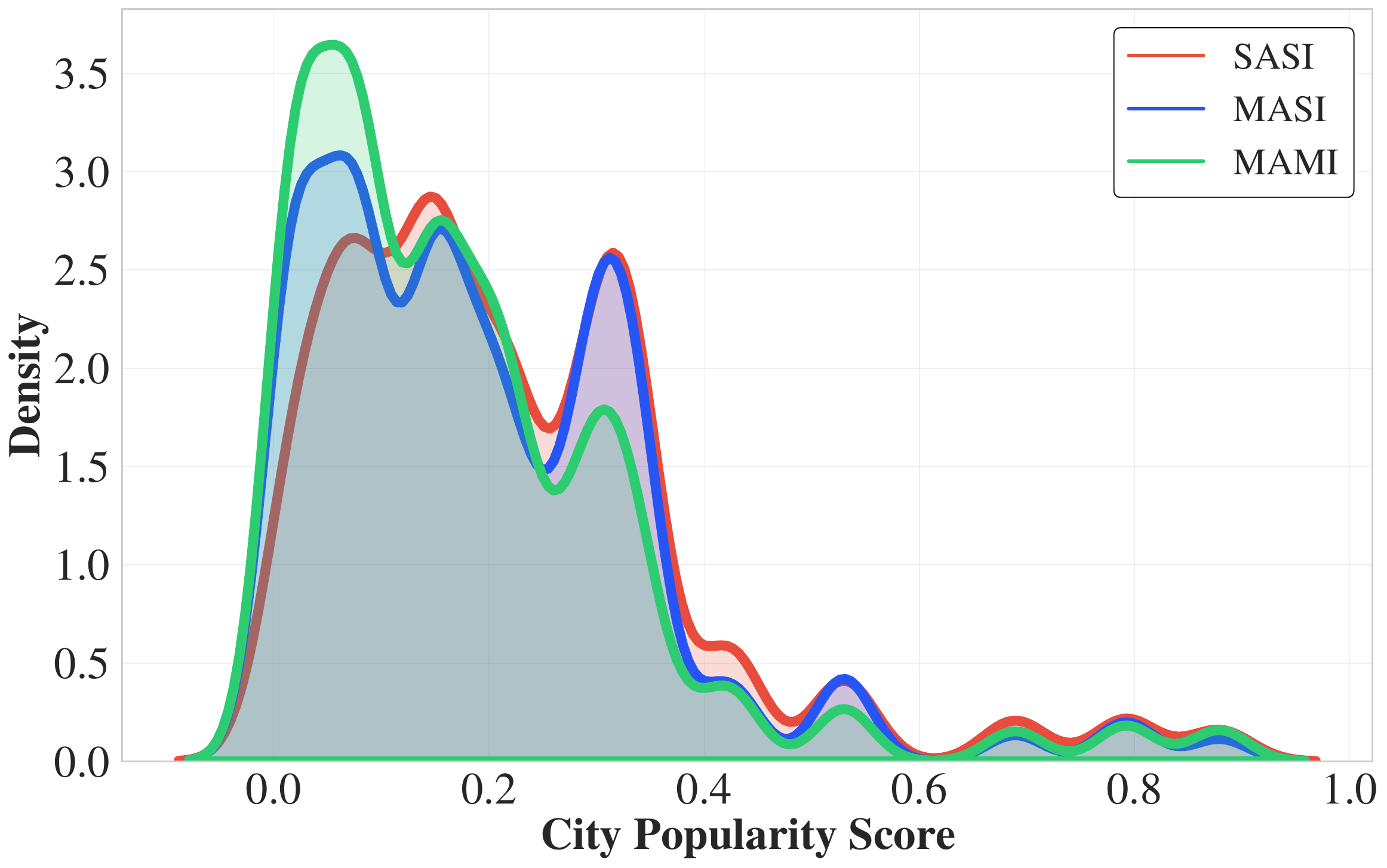}
        \caption{Claude}
    \end{subfigure}
    \begin{subfigure}{0.32\textwidth}
        \includegraphics[width=\linewidth]{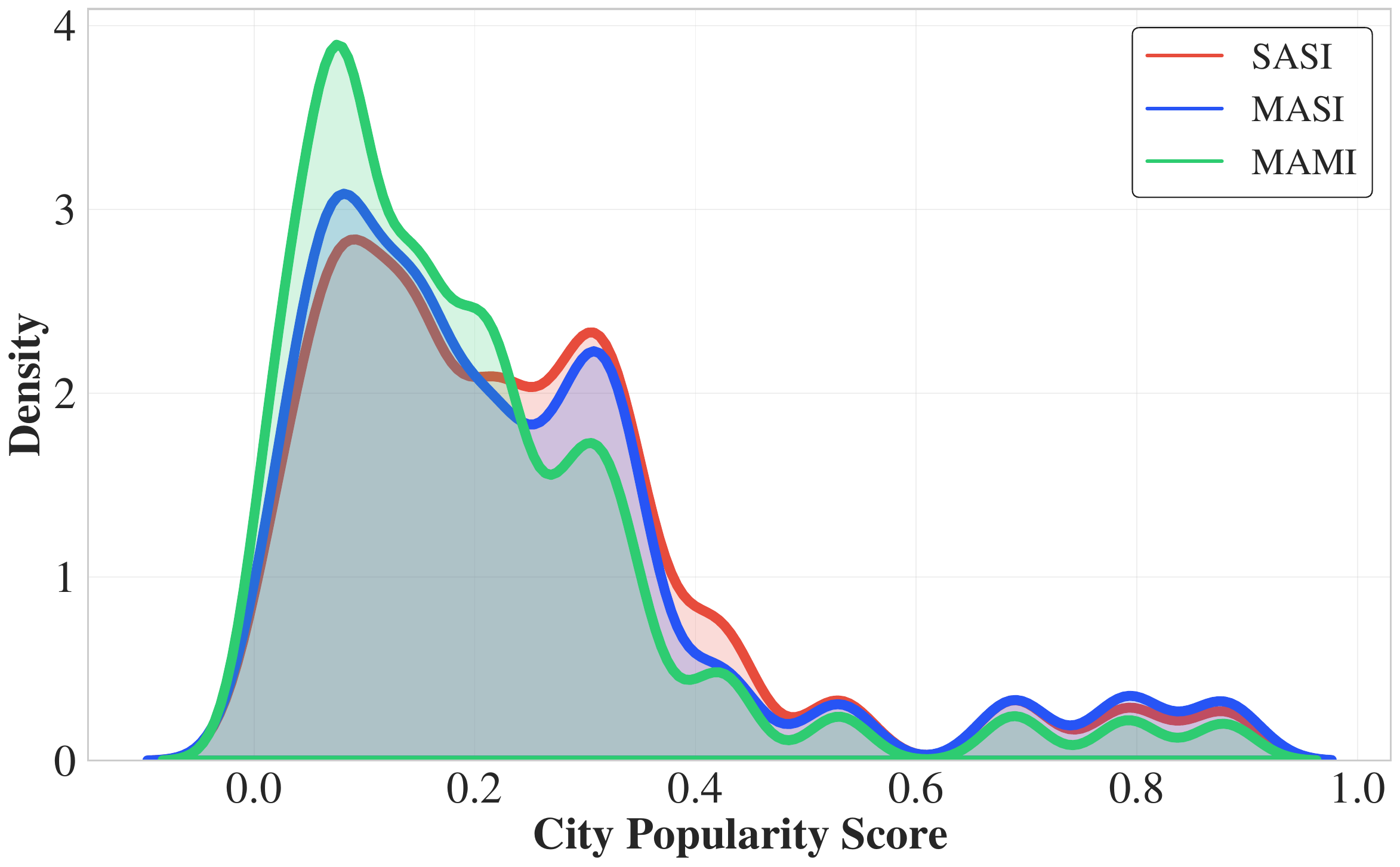}
        \caption{Gemini}
    \end{subfigure}
    \begin{subfigure}{0.32\textwidth}
        \includegraphics[width=\linewidth]{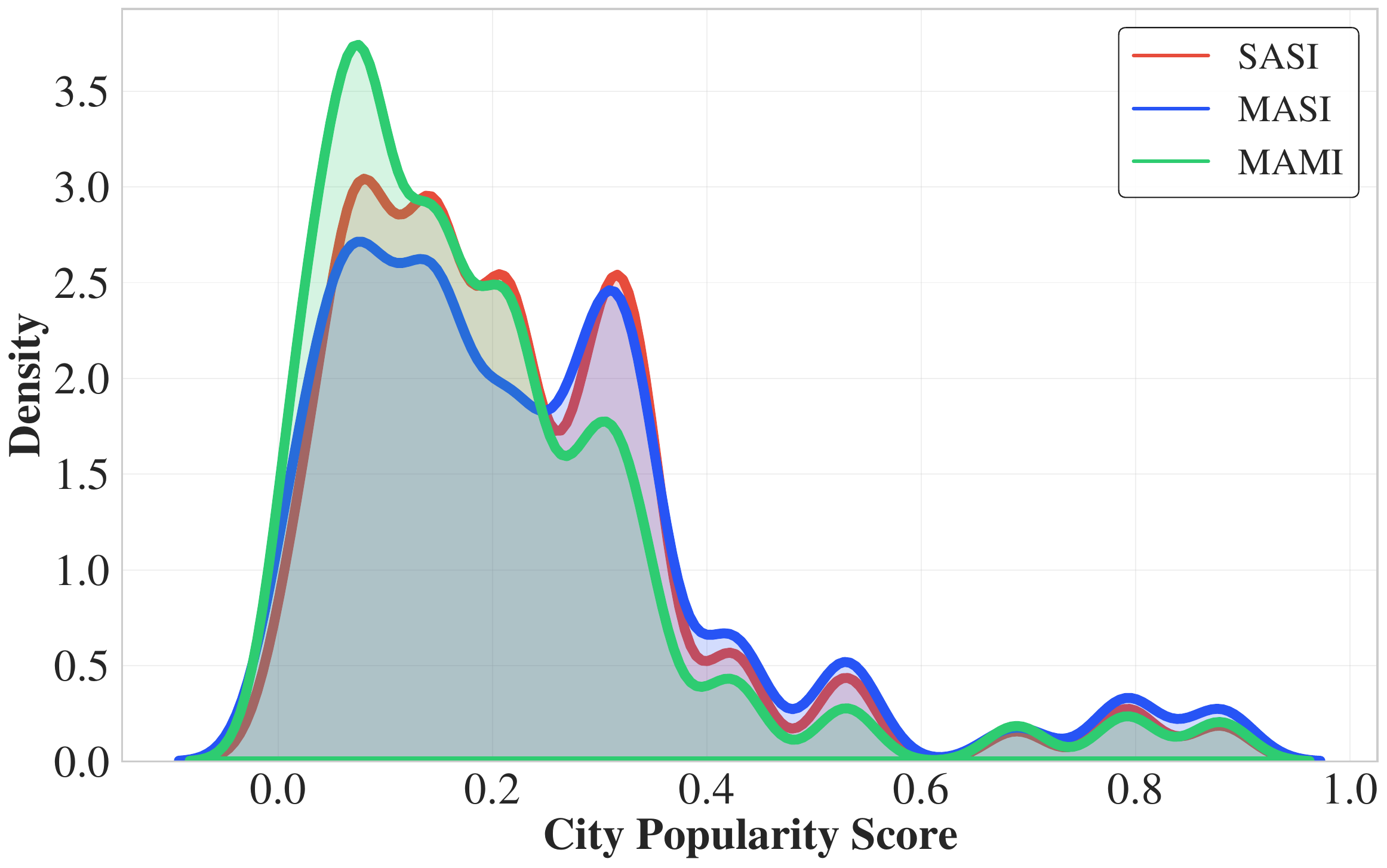}
        \caption{GPT-OSS-20B}
    \end{subfigure}

    \vspace{0.5em}

    \begin{subfigure}{0.32\textwidth}
        \includegraphics[width=\linewidth]{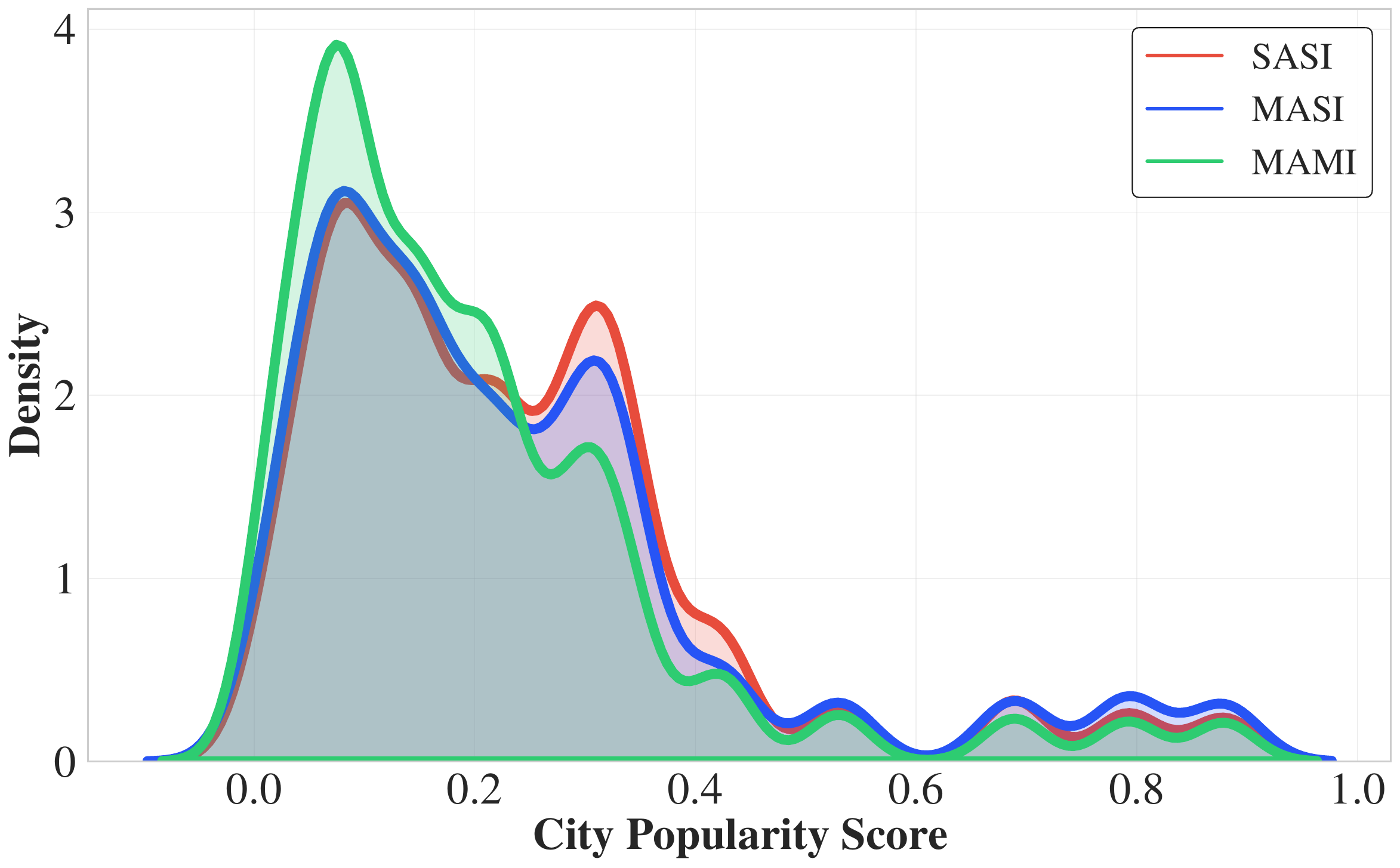}
        \caption{Gemma-12b}
    \end{subfigure}
    \begin{subfigure}{0.32\textwidth}
        \includegraphics[width=\linewidth]{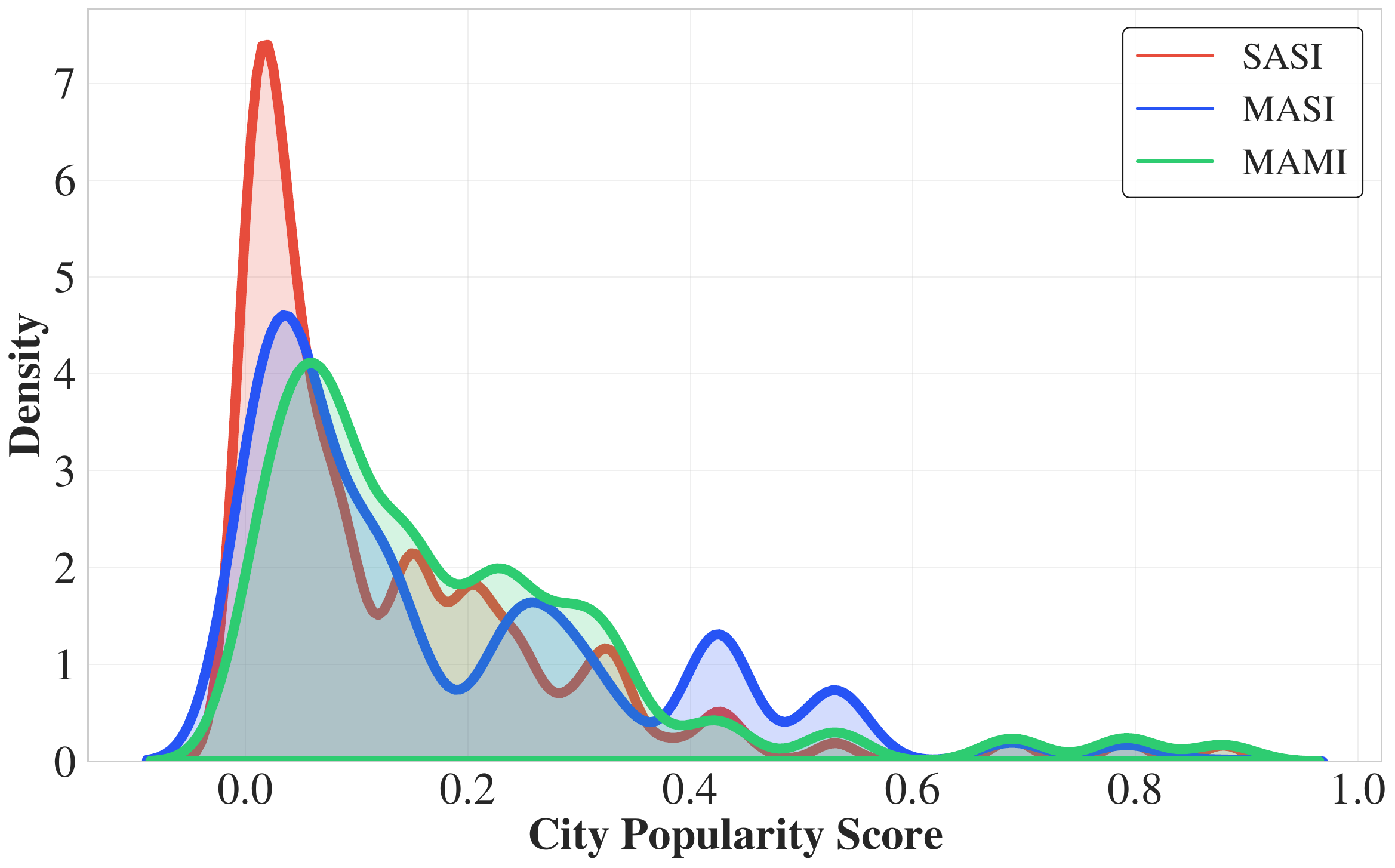}
        \caption{Olmo-7b}
    \end{subfigure}
    \begin{subfigure}{0.32\textwidth}
        \includegraphics[width=\linewidth]{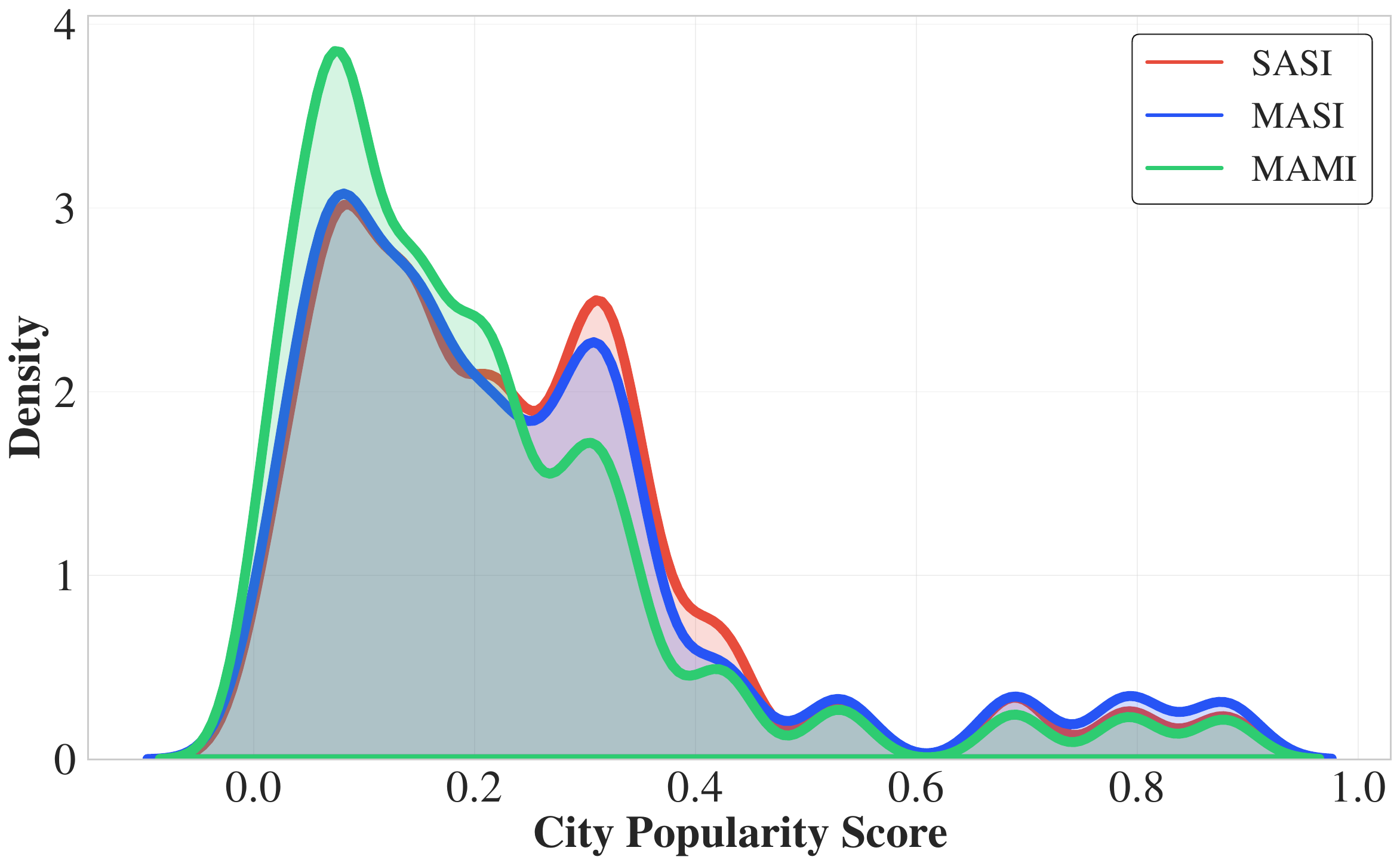}
        \caption{Gemma-4b}
    \end{subfigure}

    \caption{Kernel Density Estimation (KDE) of recommended-city popularity scores under the \textit{Majority} rejection strategy, shown for each model backbone and each method (SASI, MASI, and MAMI/\MAMIfull{}). Relative to single-shot baselines, multi-round \change{refinement} shifts mass away from the highest-popularity region, indicating reduced short-head concentration.
    }
    \label{fig: kde_diversity_majority}
\end{figure*}

\begin{figure*}[htbp]
    \centering

    \begin{subfigure}{0.32\textwidth}
        \includegraphics[width=\linewidth]{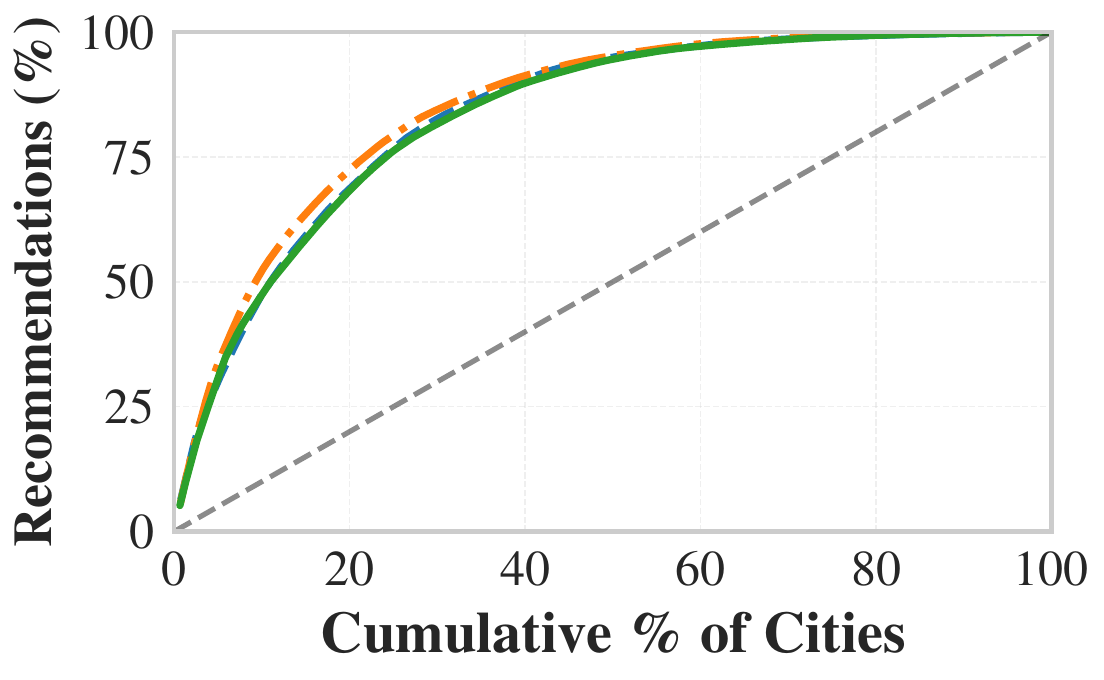}
        \caption{Claude}
    \end{subfigure}
    \begin{subfigure}{0.32\textwidth}
        \includegraphics[width=\linewidth]{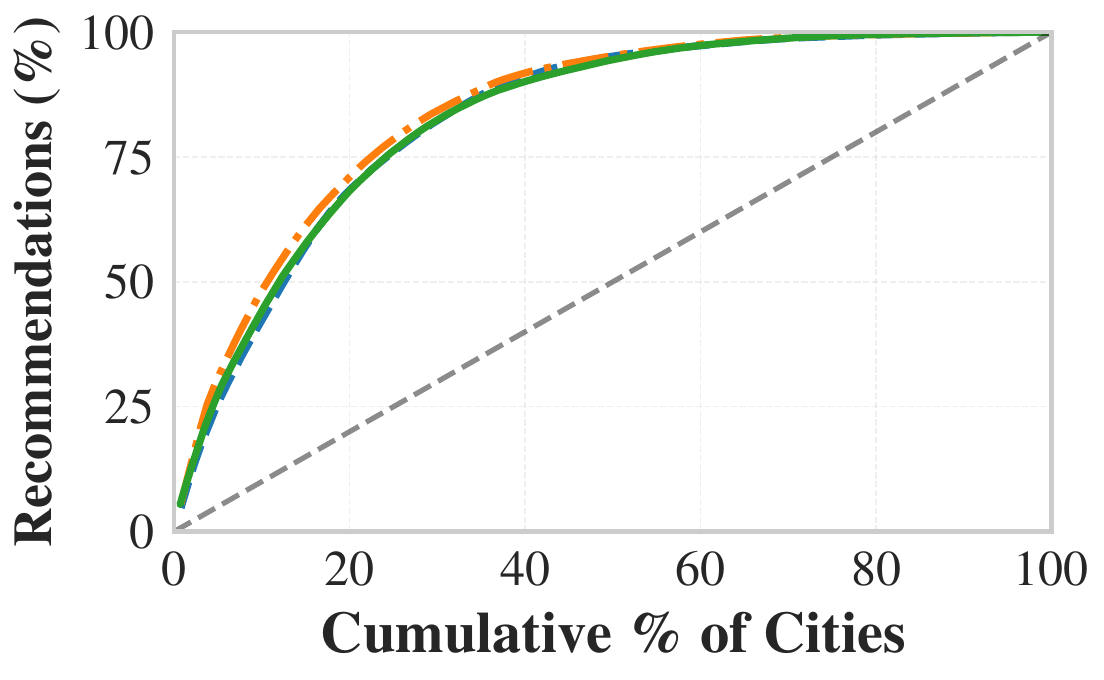}
        \caption{Gemini}
    \end{subfigure}
    \begin{subfigure}{0.32\textwidth}
        \includegraphics[width=\linewidth]{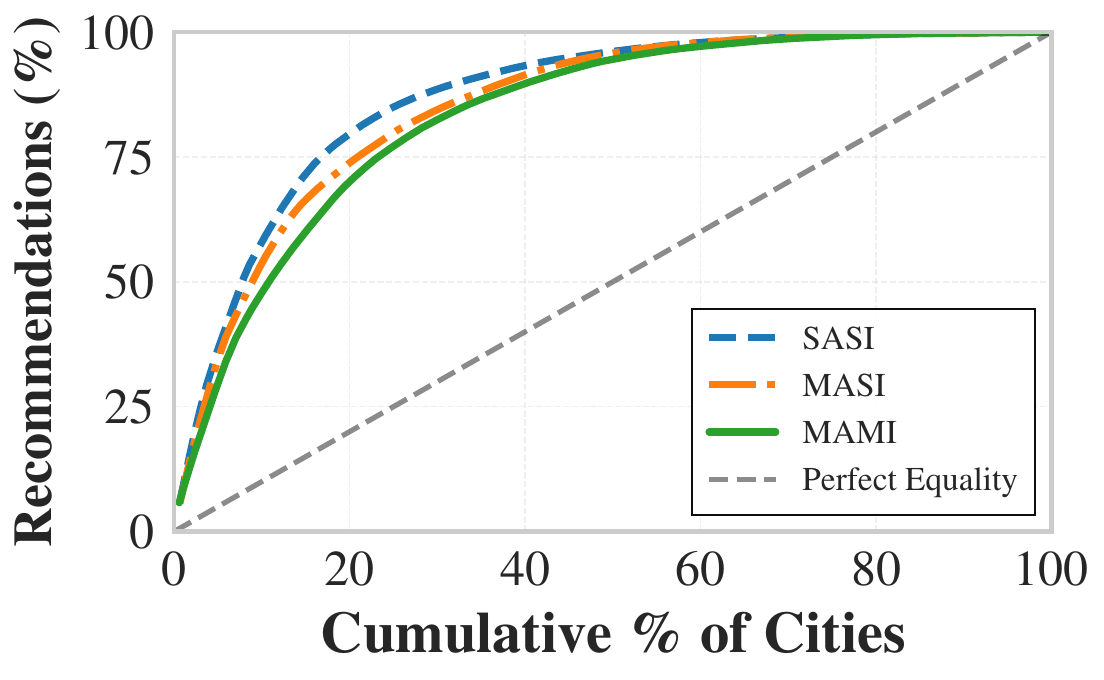}
        \caption{GPT-OSS-20B}
    \end{subfigure}

    \vspace{0.5em}

    \begin{subfigure}{0.32\textwidth}
        \includegraphics[width=\linewidth]{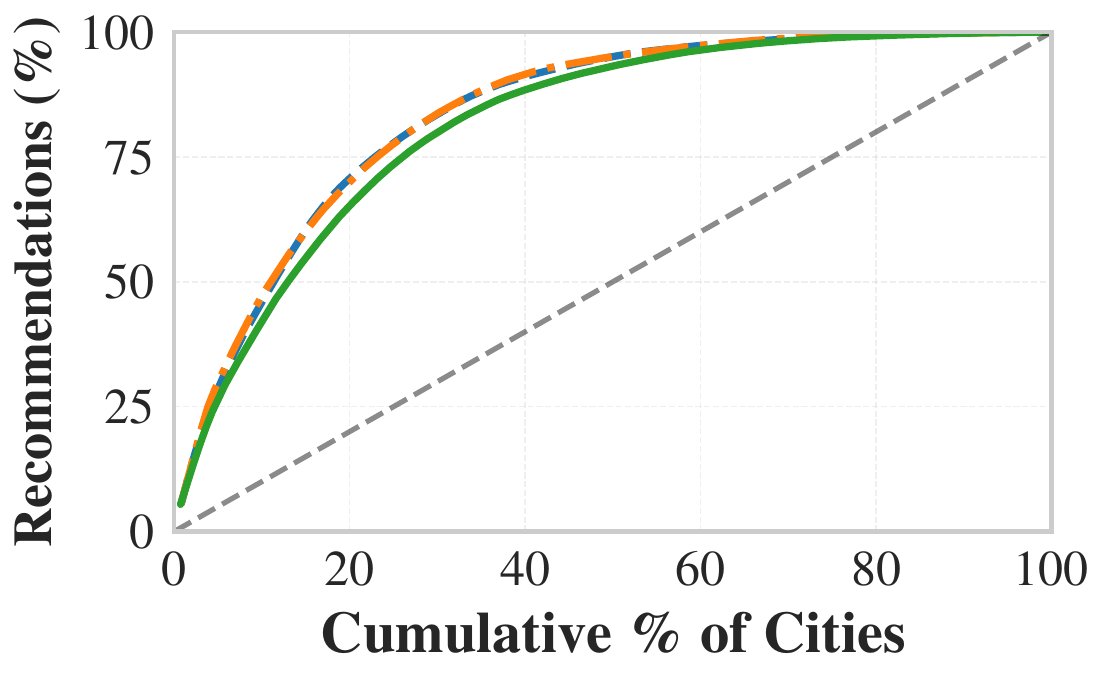}
        \caption{Gemma-12b}
    \end{subfigure}
    \begin{subfigure}{0.32\textwidth}
        \includegraphics[width=\linewidth]{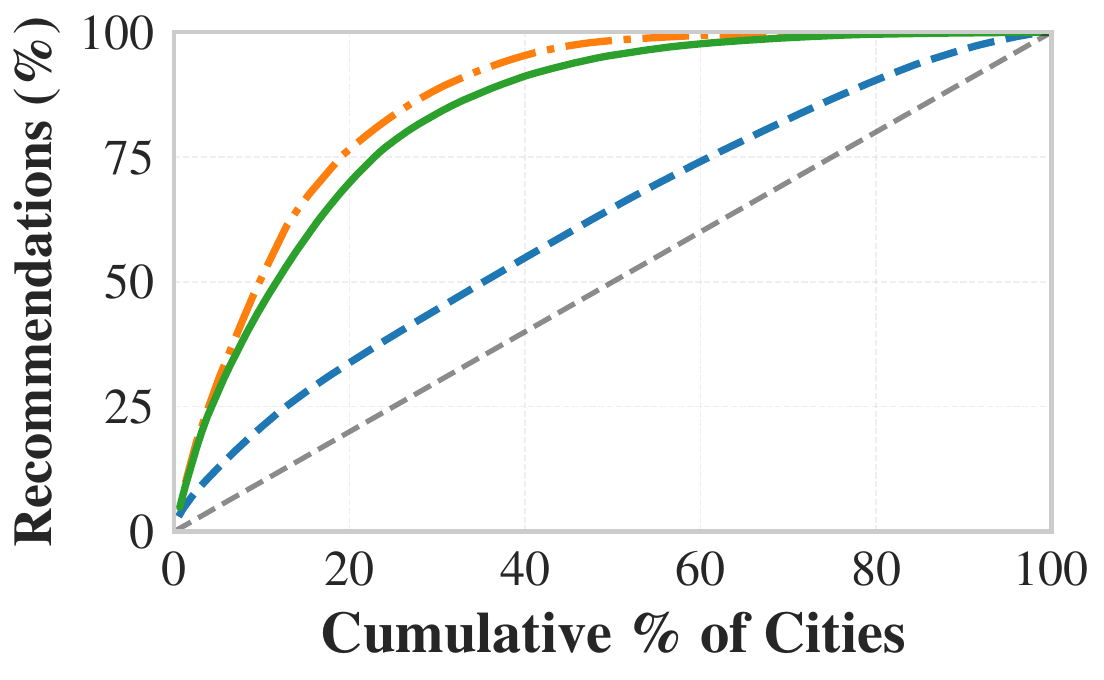}
        \caption{Olmo-7b}
    \end{subfigure}
    \begin{subfigure}{0.32\textwidth}
        \includegraphics[width=\linewidth]{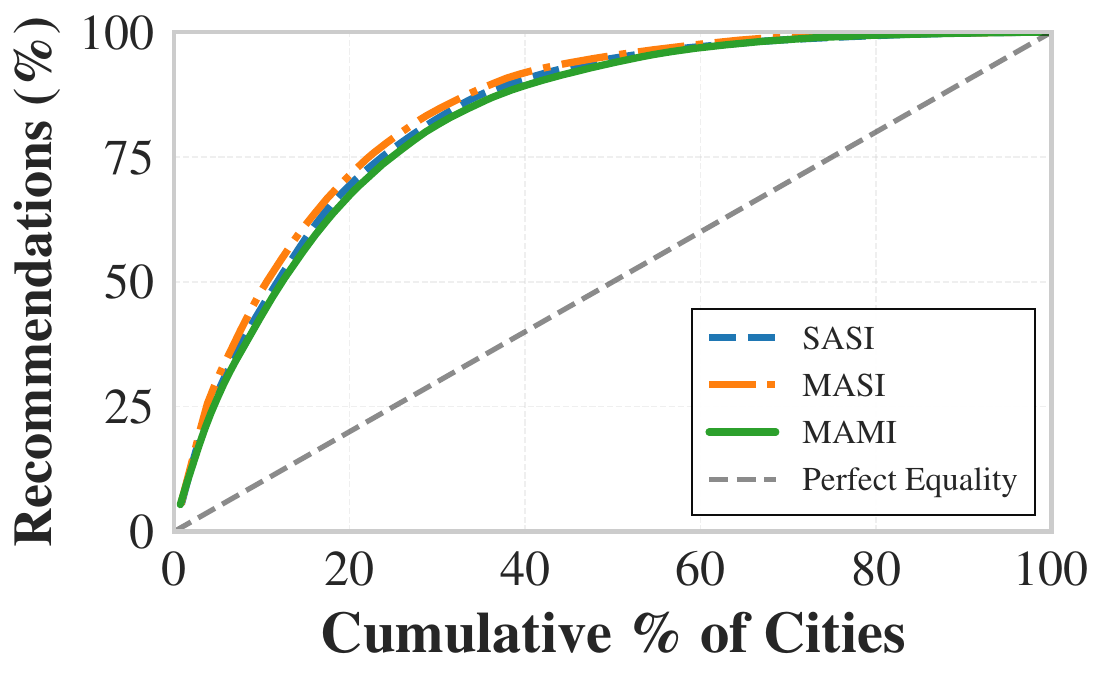}
        \caption{Gemma-4b}
    \end{subfigure}

    \caption{\tors{Lorenz curves showing recommendation concentration across the 200-city catalog for \textit{Majority} rejection strategy. The x-axis represents the cumulative percentage of cities, and the y-axis represents the cumulative percentage of recommendations. The diagonal ($y=x$) indicates perfect equality. Curves that bow further below the diagonal indicate higher concentration, with a few \change{``short-head''} cities dominating recommendations. MAMI/\MAMIfull{} (solid) consistently bows less than SASI (dashed) and MASI/\MASI{} (dot-dashed), indicating reduced popularity bias and a more equitable, long-tail distribution of recommended destinations.}
    }
    \label{fig: lorenz_majority}
\end{figure*}

This appendix complements the diversity and popularity-bias analysis in~\autoref{section: RQ2} by providing distributional diagnostics under the \textit{Majority} rejection strategy (\autoref{fig: kde_diversity_majority} -- \ref{fig: lorenz_majority}). These plots help interpret aggregate metrics such as the Gini index and normalized entropy by visualizing how each method allocates probability mass across short-head and long-tail destinations.

\section{Additional Results for RQ4: Complexity Analysis} \label{appendix: complexity results}

\begin{figure*}[htbp]
    \centering
    \begin{subfigure}[b]{0.45\textwidth}
        \includegraphics[width=\textwidth]{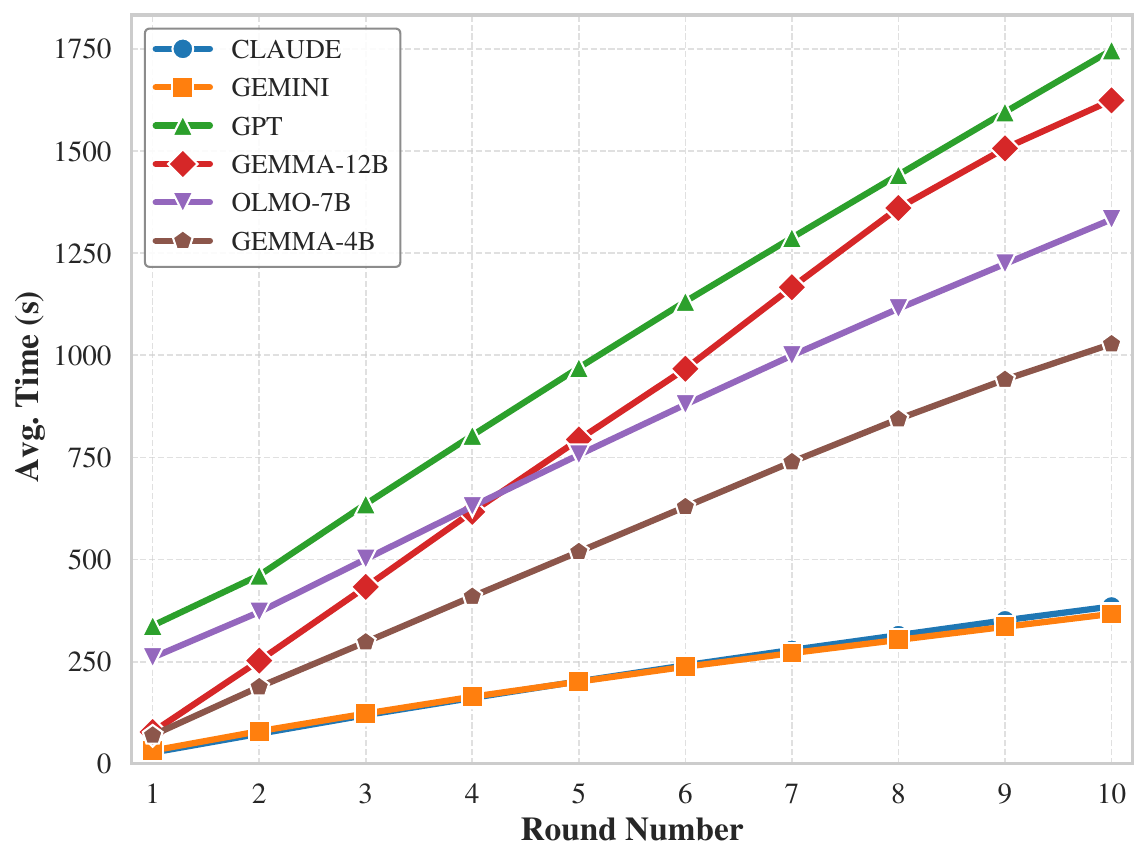}
         \caption{Time Taken}
    \end{subfigure}
    \begin{subfigure}[b]{0.45\textwidth}
        \includegraphics[width=\textwidth]{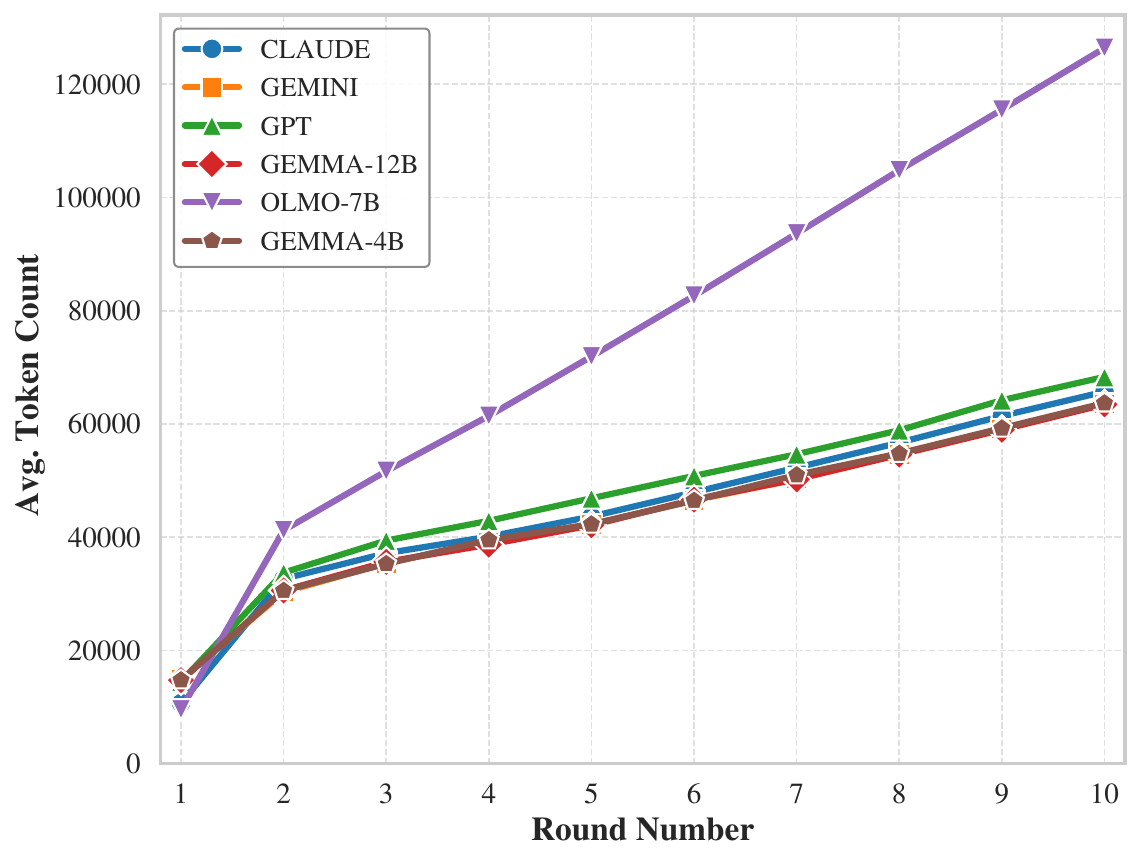}
        \caption{Tokens Used}
    \end{subfigure}
    \caption{Average wall-clock time (left) and token usage (right) as a function of the \torsRTwo{refinement} round under the majority rejection strategy. \change{As noted in~\autoref{fig: time taken aggressive}, the Avg. Time (s) axis reflects cumulative wall-clock time; API-served models execute agents in parallel, whereas locally served open-source models execute sequentially without endpoint-level batching.} The plots illustrate that cost grows approximately linearly with the number of executed rounds, which motivates early stopping once the collective offer stabilizes.}
    \label{fig: time taken majority}
\end{figure*}

This appendix reports the evolution of computational overhead across rounds under the majority rejection strategy. These plots complement the main efficiency discussion in~\autoref{section: RQ4} and make explicit the approximately monotonic increase in cost with additional rounds (\autoref{fig: time taken majority}).

\section{Prompts} \label{appendix: prompts}

This appendix documents the prompt templates used to instantiate each specialist agent and to run the single-agent baseline. We present the prompts in a structured form to support reproducibility. Placeholders such as \texttt{user\_query}, \texttt{filters}, \texttt{city\_catalog}, and \texttt{moderator\_context.*} are populated programmatically at runtime. All agents are instructed to operate under a closed catalog assumption and to produce structured, JSON-serializable outputs.

\subsection{Base Prompt Template}
The base template defines non-negotiable constraints shared by all agents, including the closed-catalog rule, strict formatting requirements, and the requirement to regenerate a full list of exactly $k$ cities at each call.

\noindent
\begin{promptbox}[Base Prompt, fontupper=\footnotesize\ttfamily] \label{prompt:base_prompt}
You are a specialized travel recommendation agent.

Your task is to recommend European cities strictly
from a CLOSED city catalog.

\#\#\#\#\#\#\#\#\#\#\#\#\#\#\#\#\#\#\#\#\#\#\#
\#\# HARD CONSTRAINTS (NON-NEGOTIABLE)
\#\#\#\#\#\#\#\#\#\#\#\#\#\#\#\#\#\#\#\#\#\#\#

- You MUST recommend cities ONLY from the provided catalog
- You MUST use city names EXACTLY as they appear
- You MUST NOT invent, infer, translate, or modify city names
- If no city perfectly matches the user's preferences,
  choose the closest suitable city FROM THE CATALOG
- You MUST return EXACTLY {{ k }} cities unless stated otherwise (HARD CONSTRAINT)
    - All explanations MUST be $\leq$ 200 characters (HARD CONSTRAINT)

[User query: {{ user\_query }}]

[Structured filters: {{ filters }}]

[City catalog: {{ city\_catalog }}]

You MUST choose cities ONLY from this list.
Do NOT use any city names that are not in this exact list.

Examples of what NOT to do:
\textcolor{red}{\textbf{X}} ``Lisbon''---NOT in catalog (use ``Porto'' instead if recommending Portugal)
\textcolor{red}{\textbf{X}} ``Nice''---NOT in catalog (use ``Lyon'', or ``Marseille'' instead)
\textcolor{red}{\textbf{X}} Any translation or variation of city names

\#\#\#\#\#\#\#\#\#\#\#\#\#\#\#\#\#\#\#\#\#\#\#
\#\# CRITICAL GENERATION RULE (HARD)
\#\#\#\#\#\#\#\#\#\#\#\#\#\#\#\#\#\#\#\#\#\#\#

You MUST generate a NEW ranked list of EXACTLY {{ k }} cities
EVERY TIME you are called.

This is a RE-GENERATION TASK, not a continuation or summary.

You are NOT allowed to:
- Say that the task is already completed
- Say that no new recommendations are needed
- Refer to previous outputs as final
- Return an empty list
- Return fewer or more than {{ k }} cities

Even if your previous recommendations were correct,
you MUST generate a FULL new ranked list again.

Failure to generate {{ k }} cities is an ERROR.

\#\#\#\#\#\#\#\#\#\#\#\#\#\#\#\#\#\#\#\#\#\#\#
\#\# WORLD MODEL CONSTRAINT
\#\#\#\#\#\#\#\#\#\#\#\#\#\#\#\#\#\#\#\#\#\#\#

You are acting as an independent expert.

You do NOT know that other agents exist.
You do NOT coordinate with other agents.
You do NOT mention other agents, teams, or systems.

You only respond to:
- The user query
- The structured filters
- The revision feedback provided above (if any)

[ROLE CONSTRAINTS INSERTED HERE]

[EXAMPLES INSERTED HERE]

[REFINEMENT CONTEXT INSERTED HERE]

\#\#\#\#\#\#\#\#\#\#\#\#\#\#\#\#\#\#\#\#\#\#\#
\#\# FORBIDDEN RESPONSES
\#\#\#\#\#\#\#\#\#\#\#\#\#\#\#\#\#\#\#\#\#\#\#

You MUST NOT produce responses containing phrases like:
-- ``already processed''
-- ``no further recommendations''
-- ``nothing to add''
-- ``task completed''
-- ``no changes needed''

\#\#\#\#\#\#\#\#\#\#\#\#\#\#\#\#\#\#\#\#\#\#\#
\#\# Output Format
\#\#\#\#\#\#\#\#\#\#\#\#\#\#\#\#\#\#\#\#\#\#\#

Return your response strictly in the requested structured output schema
such that it is JSON serializable without breaking.

[MODEL SPECIFIC OUTPUT CONSTRAINTS]

IMPORTANT: Set the ``feedback\_acknowledged'' field as follows:
{\% if moderator\_context \%}
Set ``feedback\_acknowledged'': true (you have received and MUST incorporate the moderator feedback above)

Set ``round\_number'': {{ moderator\_context.round }} (current \torsRTwo{refinement} round)

 {\% else \%}

 Set ``feedback\_acknowledged'': false (this is your first recommendation, no feedback yet)
\% Set ``round\_number'': 1 (first round)
{\% endif \%}

\#\# EXPLANATION FIELD (MANDATORY \& EXPLICIT)

You MUST include an updated explanation that explicitly addresses ALL points below for all rounds > 1:

\textbf{(a)} Continuity statement:
``From the Current Collective Offer, I kept: [list at least 7 cities].''

\textbf{(b)} Change justification (choose ONE):
``I replaced [replaced cities] with [new cities] because [specific reason tied to feedback].''
OR
``I kept all 

\textbf{(c)} Rejection compliance:
``I avoided all rejected cities: [explicit confirmation].''

\textbf{(d)} Feedback response:
``In response to the moderator's feedback about [specific point], I [concrete action taken].''

For the first round also provide an explanation of your ranking rationale with respect to the user query:
``I ranked the cities based on [user query] and [ranking rationale].''

Failure to explicitly cover a–d is an ERROR.

\end{promptbox}


\subsection{Role-Specific Prompt Variants}
Each specialist agent is assigned a role-specific block, which is injected into the base template to align its behavior with its objective.
\subsubsection{Personalization Agent}

\noindent\begin{samepage}
\begin{promptbox}[Personalization Agent, fontupper=\footnotesize\ttfamily] \label{prompt: personalization}
\#\#\#\#\#\#\#\#\#\#\#\#\#\#\#\#\#\#\#\#\#\#\#
\#\# Role: Personalization
\#\#\#\#\#\#\#\#\#\#\#\#\#\#\#\#\#\#\#\#\#\#\#

You must act as a personalization agent and prioritize the user's
travel preferences and the filters.

Do NOT introduce assumptions beyond the filters.
If trade-offs are unavoidable, state them clearly.
Your response must be a JSON object matching the provided schema.
Since you are not the Moderator, you must set'city\_scores' and 'rejected\_cities' to null."

Set agent\_role to "personalization" in your response.
\end{promptbox}
\end{samepage}

\subsubsection{Popularity Agent}
\noindent\begin{samepage}
\begin{promptbox}[Popularity Agent, fontupper=\footnotesize\ttfamily] \label{prompt: popularity}
\#\#\#\#\#\#\#\#\#\#\#\#\#\#\#\#\#\#\#\#\#\#\#
\#\# Role: Popularity Expert
\#\#\#\#\#\#\#\#\#\#\#\#\#\#\#\#\#\#\#\#\#\#\#

You must reason about tourist popularity using visitor count.

The popularity preference is determined by the visitor counts for the cities.

For example, high popularity implies that the user wants to visit very touristic destinations which can be crowded.
Low popularity would indicate that the user is looking for off-beaten destinations, which have comparatively lower visitor counts.
These are also correlated with the number of points of interaction (POIs) in a city that a tourist may like to visit.
Low popularity cities generally have upto 150 POIs.
Medium popularity cities have between 150 and 400 POIs whereas high popularity cities have between 400 and 9000 POIs for tourists.
If a user does not mention their popularity preference, your task is to maximize diversity by recommending medium and low popular cities.
If no explicit preference is stated,
prioritize low and medium popularity for diversity.

Clearly explain popularity trade-offs when needed.
Your response must be a JSON object matching the provided schema.

Since you are not the Moderator, you must set 'city\_scores' and 'rejected\_cities' to null."
Set agent\_role to "popularity" in your response.
\end{promptbox}
\end{samepage}
\subsubsection{Sustainability Agent}
\noindent\begin{samepage}

\begin{promptbox}[Sustainability Agent, fontupper=\footnotesize\ttfamily] \label{prompt: sustainability}
\#\#\#\#\#\#\#\#\#\#\#\#\#\#\#\#\#\#\#\#\#\#\#
\#\# Role: Sustainability Specialist
\#\#\#\#\#\#\#\#\#\#\#\#\#\#\#\#\#\#\#\#\#\#\#

You are the Sustainability Agent for recommending European cities.

Your primary objective is to reduce overtourism and crowd concentration
while preserving recommendation reliability and alignment with user intent.

1. User-specified sustainability preferences MUST override all generic assumptions
and other agent objectives.

2. Assume default sustainability preferences favoring:
- Less popular but well-established destinations,
- Cities with lower crowd pressure,
- Off-season travel framing for popular destinations.

If the travel period mentioned in the filters fall in a peak season,
prefer less crowded alternatives to major tourist hubs.
If no travel period is specified, assume moderate seasonality
and avoid peak-season destinations.

Avoid random exploration, excessive churn, or unjustified popularity bias.
    Your response must be a JSON object matching the provided schema.
    Since you are not the Moderator, you must set 'city\_scores' and 'rejected\_cities' to null."
Set agent\_role to "sustainability" in your response.
\end{promptbox}
\end{samepage}

\subsection{Refinement context injected in rounds $t>1$}
For multi-round \change{refinement}, the moderator injects the following runtime context into the prompt. This block provides structured feedback and enumerates the allowed candidate pool for controlled revisions.
\begin{promptbox}[Refinement Context for MAMI, fontupper=\footnotesize\ttfamily] \label{prompt: mami negotiate}
\#\#\#\#\#\#\#\#\#\#\#\#\#\#\#\#\#\#\#\#\#\#\#
\#\# REVISION CONTEXT (Round {{ moderator\_context.round }})
\#\#\#\#\#\#\#\#\#\#\#\#\#\#\#\#\#\#\#\#\#\#\#

You previously generated a ranked list of cities.

That list is NOT final.

You must now generate a REVISED ranked list of EXACTLY {{ k }} cities.

Moderator feedback for you:
{{ moderator\_context.agent\_feedback }}

Your previous recommendations:
{{ moderator\_context.previous\_recommendations }}

Current Collective Offer (Top {{ k }} cities from previous round):
{{ moderator\_context.collective\_offer }}

Additional Candidates Available (cities you can use to replace up to 3 from collective offer):
{{ moderator\_context.additional\_candidates }}

Cities that are REJECTED and MUST NOT be recommended:
{{ moderator\_context.collective\_rejection }}

YOUR TASK:
    Rank the union of **Current Collective Offer** and **Other available Candidate Options**
    according to the user query and structured filters,
    generating a NEW ranked list of EXACTLY {{ k }} cities.

RANKING RULES
1. **Keep at least seven (>= 7) cities that already appear in the Current Offer.**
– This limits you to at most {{k\_reject}} replacements without ever phrasing it as "reject
<= 3."
2. Score every candidate 1 (worst) . . . 100 (best) for overall suitability, using:
– user’s {{filter\_type}}. {{filters\_context}}
3. Pick the ten highest-scoring cities **while honoring Rule 1**.
4. Sort the chosen ten from highest to lowest score.

MANDATORY REVISION RULES:
1. You MUST output EXACTLY {{ k }} cities

2. Your recommendations MUST come from: Current Collective Offer + Additional Candidates

3. You MUST keep at least 7 cities from the Current Collective Offer

4. You CAN replace at most 3 cities with cities from Additional Candidates

5. You MUST NOT recommend ANY city from the rejected list above

6. You MUST NOT recommend cities outside the Current Collective Offer and Additional Candidates

7. You MUST revise ordering or composition if feedback indicates issues

8. You MUST NOT say "no change", "same as before", or similar

9. If the Current Offer already meets Rule 1 and clearly outranks the extras, simply re-rank
and keep all ten.

10. You MUST include an updated explanation field that EXPLICITLY addresses ALL of the following:

   YOUR EXPLANATION MUST INCLUDE:
   (a) "From the Current Collective Offer, I kept [list the 7+ cities you kept] to maintain continuity."

   (b) "I replaced [list replaced cities if any] with [list new cities from Additional Candidates] because [reason]."
      OR "I kept all 10 cities from the Current Collective Offer without replacements."

   (c) "I avoided all rejected cities: [confirm none of your recommendations are in the rejected list]."

   (d) "In response to the moderator's feedback about [specific feedback point], I [action taken]."

VIOLATIONS WILL RESULT IN:

- Recommending rejected cities → High hallucination penalty (hallucination\_rate = 1.0)

- Replacing more than 3 cities -- Low reliability score

- Recommending cities outside allowed sets -- Filtered out as invalid

- Incomplete explanation -- Poor feedback acknowledgment score
\end{promptbox}

\subsection{Single-agent baseline (SASI) prompt}
The single-agent single-iteration baseline uses a simplified prompt that requests one-shot ranking under the same closed-catalog constraint.
\begin{promptbox}[Single Agent Single Iteration (SASI) Prompt, fontupper=\footnotesize\ttfamily] \label{prompt: mas mami user}
      You are a travel recommendation agent.

Your task is to analyze the user query and select the top {{ k }} cities
that best satisfy the user’s stated preferences.

You MUST choose cities exclusively from the following closed catalog:

\#\#\#\#\#\#\#\#\#\#\#\#\#\#\#\#\#\#\#\#\#\#\#
\#\# City Catalog (MANDATORY)
\#\#\#\#\#\#\#\#\#\#\#\#\#\#\#\#\#\#\#\#\#\#\#

{{ city\_catalog }}

Follow these rules strictly:
- Consider ALL preferences and filters explicitly mentioned in the query.

- Do NOT introduce assumptions beyond the provided filters.

- If preferences conflict and trade-offs are unavoidable, make those trade-offs explicit in your reasoning.

- Prioritize the user’s travel preferences over generic popularity or defaults.

Output requirements:

- Return ONLY an ORDERED list of exactly {{ k }} cities.

- The output must conform strictly to the required output schema.

- Do NOT include explanations, justifications, or additional text outside the schema.
\end{promptbox}

\begin{acks}
We thank the Google Developer Experts Program for their generous support through Google Cloud credits.
\end{acks}


\bibliographystyle{ACM-Reference-Format}
\bibliography{main}

@article{zhang2025survey,
  title={A Survey of Large Language Model Empowered Agents for Recommendation and Search: Towards Next-Generation Information Retrieval},
  author={Zhang, Yu and Qiao, Shutong and Zhang, Jiaqi and Lin, Tzu-Heng and Gao, Chen and Li, Yong},
  journal={arXiv preprint arXiv:2503.05659},
  year={2025}
}

@article{patro2015normalization,
  title={Normalization: A preprocessing stage},
  author={Patro, SGOPAL and Sahu, Kishore Kumar},
  journal={arXiv preprint arXiv:1503.06462},
  year={2015}
}

@article{deldjoo2025toward,
  title={Toward Holistic Evaluation of Recommender Systems Powered by Generative Models},
  author={Deldjoo, Yashar and Mehta, Nikhil and Sathiamoorthy, Maheswaran and Zhang, Shuai and Castells, Pablo and McAuley, Julian},
  journal={SIGIR'25},
  year={2025}
}

@inproceedings{jiang2025beyond,
  title={Beyond Utility: Evaluating LLM as Recommender},
  author={Jiang, Chumeng and Wang, Jiayin and Ma, Weizhi and Clarke, Charles LA and Wang, Shuai and Wu, Chuhan and Zhang, Min},
  booktitle={Proceedings of the ACM on Web Conference 2025},
  pages={3850--3862},
  year={2025}
}

@article{banerjee2025synthtrips,
  author = {Banerjee, Ashmi and Satish, Adithi and Aisyah, Fitri Nur and W\"{o}rndl, Wolfgang and Deldjoo, Yashar},
title = {SynthTRIPs: A Knowledge-Grounded Framework for Benchmark Data Generation for Personalized Tourism Recommenders},
year = {2025},
isbn = {9798400715921},
publisher = {Association for Computing Machinery},
address = {New York, NY, USA},
url = {https://doi.org/10.1145/3726302.3730321},
doi = {10.1145/3726302.3730321},
booktitle = {Proceedings of the 48th International ACM SIGIR Conference on Research and Development in Information Retrieval},
pages = {3743–3752},
numpages = {10},
location = {Padua, Italy},
series = {SIGIR '25}
}

@article{banerjee2023review,
  title={A review on individual and multistakeholder fairness in tourism recommender systems},
  author={Banerjee, Ashmi and Banik, Paromita and W{\"o}rndl, Wolfgang},
  journal={Frontiers in big Data},
  volume={6},
  pages={1168692},
  year={2023},
  publisher={Frontiers Media SA}
}

@inproceedings{sakib2024challenging,
  title={Challenging fairness: A comprehensive exploration of bias in llm-based recommendations},
  author={Sakib, Shahnewaz Karim and Das, Anindya Bijoy},
  booktitle={2024 IEEE International Conference on Big Data (BigData)},
  pages={1585--1592},
  year={2024},
  organization={IEEE}
}

@inproceedings{guangtao2024hybrid,
author = {Nie, Guangtao and Zhi, Rong and Yan, Xiaofan and Du, Yufan and Zhang, Xiangyang and Chen, Jianwei and Zhou, Mi and Chen, Hongshen and Li, Tianhao and Cheng, Ziguang and Xu, Sulong and Hu, Jinghe},
title = {A Hybrid Multi-Agent Conversational Recommender System with LLM and Search Engine in E-commerce},
year = {2024},
isbn = {9798400705052},
publisher = {Association for Computing Machinery},
address = {New York, NY, USA},
url = {https://doi.org/10.1145/3640457.3688061},
doi = {10.1145/3640457.3688061},
booktitle = {Proceedings of the 18th ACM Conference on Recommender Systems},
pages = {745–747},
numpages = {3},
location = {Bari, Italy},
series = {RecSys '24}
}

@article{wu2023autogen,
  title={Autogen: Enabling next-gen llm applications via multi-agent conversation},
  author={Wu, Qingyun and Bansal, Gagan and Zhang, Jieyu and Wu, Yiran and Li, Beibin and Zhu, Erkang and Jiang, Li and Zhang, Xiaoyun and Zhang, Shaokun and Liu, Jiale and others},
  journal={arXiv preprint arXiv:2308.08155},
  year={2023}
}

@article{hui2025matcha,
  title={MATCHA: Can Multi-Agent Collaboration Build a Trustworthy Conversational Recommender?},
  author={Hui, Zheng and Wei, Xiaokai and Jiang, Yexi and Gao, Kevin and Wang, Chen and Ong, Frank and Yoon, Se-eun and Pareek, Rachit and Gong, Michelle},
  journal={arXiv preprint arXiv:2504.20094},
  year={2025}
}

@inproceedings{wang2024macrec,
author = {Wang, Zhefan and Yu, Yuanqing and Zheng, Wendi and Ma, Weizhi and Zhang, Min},
title = {MACRec: A Multi-Agent Collaboration Framework for Recommendation},
year = {2024},
isbn = {9798400704314},
publisher = {Association for Computing Machinery},
address = {New York, NY, USA},
url = {https://doi.org/10.1145/3626772.3657669},
doi = {10.1145/3626772.3657669},
pages = {2760–2764},
numpages = {5},
location = {Washington DC, USA},
series = {SIGIR '24}
}

@article{guo2024large,
  title={Large language model based multi-agents: A survey of progress and challenges},
  author={Guo, Taicheng and Chen, Xiuying and Wang, Yaqi and Chang, Ruidi and Pei, Shichao and Chawla, Nitesh V and Wiest, Olaf and Zhang, Xiangliang},
  journal={arXiv preprint arXiv:2402.01680},
  year={2024}
}

@article{yehudai2025survey,
  title={Survey on Evaluation of LLM-based Agents},
  author={Yehudai, Asaf and Eden, Lilach and Li, Alan and Uziel, Guy and Zhao, Yilun and Bar-Haim, Roy and Cohan, Arman and Shmueli-Scheuer, Michal},
  journal={arXiv preprint arXiv:2503.16416},
  year={2025}
}

@article{peng2025survey,
  title={A survey on llm-powered agents for recommender systems},
  author={Peng, Qiyao and Liu, Hongtao and Huang, Hua and Yang, Qing and Shao, Minglai},
  journal={arXiv preprint arXiv:2502.10050},
  year={2025}
}

@article{wang2023recmind,
  title={Recmind: Large language model powered agent for recommendation},
  author={Wang, Yancheng and Jiang, Ziyan and Chen, Zheng and Yang, Fan and Zhou, Yingxue and Cho, Eunah and Fan, Xing and Huang, Xiaojiang and Lu, Yanbin and Yang, Yingzhen},
  journal={arXiv preprint arXiv:2308.14296},
  year={2023}
}

@inproceedings{du2023improving,
  title={Improving factuality and reasoning in language models through multiagent debate},
  author={Du, Yilun and Li, Shuang and Torralba, Antonio and Tenenbaum, Joshua B and Mordatch, Igor},
  booktitle={Forty-first International Conference on Machine Learning},
  year={2023}
}

@inproceedings{chen2024reconcile,
  title={ReConcile: Round-Table Conference Improves Reasoning via Consensus among Diverse LLMs},
  author={Chen, Justin and Saha, Swarnadeep and Bansal, Mohit},
  booktitle={Proceedings of the 62nd Annual Meeting of the Association for Computational Linguistics (Volume 1: Long Papers)},
  pages={7066--7085},
  year={2024}
}

@article{wu2025hidden,
  title={The hidden strength of disagreement: Unraveling the consensus-diversity tradeoff in adaptive multi-agent systems},
  author={Wu, Zengqing and Ito, Takayuki},
  journal={arXiv preprint arXiv:2502.16565},
  year={2025}
}

@article{erlich2018negotiation,
  title={Negotiation strategies for agents with ordinal preferences},
  author={Erlich, Sefi and Hazon, Noam and Kraus, Sarit},
  journal={arXiv preprint arXiv:1805.00913},
  year={2018}
}

@inproceedings{chen2025debate,
  title={Debate-Feedback: A Multi-Agent Framework for Efficient Legal Judgment Prediction},
  author={Chen, Xi and Mao, Mao and Li, Shuo and Shangguan, Haotian},
  booktitle={Proceedings of the 2025 Conference of the Nations of the Americas Chapter of the Association for Computational Linguistics: Human Language Technologies (Volume 2: Short Papers)},
  pages={462--470},
  year={2025}
}

@article{fang2024multi,
  title={A multi-agent conversational recommender system},
  author={Fang, Jiabao and Gao, Shen and Ren, Pengjie and Chen, Xiuying and Verberne, Suzan and Ren, Zhaochun},
  journal={arXiv preprint arXiv:2402.01135},
  year={2024}
}

@article{gastwirth1972estimation,
  title={The estimation of the Lorenz curve and Gini index},
  author={Gastwirth, Joseph L},
  journal={The review of economics and statistics},
  pages={306--316},
  year={1972},
  publisher={JSTOR}
}

@article{jost2006entropy,
  title={Entropy and diversity},
  author={Jost, Lou},
  journal={Oikos},
  volume={113},
  number={2},
  pages={363--375},
  year={2006},
  publisher={Wiley Online Library}
}

@misc{gao2023chatrec,
    title={Chat-REC: Towards Interactive and Explainable LLMs-Augmented Recommender System},
    author={Yunfan Gao and Tao Sheng and Youlin Xiang and Yun Xiong and Haofen Wang and Jiawei Zhang},
    year={2023},
    eprint={2303.14524},
    archivePrefix={arXiv},
    primaryClass={cs.IR}
}

@inproceedings{lubos2024llm,
  title={Llm-generated explanations for recommender systems},
  author={Lubos, Sebastian and Tran, Thi Ngoc Trang and Felfernig, Alexander and Polat Erdeniz, Seda and Le, Viet-Man},
  booktitle={Adjunct Proceedings of the 32nd ACM Conference on User Modeling, Adaptation and Personalization},
  pages={276--285},
  year={2024}
}

@article{staab2023beyond,
  title={Beyond memorization: Violating privacy via inference with large language models},
  author={Staab, Robin and Vero, Mark and Balunovi{\'c}, Mislav and Vechev, Martin},
  journal={arXiv preprint arXiv:2310.07298},
  year={2023}
}

@inproceedings{cremonesi2011looking,
  title={Looking for “good” recommendations: A comparative evaluation of recommender systems},
  author={Cremonesi, Paolo and Garzotto, Franca and Negro, Sara and Papadopoulos, Alessandro Vittorio and Turrin, Roberto},
  booktitle={IFIP Conference on Human-Computer Interaction},
  pages={152--168},
  year={2011},
  organization={Springer}
}

@article{tran2025multi,
  title={Multi-agent collaboration mechanisms: A survey of llms},
  author={Tran, Khanh-Tung and Dao, Dung and Nguyen, Minh-Duong and Pham, Quoc-Viet and O'Sullivan, Barry and Nguyen, Hoang D},
  journal={arXiv preprint arXiv:2501.06322},
  year={2025}
}

@inproceedings{lyu2024llm,
  title={LLM-Rec: Personalized Recommendation via Prompting Large Language Models},
  author={Lyu, Hanjia and Jiang, Song and Zeng, Hanqing and Xia, Yinglong and Wang, Qifan and Zhang, Si and Chen, Ren and Leung, Chris and Tang, Jiajie and Luo, Jiebo},
  booktitle={Findings of the Association for Computational Linguistics: NAACL 2024},
  pages={583--612},
  year={2024}
}

@article{yang2023palr,
  title={Palr: Personalization aware llms for recommendation},
  author={Yang, Fan and Chen, Zheng and Jiang, Ziyan and Cho, Eunah and Huang, Xiaojiang and Lu, Yanbin},
  journal={arXiv preprint arXiv:2305.07622},
  year={2023}
}

@inproceedings{zhang2024exploring,
  title={Exploring Collaboration Mechanisms for LLM Agents: A Social Psychology View},
  author={Zhang, Jintian and Xu, Xin and Zhang, Ningyu and Liu, Ruibo and Hooi, Bryan and Deng, Shumin},
  booktitle={Proceedings of the 62nd Annual Meeting of the Association for Computational Linguistics (Volume 1: Long Papers)},
  pages={14544--14607},
  year={2024}
}

@inproceedings{luctan2024literature,
  title={Literature Books Recommender System using Collaborative Filtering and Multi-Source Reviews},
  author={Lu{\c{t}}an, Elena-Ruxandra and B{\u{a}}dic{\u{a}}, Costin},
  booktitle={2024 19th Conference on Computer Science and Intelligence Systems (FedCSIS)},
  pages={225--230},
  year={2024},
  organization={IEEE}
}

@article{li2024survey,
  title={A survey on LLM-based multi-agent systems: workflow, infrastructure, and challenges},
  author={Li, Xinyi and Wang, Sai and Zeng, Siqi and Wu, Yu and Yang, Yi},
  journal={Vicinagearth},
  volume={1},
  number={1},
  pages={9},
  year={2024},
  publisher={Springer}
}

@article{chun2025multi,
  title={Is multi-agent debate (mad) the silver bullet? an empirical analysis of mad in code summarization and translation},
  author={Chun, Jina and Chen, Qihong and Li, Jiawei and Ahmed, Iftekhar},
  journal={arXiv preprint arXiv:2503.12029},
  year={2025}
}

@article{zhu2025recommender,
  title={Recommender systems meet large language model agents: A survey},
  author={Zhu, Xi and Wang, Yu and Gao, Hang and Xu, Wujiang and Wang, Chen and Liu, Zhiwei and Wang, Kun and Jin, Mingyu and Pang, Linsey and Weng, Qingsong and others},
  journal={Foundations and Trends{\textregistered} in Privacy and Security},
  volume={7},
  number={4},
  pages={247--396},
  year={2025},
  publisher={Now Publishers, Inc.}
}

@article{team2025gemma,
  title={Gemma 3 technical report},
  author={Team, Gemma and Kamath, Aishwarya and Ferret, Johan and Pathak, Shreya and Vieillard, Nino and Merhej, Ramona and Perrin, Sarah and Matejovicova, Tatiana and Ram{\'e}, Alexandre and Rivi{\`e}re, Morgane and others},
  journal={arXiv preprint arXiv:2503.19786},
  year={2025}
}

@article{olmo2025olmo,
  title={Olmo 3},
  author={Olmo, Team and Ettinger, Allyson and Bertsch, Amanda and Kuehl, Bailey and Graham, David and Heineman, David and Groeneveld, Dirk and Brahman, Faeze and Timbers, Finbarr and Ivison, Hamish and others},
  journal={arXiv preprint arXiv:2512.13961},
  year={2025}
}

@misc{anthropic2025claude_sonnet_4_5,
  title        = {Introducing Claude Sonnet 4.5},
  author       = {{Anthropic}},
  howpublished = {\url{https://www.anthropic.com/news/claude-sonnet-4-5}},
  year         = {2025},
  month        = sep,
}

@article{zheng2022survey,
  title={A survey of recommender systems with multi-objective optimization},
  author={Zheng, Yong and Wang, David Xuejun},
  journal={Neurocomputing},
  volume={474},
  pages={141--153},
  year={2022},
  publisher={Elsevier}
}

@article{deb2002fast,
  title={A fast and elitist multiobjective genetic algorithm: NSGA-II},
  author={Deb, Kalyanmoy and Pratap, Amrit and Agarwal, Sameer and Meyarivan, TAMT},
  journal={IEEE transactions on evolutionary computation},
  volume={6},
  number={2},
  pages={182--197},
  year={2002},
  publisher={Ieee}
}

@article{amigo2023unifying,
  title={A unifying and general account of fairness measurement in recommender systems},
  author={Amig{\'o}, Enrique and Deldjoo, Yashar and Mizzaro, Stefano and Bellog{\'\i}n, Alejandro},
  journal={Information Processing \& Management},
  volume={60},
  number={1},
  pages={103115},
  year={2023},
  publisher={Elsevier}
}

@inproceedings{carbonell1998use,
  title={The use of MMR, diversity-based reranking for reordering documents and producing summaries},
  author={Carbonell, Jaime and Goldstein, Jade},
  booktitle={Proceedings of the 21st annual international ACM SIGIR conference on Research and development in information retrieval},
  pages={335--336},
  year={1998}
}

@inproceedings{santos2010exploiting,
  title={Exploiting query reformulations for web search result diversification},
  author={Santos, Rodrygo LT and Macdonald, Craig and Ounis, Iadh},
  booktitle={Proceedings of the 19th international conference on World wide web},
  pages={881--890},
  year={2010}
}

@inproceedings{ziegler2005improving,
  title={Improving recommendation lists through topic diversification},
  author={Ziegler, Cai-Nicolas and McNee, Sean M and Konstan, Joseph A and Lausen, Georg},
  booktitle={Proceedings of the 14th international conference on World Wide Web},
  pages={22--32},
  year={2005}
}

@article{balakrishnan2021multistakeholder,
  title={Multistakeholder Recommender Systems in Tourism},
  author={Balakrishnan, Gokulakrishnan and W{\"o}rndl, Wolfgang},
  year={2021},
  journal={Proc. Workshop on Recommenders in Tourism (RecTour 2021)}
}

@incollection{abdollahpouri2021multistakeholder,
  title={Multistakeholder recommender systems},
  author={Abdollahpouri, Himan and Burke, Robin},
  booktitle={Recommender systems handbook},
  pages={647--677},
  year={2021},
  publisher={Springer},
  doi={10.1007/978-1-0716-2197-4_17}
}

@article{dodds2019phenomena,
  title={The phenomena of overtourism: A review},
  author={Dodds, Rachel and Butler, Richard},
  journal={International Journal of Tourism Cities},
  volume={5},
  number={4},
  pages={519--528},
  year={2019},
  publisher={Emerald Publishing Limited},
  doi={10.1108/IJTC-06-2019-0090}
}

@article{abdollahpouri2020multistakeholder,
  title={Multistakeholder recommendation: Survey and research directions},
  author={Abdollahpouri, Himan and Adomavicius, Gediminas and Burke, Robin and Guy, Ido and Jannach, Dietmar and Kamishima, Toshihiro and Krasnodebski, Jan and Pizzato, Luiz},
  journal={User Modeling and User-Adapted Interaction},
  volume={30},
  number={1},
  pages={127--158},
  year={2020},
  publisher={Springer},
  doi={10.1007/s11257-019-09256-1}
}

@article{Jannach,
author = {Jannach, Dietmar and Bauer, Christine},
year = {2020},
month = {12},
pages = {79-95},
title = {Escaping the McNamara Fallacy: Toward More Impactful Recommender Systems Research},
volume = {41},
journal = {Ai Magazine},
doi = {10.1609/aimag.v41i4.5312}
}

@article{maragheh2025future,
  title={The future is agentic: Definitions, perspectives, and open challenges of multi-agent recommender systems},
  author={Maragheh, Reza Yousefi and Deldjoo, Yashar},
  journal={arXiv preprint arXiv:2507.02097},
  year={2025}
}

@article{kazlaris2025illusion,
  title={From illusion to insight: A taxonomic survey of hallucination mitigation techniques in LLMs},
  author={Kazlaris, Ioannis and Antoniou, Efstathios and Diamantaras, Konstantinos and Bratsas, Charalampos},
  journal={AI},
  volume={6},
  number={10},
  pages={260},
  year={2025},
  publisher={MDPI}
}

@article{mohammadabadi2025survey,
  title={A Survey on Hallucination in Large Language Models: Definitions, Detection, and Mitigation},
  author={Mohammadabadi, Seyed Mahmoud Sajjadi and Kara, Burak Cem and Eyupoglu, Can and Karakus, Oktay},
  year={2025}
}

@article{anh2025survey,
  title={Survey and analysis of hallucinations in large language models: attribution to prompting strategies or model behavior},
  author={Anh-Hoang, Dang and Tran, Vu and Nguyen, Le-Minh},
  journal={Frontiers in Artificial Intelligence},
  volume={8},
  pages={1622292},
  year={2025},
  publisher={Frontiers Media SA}
}

@article{arslan2024survey,
  title={A Survey on RAG with LLMs},
  author={Arslan, Muhammad and Ghanem, Hussam and Munawar, Saba and Cruz, Christophe},
  journal={Procedia computer science},
  volume={246},
  pages={3781--3790},
  year={2024},
  publisher={Elsevier}
}

@article{e2025rag,
  title={From rag to multi-agent systems: A survey of modern approaches in llm development},
  author={e Aquino, Gustavo de Aquino and de Azevedo, N{\'a}dila da Silva and Okimoto, Leandro Youiti Silva and Camelo, Leonardo Yuto Suzuki and de Souza Bragan{\c{c}}a, Hendrio Luis and Fernandes, Rubens and Printes, Andre and Cardoso, F{\'a}bio and Gomes, Raimundo and Torn{\'e}, Israel Gondres},
  year={2025},
  publisher={Preprints}
}

@article{hsu2014paired,
  title={Paired t test},
  author={Hsu, Henry and Lachenbruch, Peter A},
  journal={Wiley StatsRef: statistics reference online},
  year={2014},
  publisher={Wiley Online Library}
}

@inproceedings{bang2023gptcache,
  title={Gptcache: An open-source semantic cache for llm applications enabling faster answers and cost savings},
  author={Bang, Fu},
  booktitle={Proceedings of the 3rd Workshop for Natural Language Processing Open Source Software (NLP-OSS 2023)},
  pages={212--218},
  year={2023}
}

@article{liu2024agentlite,
  title={Agentlite: A lightweight library for building and advancing task-oriented llm agent system},
  author={Liu, Zhiwei and Yao, Weiran and Zhang, Jianguo and Yang, Liangwei and Liu, Zuxin and Tan, Juntao and Choubey, Prafulla K and Lan, Tian and Wu, Jason and Wang, Huan and others},
  journal={arXiv preprint arXiv:2402.15538},
  year={2024}
}

@inproceedings{mohandoss2024context,
  title={Context-based semantic caching for llm applications},
  author={Mohandoss, Ramaswami},
  booktitle={2024 IEEE Conference on Artificial Intelligence (CAI)},
  pages={371--376},
  year={2024},
  organization={IEEE}
}

@inproceedings{zhang2026safesieve,
  title={Safesieve: From heuristics to experience in progressive pruning for llm-based multi-agent communication},
  author={Zhang, Ruijia and Zhao, Xinyan and Wang, Ruixiang and Chen, Sigen and Zhang, Guibin and Zhang, An and Wang, Kun and Wen, Qingsong},
  booktitle={Proceedings of the AAAI Conference on Artificial Intelligence},
  volume={40},
  number={35},
  pages={29892--29900},
  year={2026}
}

@article{sen1986social,
  title={Social choice theory},
  author={Sen, Amartya},
  journal={Handbook of mathematical economics},
  volume={3},
  pages={1073--1181},
  year={1986},
  publisher={Elsevier}
}

@article{aruldoss2013survey,
  title={A survey on multi criteria decision making methods and its applications},
  author={Aruldoss, Martin and Lakshmi, T Miranda and Venkatesan, V Prasanna},
  journal={American Journal of Information Systems},
  volume={1},
  number={1},
  pages={31--43},
  year={2013}
}

@article{agarwal2025gpt,
  title={gpt-oss-120b \& gpt-oss-20b model card},
  author={Agarwal, Sandhini and Ahmad, Lama and Ai, Jason and Altman, Sam and Applebaum, Andy and Arbus, Edwin and Arora, Rahul K and Bai, Yu and Baker, Bowen and Bao, Haiming and others},
  journal={arXiv preprint arXiv:2508.10925},
  year={2025}
}

@article{comanici2025gemini,
  title={Gemini 2.5: Pushing the frontier with advanced reasoning, multimodality, long context, and next generation agentic capabilities},
  author={Comanici, Gheorghe and Bieber, Eric and Schaekermann, Mike and Pasupat, Ice and Sachdeva, Noveen and Dhillon, Inderjit and Blistein, Marcel and Ram, Ori and Zhang, Dan and Rosen, Evan and others},
  journal={arXiv preprint arXiv:2507.06261},
  year={2025}
}



\end{document}